\documentclass{ieeeaccess}
\usepackage{cite}
\usepackage{url}
\usepackage[utf8]{inputenc}
\usepackage{amsmath,amssymb,amsfonts}
\usepackage{graphicx}
\usepackage{textcomp}
\usepackage{bm}
\usepackage{bbm}
\usepackage{booktabs}
\usepackage{multirow}
\usepackage{xcolor}

\usepackage{threeparttable}
\usepackage{calc}
\makeatletter
\let\caption@ieeeaccess@makecaption\@makecaption
\AtBeginDocument{\let\@makecaption\caption@ieeeaccess@makecaption}
\makeatother
\usepackage{subcaption}

\definecolor{lightpurple}{RGB}{200, 160, 255}
\definecolor{lightgreen}{RGB}{144, 238, 144}

\makeatletter
\def\@IEEEprocessthesectionargument#1{%
\@ifmtarg{#1}{%
\section*{\thesectiondis}%
\addcontentsline{toc}{section}{Appendix \thesection}}{%
\section*{\thesectiondis \\* #1}%
\addcontentsline{toc}{section}{Appendix \thesection: #1}}}
\makeatother

\newcommand{\imagewithlabel}[6][black]{%
    \begin{subfigure}[b]{#4}%
        \captionsetup[subfigure]{aboveskip=2pt,belowskip=0pt,skip=0pt,margin=0pt,font=footnotesize}%
        \begingroup
        \setlength{\fboxsep}{1.5pt}%
        \sbox0{\includegraphics[width=\linewidth]{#2}}%
        \sbox1{\fcolorbox{#1}{#1}{\textcolor{white}{\sffamily\tiny\,#3}}}%
        \raisebox{0pt}[\ht0][\dp0]{%
            \makebox[\wd0][l]{%
                \rlap{\copy0}%
                \rlap{\raisebox{\dimexpr\dp1-\dp0\relax}{\copy1}}%
            }%
        }%
        \endgroup
        \subcaption{}%
        \if\relax\detokenize{#6}\relax\else\label{#6}\fi%
    \end{subfigure}%
}

\newcommand{\FigureContentSingle}[3][tbp]{%
  \begin{figure}[#1]%
    \centering%
    \vspace*{5pt}%
    #2%
    \caption{#3}%
  \end{figure}%
}
\newcommand{\FigureContentWide}[3][tbp]{%
  \begin{figure*}[#1]%
    \centering%
    \vspace*{5pt}%
    #2%
    \caption{#3}%
  \end{figure*}%
}
\makeatletter
\AtBeginDocument{\DeclareMathVersion{bold}
\SetSymbolFont{operators}{bold}{T1}{times}{b}{n}
\SetSymbolFont{NewLetters}{bold}{T1}{times}{b}{it}
\SetMathAlphabet{\mathrm}{bold}{T1}{times}{b}{n}
\SetMathAlphabet{\mathit}{bold}{T1}{times}{b}{it}
\SetMathAlphabet{\mathbf}{bold}{T1}{times}{b}{n}
\SetMathAlphabet{\mathtt}{bold}{OT1}{pcr}{b}{n}
\SetSymbolFont{symbols}{bold}{OMS}{cmsy}{b}{n}
\renewcommand\boldmath{\@nomath\boldmath\mathversion{bold}}}
\makeatother

\def\BibTeX{{\rm B\kern-.05em{\sc i\kern-.025em b}\kern-.08em
    T\kern-.1667em\lower.7ex\hbox{E}\kern-.125emX}}

\begin{document}

\history{Date of publication 2026, date of current version 2026.}
\doi{10.1109/ACCESS.2026.3752444}

\title{SunnyParking: Multi-Shot Trajectory Generation and Motion State Awareness for Human-like Parking}

\author{
\uppercase{Jishu Miao}\authorrefmark{1},
\uppercase{Han Chen}\authorrefmark{2},
\uppercase{Jiankun Zhai}\authorrefmark{3},
\uppercase{Qi Liu}\authorrefmark{4},
\uppercase{Tsubasa Hirakawa}\authorrefmark{1}, \IEEEmembership{Member, IEEE},
\uppercase{Takayoshi Yamashita}\authorrefmark{1}, \IEEEmembership{Member, IEEE}, AND
\uppercase{Hironobu Fujiyoshi}\authorrefmark{5}, \IEEEmembership{Member, IEEE}
}

\address[1]{Department of Computer Science, Chubu University, Kasugai, Aichi 487-8501, Japan}
\address[2]{College of Automotive Engineering, Tongji University, Shanghai 201804, China}
\address[3]{School of Mechanical Engineering and Automation, Harbin Institute of Technology, Harbin 150001, China}
\address[4]{Independent Researcher}
\address[5]{Department of Robotics, Chubu University, Kasugai, Aichi 487-8501, Japan}

\tfootnote{This article has been accepted for publication in IEEE Access. DOI: 10.1109/ACCESS.2026.3752444.}

\markboth
{Miao \headeretal: SunnyParking: Multi-Shot Trajectory Generation and Motion State Awareness for Human-like Parking}
{Miao \headeretal: SunnyParking: Multi-Shot Trajectory Generation and Motion State Awareness for Human-like Parking}

\corresp{Corresponding author: Jishu Miao (e-mail: miaoj1229@mprg.cs.chubu.ac.jp).}

\begin{abstract}
Automated parking fundamentally differs from on-road driving due to its frequent direction changes and complex maneuvering requirements.
However, existing End-to-End (E2E) planning methods often simplify the parking task into a geometric path regression problem, neglecting explicit modeling of the vehicle's kinematic state. 
This ``dimensionality deficiency'' easily leads to physically infeasible trajectories that deviate from real human driving behavior, particularly at critical gear-shift points in multi-shot parking scenarios.
In this paper, we propose SunnyParking, a novel dual-branch E2E architecture that achieves motion state awareness by jointly predicting spatial trajectories and discrete motion state sequences (e.g., forward/reverse).
Additionally, we introduce a Fourier feature-based representation of target parking slots to overcome the resolution limitations of traditional bird's-eye view (BEV) approaches, enabling high-precision target interactions. 
Experimental results demonstrate that our framework generates more robust and human-like trajectories in complex multi-shot parking scenarios, while substantially improving gear-shift point position accuracy compared to state-of-the-art methods. 
We open-source a new parking dataset based on the CARLA simulator, specifically designed to evaluate full prediction capabilities under complex maneuvers. 
\end{abstract}


\titlepgskip=-21pt

\maketitle

\section{Introduction}
\label{sec:introduction}
\PARstart{I}{n} the field of autonomous driving, 
driving and parking functions complement each other to form a complete, advanced user experience. 
Since its market launch, Automated Parking Assist (APA) has undergone continuous technological evolution, 
gradually upgrading into a fundamental function that supports high-level features such as Valet Parking Assist (VPA) and Automated Valet Parking (AVP). 
Its performance dictates the final quality of the parking task.

Unlike on-road driving, which is predominantly forward-moving, the parking task is fundamentally different. 
It typically involves multiple gear shifts (i.e., forward-reverse maneuvers), requires larger front-wheel steering angles, and operates in closer proximity to surrounding obstacles, leading to a higher risk of collision. 
Concurrently, the vehicle experiences more frequent acceleration and deceleration. 

\Figure[tbp][width=0.99\linewidth]{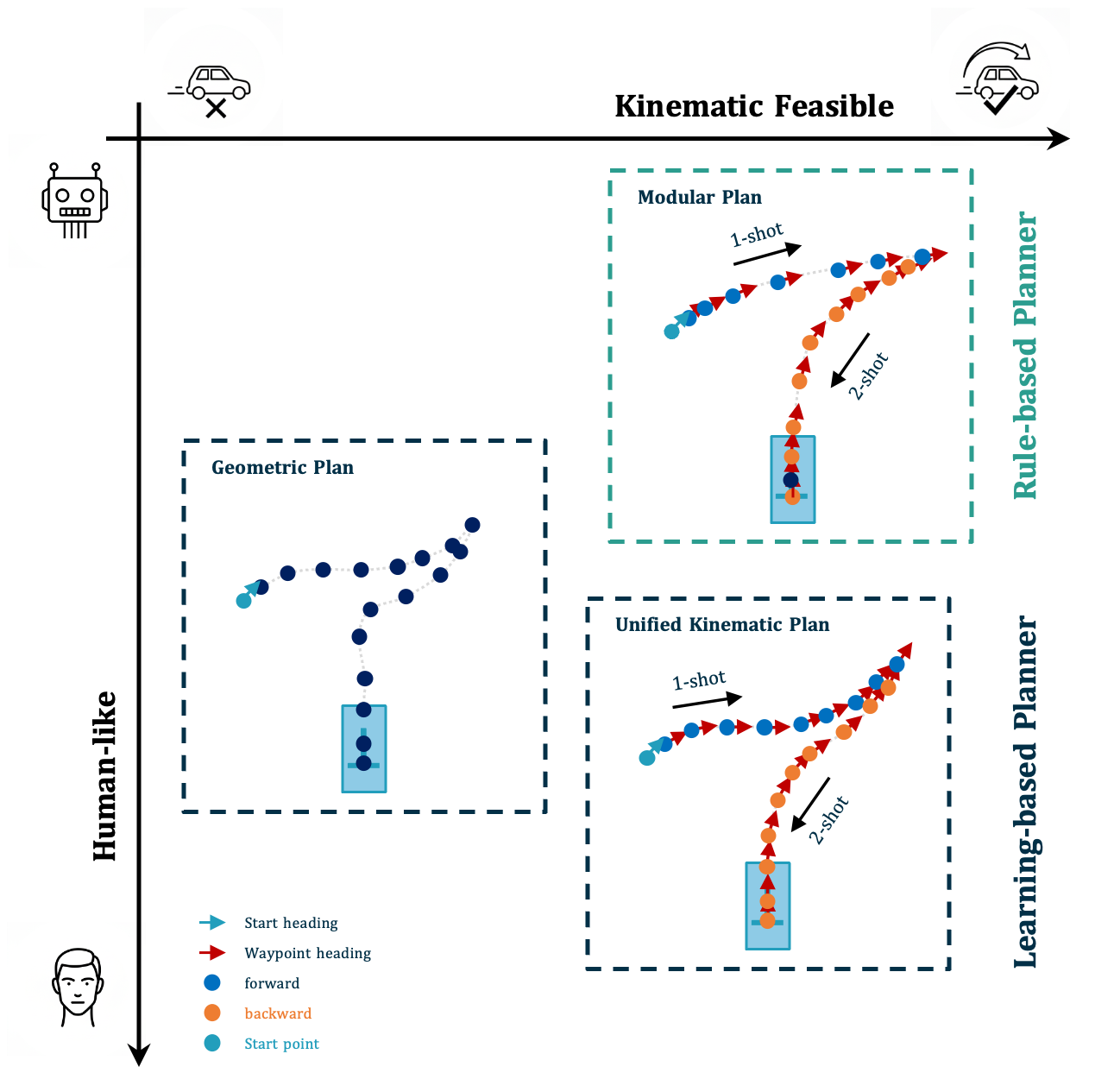}{\textbf{Comparison of parking planning paradigms.
Geometric Plan represents many current E2E approaches by focusing primarily on geometric features. 
Modular Plan represents the traditional rule-based paradigm (e.g., Hybrid A*). 
While it generates a kinematically feasible plan, achieving human-like planning remains a challenge. 
Our Unified Kinematic Plan explicitly models kinematic states while more closely matching expert human trajectories.} \label{fig:compare_planner}}

Although traditional planning methods like Hybrid A* \cite{dolgov2010path} still dominate mass production solutions, 
their practical application is often hindered by a heavy reliance on complex calibration and their inherent difficulty in achieving human-like driving behavior. 
Modern End-to-End (E2E) learning-based approaches have become a research hotspot, showing great promise in handling complex sensor fusion, temporal prediction, and continuous action spaces \cite{10205112, 10377337, jiang2026vadv2}. 
Moving beyond classical regression, generative models, diffusion policies, and foundation agents have rapidly reshaped on-road planning~\cite{jiang2023motiondiffuser,liao2025diffusiondrive, 10.1007/978-3-031-73195-2_4, sima_drivelm_2025}.
Recently, this E2E paradigm has been rapidly extended to the automated parking domain, driving innovations from perception-control integration to multimodal trajectory generation \cite{10588551, 10801763, chen2025endtoendvisualautonomousparking, du2025transparking}.

However, trajectory planning studies have primarily focused on improving the spatial accuracy of the predicted trajectory, with the core objective being the output of precise 2D coordinate waypoints. 
When this paradigm is directly applied to the complex, spatially-constrained environment of automated parking, its limitations are exposed. The parking task is not merely a simple path-finding problem; it is fundamentally a complex maneuvering problem. 
An executable and safe parking trajectory must not only provide precise $(x,y)$ positions, but also satisfy two other critical constraints: 
an accurate continuous heading $\theta$ and the correct corresponding motion state.  

The gap in current E2E parking models is twofold. First, they rely on discretized grid representations that bottleneck the precision of spatial priors. Second, they lack explicit modeling of the vehicle's kinematic states. 
As shown in Fig.~\ref{fig:compare_planner}, the ``Geometric Plan'' approach involves training a model to fit a geometric curve but does not understand the intrinsic physical commands required to drive the vehicle along this curve. 
This leads to a severe deficiency in its ability to predict practically feasible trajectories. It often generates unclear or physically infeasible maneuvers during critical direction changes, primarily because it assumes a continuous flow rather than distinctly identifying the necessary forward and reverse driving patterns.

Compared to the abundance of driving datasets (e.g., NuPlan \cite{caesar2021nuplan}, Waymo Open Dataset \cite{9709630}, Argoverse \cite{Argoverse, Argoverse2}), 
dedicated parking datasets are limited, with the majority focusing on bird's-eye view (BEV) images or perception labels (e.g., CNRPark+EXT \cite{amato2017deep}). 

To fill the aforementioned research gap, we aim to elevate parking planning from a mere geometric path regression problem to a comprehensive prediction problem of both high-fidelity trajectories and kinematic states. 
To overcome the spatial quantization errors of grid-based representations, 
we introduce a Fourier target encoding mechanism that maps slot coordinates into a continuous frequency space, 
ensuring that the planner is conditioned on continuous, non-quantized spatial priors. 
Simultaneously, 
we propose explicit motion state modeling that decodes forward and reverse primitives to improve physical feasibility at gear-shift boundaries. 
Compared to the state-of-the-art approaches, our method not only enhances overall trajectory performance but also maintains robustness even in complex scenarios requiring multiple gear shifts. 

To summarize, our contributions are as follows:

\begin{itemize}
  \item We propose SunnyParking, a novel dual-branch architecture for the joint prediction of spatial trajectory and motion state. Our architecture can simultaneously output a high-precision spatial trajectory and a strictly aligned sequence of discrete motion states.  

  \item We design a cross-domain target fusion mechanism that represents the target slot using Fourier feature mapping. This approach is not constrained by the fixed BEV resolution or performance degradation at the grid's edges, overcoming a key limitation of traditional BEV-based schemes. 

  \item We construct and open-source a CARLA \cite{pmlr-v78-dosovitskiy17a} dataset containing multi-shot parking maneuvers. The dataset features parking demonstrations by expert drivers, supplying all required RGB camera and chassis data for E2E approaches. 
\end{itemize}

\section{Related Works}
Focusing on automated parking, we structure this section around three core parts: target representation, trajectory planner paradigms, and recent developments in generative and foundation models that inform our design choices.

\subsection{Target Representation in Goal-Conditioned Planning}
As a quintessential goal-conditioned planning task, the representation of the target parking slot critically influences performance. 

\subsubsection{Goal-state Representation} The most straightforward approach is to represent the goal as a sparse vector of its state. 
MPNet's \cite{9154607} Pnet explicitly uses the goal state $(x, y, \theta)$ as an input for a bidirectional path generation heuristic. 
It simultaneously extends a path from the start towards the goal and a second path from the goal towards the start, driving the two trajectories to converge until they can be connected. 

\subsubsection{BEV Pseudo-image Representation} To provide richer spatial priors, 
many studies have drawn inspiration from concepts in BEV perception \cite{philion2020lift, huang2021bevdet, liu2022bevfusion, 10377285, xia2024henet}. 

E2E Parking \cite{10588551} represents the target slot as a binary pseudo-image within the vehicle-centric BEV space. 
This is achieved by first transforming the target slot's coordinates and then setting the hard target heatmap that forms a fixed-size square at that location. 
TransParking \cite{du2025transparking} further refines this approach with a Gaussian kernel distribution representation. 
This method generates a soft target heatmap. 
This direct alignment allows the target features to be fused with the sensor features via element-wise addition (i.e., one-domain fusion). 

However, the precision of these heatmap-based methods is fundamentally bottlenecked by spatial quantization errors inherent to the BEV grid. 
Furthermore, downstream spatial decoders struggle to extract the precise, continuous heading of the target slot from these discretized representations. 

\subsubsection{Fourier Feature Representation}
\label{subsec:fourier_background}
Tancik et al.~\cite{tancik2020fourier} show that mapping low-dimensional coordinates into a
higher-dimensional sinusoidal feature space helps MLPs learn high-frequency details,
overcoming the ``spectral bias'' of coordinate-based networks~\cite{pmlr-v97-rahaman19a}.
This motivates replacing BEV heatmaps with a continuous target encoding.
Our concrete axis-aligned instantiation, defined in Sec.~\ref{subsec:position_encoding},
uses deterministic multi-band sinusoids rather than random Fourier features~\cite{tancik2020fourier},
which is better suited to the axis-aligned vehicle coordinate system (VCS) in parking. 

\subsection{End-to-End Planner Paradigms}
The E2E approach learns a direct mapping from sensor inputs to maneuvers. 
DriveAdapter~\cite{10378531} has explored new architectures to optimize this perception-to-planning pipeline. 
Focusing on automated parking, current E2E methods typically fall into two main categories based on their final outputs: 
direct control and trajectory prediction.

\subsubsection{Control-based planning paradigm} 
Pioneering works \cite{DBLP:journals/corr/BojarskiTDFFGJM16, 10588551} bypass the need for a separate controller by outputting steering and throttle commands directly. 
However, this reactive approach is unsuitable for the deliberate, 
multi-step nature of parking, as its short planning horizon amplifies control errors, often leading to hesitation or deadlocks. 

\subsubsection{Trajectory-based planning paradigm} 
Chitta et al. \cite{chitta2022transfuser} employ a recurrent architecture for driving trajectory generation. 
The fused spatial context vector subsequently initializes the hidden state of a GRU decoder \cite{cho-etal-2014-properties}, which then predicts the future trajectory as an ordered sequence of waypoints in an autoregressive manner. 
UniAD \cite{10205112}, which leverages Transformer-based backbones, has demonstrated impressive performance with a simple attention-based planner in generating smooth, collision-free paths. 

For the parking task, ParkingE2E \cite{10801763} also generates waypoints in an autoregressive manner, employing a Transformer decoder \cite{10.5555/3295222.3295349} to predict the $(x, y)$ coordinates of waypoints for $N$ timesteps starting from the current time $t_0$, at a fixed time interval. 
This study was validated on a real vehicle, and the experimental results demonstrated its ability to generate a solid, 1-shot parking trajectory in real-world scenarios. 
CAA-Policy \cite{chen2025endtoendvisualautonomousparking} jointly predicts control signals and waypoints with a dual-branch architecture. 
Its short-horizon waypoint prediction is introduced as an auxiliary task to improve control accuracy and temporal consistency. 
Nevertheless, waypoint prediction remains subordinate to control and is confined to a short horizon, 
which limits full maneuver modeling and makes it difficult to capture the uncertainty inherent in complex parking scenarios. 

These studies are primarily based on the imitation learning (IL) paradigm. 
However, models relying purely on IL suffer from two main inherent defects: limited generalization and distributional shift (i.e., covariate shift \cite{ross2011reduction}). 
The compounding errors induced by these covariate shifts \cite{foster2024behavior} are particularly catastrophic at critical decision points such as gear transitions. 
To mitigate this, a prevailing trend in complex driving tasks is to employ reinforcement learning (RL) to fine-tune an IL cold start \cite{10938794, xiong2026recogdrive}.

While RL is a valid countermeasure, it inevitably introduces significant training instability and computational overhead.  
A recent study has demonstrated that these compounding shifts can be intrinsically mitigated without RL by embedding strong architectural inductive biases and auxiliary state constraints into the network design \cite{cheng2024pluto}.

\subsection{Generative, World-Model, and Foundation-Model Planning}
\label{sec:related_generative}
Beyond the regression-based E2E planners above, recent work also treats trajectory generation as a generative or forecasting problem.
Recent studies~\cite{jiang2023motiondiffuser, liao2025diffusiondrive} use diffusion models to sample multimodal on-road trajectories, and ParkDiffusion~\cite{wei2025parkdiffusion} brings the same idea to automated parking. 
World models such as DriveDreamer~\cite{10.1007/978-3-031-73195-2_4} go further by predicting future scene videos or occupancy together with ego motion~\cite{zheng2024occworld, yang2025driving}; 
Feng et al.~\cite{Feng2025ASO} survey this line of research.
These methods are attractive for capturing multimodal futures and dynamic traffic.

A related trend is foundation-model and vision--language--action (VLA) driving.
DriveLM~\cite{sima_drivelm_2025} adds graph-structured visual question answering for multi-step reasoning, 
while FlashDrive~\cite{li2026flashdrive} and RecogDrive~\cite{xiong2026recogdrive} explore faster VLA inference and RL-based fine-tuning.
Such systems improve general on-road competence, 
but they are tailored to the driving task and high-level reasoning.
They do not address continuous slot encoding or explicit gear-shift motion states.

\section{Methodology} 
\Figure[htbp][width=1\textwidth]{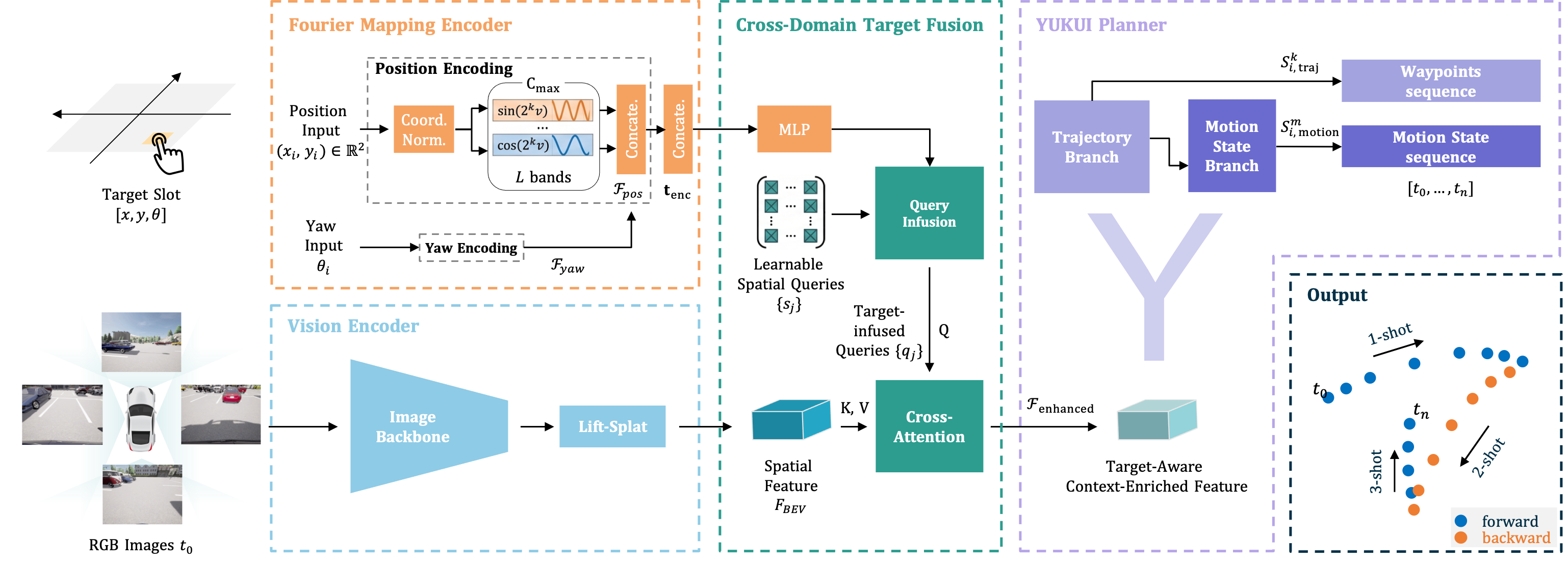}{\textbf{Overview of SunnyParking's Network Architecture. It takes four RGB images and the selected target slot as input. 
The Vision Encoder extracts the vision-domain features and projects them to the BEV space as the spatial features $\mathcal{F}_{\mathrm{BEV}}$. 
The Fourier Mapping Encoder first represents the target including its $(x, y, \theta)$ as Fourier features. 
The Cross-domain Target Fusion module infuses the encoded target into learnable spatial queries and attends to $\mathcal{F}_{\mathrm{BEV}}$, yielding the target-aware context-enriched features $\mathcal{F}_{\mathrm{enhanced}}$. 
In the YUKUI module, the Trajectory Branch takes $\mathcal{F}_{\mathrm{enhanced}}$ and then predicts a set of waypoints. With the additional input from the Trajectory Branch's hidden layer, the Motion State Branch predicts the motion state corresponding to each waypoint.}\label{fig:network_overview}}

We introduce SunnyParking, named after the sunny midday conditions under which our dataset was collected.
In this section, we introduce the problem definition, the architecture of each module, and the design of the loss functions.

\subsection{Preliminaries: Problem Definition} 
We formulate the task of imitating expert driving trajectories as a supervised learning problem. 
The core idea is to train a neural network to map sensor inputs to future vehicle trajectories based on expert behaviors.  

Discretizing continuous trajectories into tokens is a well-established paradigm in trajectory transformers outside the parking domain. 
Existing works in continuous control~\cite{janner2021offline, chen2021decision} and on-road motion forecasting~\cite{10377717} demonstrate that serializing continuous states, actions, 
or spatial coordinates into sequential tokens enables standard Transformers to generate trajectories autoregressively. 
Extending this tokenization scheme, we jointly tokenize continuous spatial waypoints alongside their discrete motion states, 
formulating multi-shot parking as a unified sequence prediction problem.

Our dataset consists of $M$ expert-driven trajectories collected from multiple drivers. 
Path shape and gear-shift timing can differ across samples even when the scene and target slot are similar. 
Each trajectory $i$ is a time-ordered sequence of states of length $N_i$. For any time step $j$ within trajectory $i$, the available data include
\begin{equation}
\mathcal{D} = \{(I_{i,j}^k, T_{i,j}, M_{i,j}, S_i)\},
\label{eq:data_definition}
\end{equation}
where $i \in [1, M]$ indexes trajectories, 
$j \in [1, N_i]$ indexes waypoints, 
$k \in [1, R]$ indexes cameras, 
$I$ is an RGB image, 
$T$ is a trajectory waypoint, 
$M$ is a motion state, 
and $S$ is the target-slot pose.

From the raw dataset, we reorganize the dataset $\mathcal{D}'$. 
Each sample in $\mathcal{D}'$ consists of an input pair $(I_{i,j}^k, S_i)$ and a corresponding target output pair $(T'_{i,j}, M'_{i,j})$. The targets represent the ground-truth future sequences of length $Q$:
\begin{equation}
\mathcal{D}' = \{(I_{i,j}^k, T'_{i,j}, M'_{i,j}, S_i)\}. 
\end{equation} 

The expert future waypoint and motion state sequences of length $Q$ are
\begin{equation}
    T'_{i,j} = \{P_{i, \min(j+b, N_i)}\}_{b=1}^Q, 
\end{equation}
\begin{equation}
    M'_{i,j} = \{M_{i, \min(j+b, N_i)}\}_{b=1}^Q. 
\end{equation}
To obtain motion state labels, we combine the gearbox state with longitudinal speed $v$, using a speed threshold $\tau$. 
When $|v| > \tau$, the gearbox sets the direction of that moving segment. 
When $|v| \le \tau$, the frame is a dwell: it keeps the motion state of the preceding moving segment and does not become a third class. 
Neutral-gear transients are removed in preprocessing and never form a segment, leaving a two-class sequence $\{\text{forward}, \text{reverse}\}$. 
The proposed method uses $\tau = 0.05\,\mathrm{m/s}$. 
This choice follows the dead zone of a practical speed sensor and is analyzed in Appendix~\ref{appendix:motion_state_gt}. 

The network $N_{\theta}$ takes the current images and goal as input and outputs predictions for the future trajectory and motion states, structured to match the ground-truth targets: 
\begin{equation}
        N_{\theta}(I, S) = (\hat{T}', \hat{M}'), 
\end{equation} 
where $\hat{T}'$ is the predicted waypoint sequence and $\hat{M}'$ is the predicted motion state sequence. 

The network's parameters $\theta$ are optimized by minimizing a composite loss function, $\mathcal{L}_{\text{total}}$, which measures the discrepancy between the predictions $(\hat{T}', \hat{M}')$ and the ground-truth targets $(T', M')$: 
\begin{equation}
    \theta' = \arg\min_{\theta} \mathbb{E}_{(I, T', M', S) \sim \mathcal{D}'} [\mathcal{L}((T', M'), N_{\theta}(I, S))], 
\end{equation}
where $\mathcal{L}$ denotes the loss function. 

\subsection{Fourier Mapping Encoder} 

The key information of the target slot $S_i$ is its position $\mathbf{p}_i=(x_i, y_i) \in \mathbb{R}^2$ and heading $\theta_i$. 
These define the goal position and orientation of the vehicle when it completes the parking task. 

In prior work, the target slot is rasterized into a BEV grid of fixed resolution (typically $0.1$--$0.2$~m per grid). 
This inherently limits positional precision and introduces quantization artifacts near slot boundaries. 
As shown in our attention analysis (Fig.~\ref{fig:heatmap_viz}), BEV-based encoders often fail to maintain consistent localization as the slot moves across pixel boundaries, leading to periodic distortions in the planner's spatial reasoning. 
Motivated by this limitation, we adopt a continuous Fourier feature mapping that provides sub-centimeter-scale spatial frequency components without the quantization imposed by the BEV grid. 
This representation allows the target slot to condition the planner with high-frequency location cues, enabling more reliable attention modulation during multi-shot maneuvers. 
As shown in Fig.~\ref{fig:network_overview}, the Fourier Mapping Encoder maps both the target slot's position and heading to the Fourier-encoded target $\mathbf{t}_{\mathrm{enc}}$. 

\subsubsection{Position Encoding}
\label{subsec:position_encoding}

We encode each scalar coordinate with deterministic, axis-aligned frequency bands.
The position $\mathbf{p}_i=(x_i,y_i)$ is first normalized to $[-\pi,\pi]$:
\begin{equation}
    x_i' = \frac{\pi}{C_{\max}} x_i, \quad y_i' = \frac{\pi}{C_{\max}} y_i,
    \label{eq:coord_normalize}
\end{equation}
where $C_{\max}$ is the half-extent of the ROI in meters (coordinates lie in $[-C_{\max}, C_{\max}]$).
For a normalized scalar $v \in [-\pi,\pi]$, the axis-aligned Fourier mapping over $L$ frequency bands is
\begin{equation}
    \label{eq:gamma_impl}
    \gamma(v) = \bigl[ \sin(2^k v)\bigr]_{k=0}^{L-1} \Vert \bigl[\cos(2^k v) \bigr]_{k=0}^{L-1}.
\end{equation}
The position embedding concatenates the normalized coordinates with their Fourier features:
\begin{equation}
    \label{eq:final_position_encoding}
    \mathcal{F}_{\mathrm{pos}} = \bigl[ x_i', y_i', \gamma(x_i'), \gamma(y_i') \bigr].
\end{equation}

Each band $k$ has phase $\phi_k(v)=2^k v$, so one period in normalized space spans $\Delta v_k = 2\pi / 2^k$.
Mapping back to physical coordinates via $v = (\pi / C_{\max})\, x$ yields the wavelength
\begin{equation}
    \label{eq:wavelength}
    \lambda_k = \frac{C_{\max}}{\pi}\,\Delta v_k
            = \frac{2 C_{\max}}{2^k}.
\end{equation}
The lowest band ($k=0$) therefore covers the full ROI extent $2C_{\max}$, 
while the highest band ($k=L{-}1$) provides the finest resolution.
With $C_{\max}=10$\,m and $L=12$, the finest wavelength is $\lambda_{11} = 20/2^{11} \approx 9.8$\,mm, 
more than an order of magnitude finer than our BEV grid resolution of $0.1$\,m per pixel ($200\times200$ over a $20\times20$\,m area).

\subsubsection{Yaw Encoding} 
To handle the heading angle $\theta_i$ of the target slot, the heading is mapped to a 2D vector $\mathcal{F}_{\mathrm{yaw}} = [\cos(\theta_i), \sin(\theta_i)]$. 
For the final output of the target encoder, the Fourier-encoded target $\mathbf{t}_{\mathrm{enc}}$ is obtained by concatenating positional and yaw features: 
$\mathbf{t}_{\mathrm{enc}} = [\mathcal{F}_{\mathrm{pos}}, \mathcal{F}_{\mathrm{yaw}}]$. 

\subsection{Cross-domain Target Fusion} 
To leverage the precise position and heading information provided by the Fourier-encoded target $\mathbf{t}_{\mathrm{enc}}$, it is fused with the BEV feature map $\mathcal{F}_{\mathrm{BEV}}$. 
Unlike conventional methods that fuse two BEV-based feature maps directly, we design a Cross-domain Target Fusion module. 
It uses the Fourier-encoded target to modulate a set of learnable spatial queries, and then uses these target-aware queries to extract the relevant information from the BEV feature map. 
These queries are dynamically constructed from two sources of information. 

\subsubsection{Target projection} An MLP projects the low-dimensional $\mathbf{t}_{\mathrm{enc}}$ to a $C$-dimensional vector $\mathbf{t}_{\mathrm{global}}$, matching the channel width of the BEV feature map. 

\subsubsection{Query Infusion} We introduce a set of learnable spatial query tokens $\{\mathbf{s}_{j}\}_{j=1}^N$, where $N=H \times W$ matches the shape of the BEV feature map. 
This set learns the prior information for each of the $N$ locations. 
We adopt gated fusion to infuse $\mathbf{t}_{\mathrm{global}}$ into each spatial query $\mathbf{s}_j$. 
Let $\boldsymbol{\phi}_j = [\mathbf{s}_j;\mathbf{t}_{\mathrm{global}}]$ and $\mathbf{g}_j = \sigma(\mathbf{W}_g \boldsymbol{\phi}_j + \mathbf{b}_g) \in (0,1)^{C}$.
\begin{equation} 
    \mathbf{q}_j = (1-\mathbf{g}_j)\odot(\mathbf{s}_j + \mathbf{t}_{\mathrm{global}}) + \mathbf{g}_j \odot (\mathbf{W}_c \boldsymbol{\phi}_j + \mathbf{b}_c), 
\end{equation} 
where $\odot$ denotes element-wise multiplication. 
Thus, $\mathbf{q}_j$ retains the location-specific prior in $\mathbf{s}_j$ while incorporating shared target context, with per-location, per-channel control via $\mathbf{g}_j$.

\subsubsection{Target-Spatial Information Fusion} 
After flattening the BEV feature map $\mathcal{F}_{\mathrm{BEV}}$ and adding a positional embedding, 
we use it as Key and Value and the target-infused queries $\mathbf{q}_j$ as Query in a Transformer decoder, 
which outputs the target-aware context-enriched features $\mathcal{F}_{\mathrm{enhanced}}$ for their corresponding spatial locations.

\subsection{YUKUI Planner} 
The philosophy of our neural planner is to predict the trajectory while explicitly considering the vehicle motion state at each waypoint. 
As shown in Fig.~\ref{fig:network_overview}, 
the combination of the trajectory branch and motion state branch resembles a \textbf{Y}-shaped \textbf{U}nified \textbf{K}inematics for m\textbf{U}lti-shot \textbf{I}mitation planner (\textbf{YUKUI}). 
Multi-shot parking fundamentally entails alternating forward and reverse segments with discrete gear shifts. 
To model this coupled behavior, both branches take $\mathcal{F}_{\mathrm{enhanced}}$ as their shared input feature map. 
The trajectory branch serves as the primary decoder, mapping $\mathcal{F}_{\mathrm{enhanced}}$ to a waypoint sequence. 
To prevent kinematic contradictions, the motion state branch conditions on trajectory features rather than operating independently. 
This conditioned design ensures strict one-to-one temporal alignment between waypoints and motion primitives in the joint output $(\hat{T}', \hat{M}')$, imitating the tightly coupled trajectory--kinematics structure exhibited in expert maneuvers. 

\subsubsection{Goal Serialization} 
The goal of YUKUI is to determine the trajectory and the motion state. 
By serializing the goal (i.e., waypoints and motion states), the prediction task is converted into a token prediction problem. 

The serialization method for trajectories is defined as follows: 
\begin{equation} 
	\text{Ser}(P_{i,j}^k) = \left\lfloor \frac{|P_{i,j}^k + R_k|}{2R_k} \right\rfloor \times N_u,
\label{eq:serialization} 
\end{equation} 
where $k \in \{x, y, \cos\theta, \sin\theta\}$, $\text{Ser}(\cdot)$ serializes each waypoint component, $N_u$ is the maximum trajectory-token value, and $R_k$ is the preset normalization bound for component $k$. 

We serialize only forward and reverse motion tokens.
A stationary class is not used at any stage: it is absent from the label vocabulary, the decoder output, and the motion state losses in Sec.~\ref{subsec:loss_function}.
The motion branch therefore predicts a binary sequence that stays one-to-one aligned with $T'_{i,j}$. 

The serialization method for the motion state is defined as follows:
\begin{equation}
    \text{Ser}(M_{i,j}^m) = \left\lfloor M_{i,j}^m \cdot N_v \right\rfloor,
    \quad m \in \{\text{fwd}, \text{rev}\}.
\end{equation}
Here, $N_v$ is the maximum motion-token index, $\text{Ser}(\cdot)$ serializes each component, 
and $M_{i,j}^{\text{fwd}}, M_{i,j}^{\text{rev}} \in \{0,1\}$ indicate forward and reverse, respectively.

Both sequences share the same temporal length of $N_i$ steps and maintain a one-to-one correspondence at each time step. 
The complete sequence formulations are as follows: 
\begin{equation} 
\begin{aligned} 
    S_{i, \text{traj}}^{k} &= \left[\mathrm{BOS}, \text{Ser}(P_{i,1}^k), \dots, \text{Ser}(P_{i,N_i}^k), \mathrm{EOS} \right], \\
    S_{i, \text{motion}}^{m} & = \left[\mathrm{BOS}, \text{Ser}(M_{i,1}^m), \dots, \text{Ser}(M_{i,N_i}^m), \mathrm{EOS} \right].
\end{aligned} 
\end{equation} 

\subsubsection{Trajectory Branch} 
We introduce a Transformer decoder for predicting waypoints. 
The target-aware features $\mathcal{F}_{\mathrm{enhanced}}$ serve as the Key and Value, while a set of $N$ learnable queries serve as the Query input.

\subsubsection{Motion State Branch} 
To predict the vehicle motion state (Fig.~\ref{fig:mot_arch}), we use a two-stage query-fusion mechanism. 
First, the Motion Query embeddings are refined by attending to the hidden layer features from the trajectory branch, 
making them contextually aware of the planned path. 
Then, a lighter Transformer decoder uses the target-aware spatial features $\mathcal{F}_{\mathrm{enhanced}}$ as Key and Value, 
allowing the trajectory-conditioned queries to extract spatial and contextual information from it.

\Figure[tbp][width=0.99\linewidth]{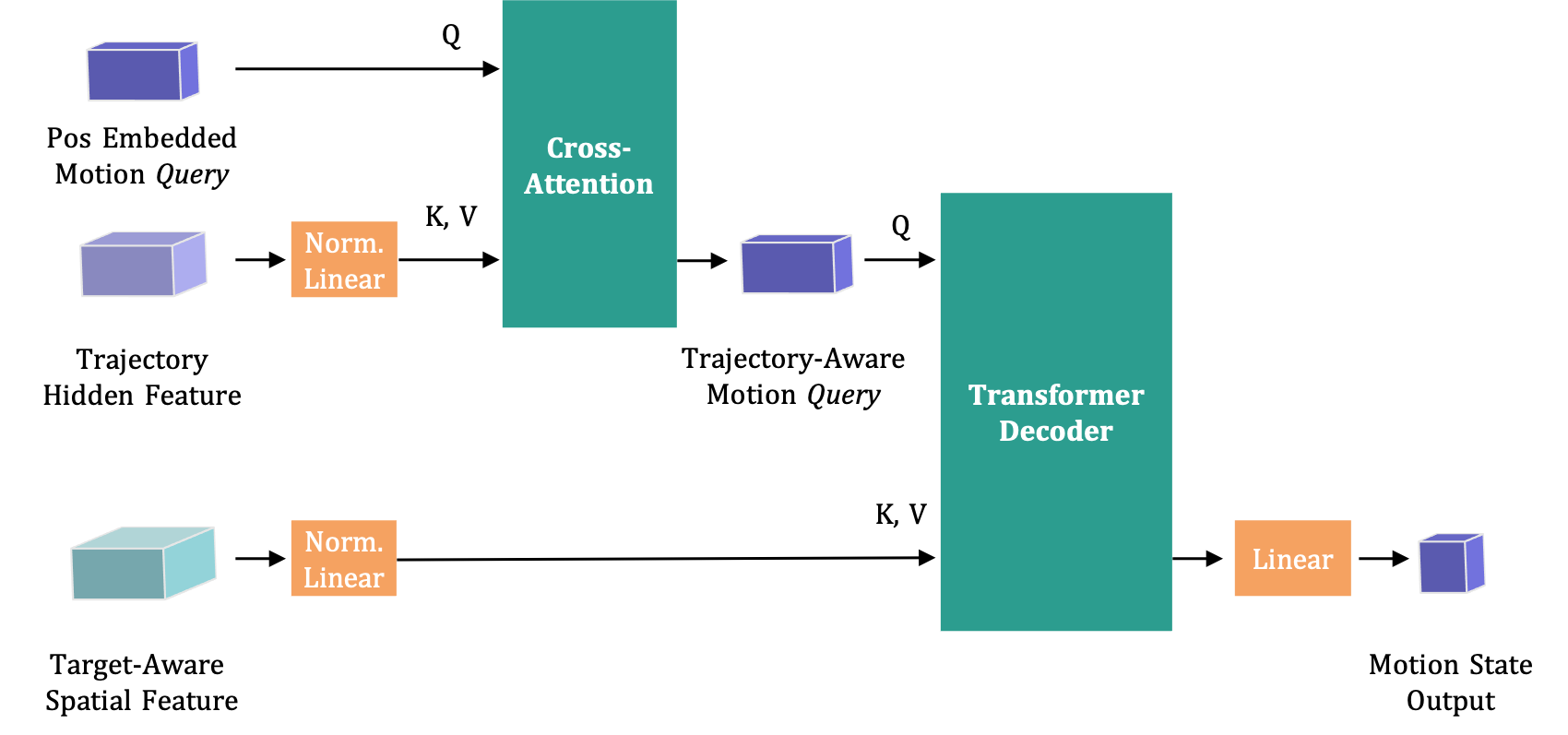}{\textbf{Architecture of the motion state branch.} Motion queries are enriched by trajectory features via cross-attention, then fed into a Transformer decoder.\label{fig:mot_arch}}

\subsection{Loss Function}\label{subsec:loss_function}
The total loss is a weighted sum of two component losses: 
\begin{equation}
    \mathcal{L}_{\text{total}} = \lambda_{\text{waypoints}} \mathcal{L}_{\text{waypoints}} + \lambda_{\text{motion}} \mathcal{L}_{\text{motion}}, 
\end{equation}

where $\lambda_{\text{waypoints}}$ and $\lambda_{\text{motion}}$ are weighting coefficients that balance the contributions from the trajectory loss $\mathcal{L}_{\text{waypoints}}$ and the motion state loss $\mathcal{L}_{\text{motion}}$. 
We use fixed weights of $\lambda_{\text{waypoints}}=1.0$ and $\lambda_{\text{motion}}=0.2$ to stably optimize both branches.

\subsubsection{Waypoint Loss}
The trajectory branch is trained with a composite waypoint loss:
\begin{equation}
    \mathcal{L}_{\text{waypoints}} = \mathcal{L}_{\text{MSE}} + \lambda_{\text{FDE}} \mathcal{L}_{\text{FDE}} + \lambda_{\text{FHE}} \mathcal{L}_{\text{FHE}}.
\label{eq:waypoints_loss}
\end{equation}
Here, $\mathcal{L}_{\text{MSE}}$ is the Mean Squared Error (MSE) between the predicted waypoints and the ground truth (GT) over the full sequence. 
It provides dense supervision along the path, but step-wise point accuracy alone does not always guarantee a precise terminal pose, which is critical for trajectory quality and downstream control. 
We therefore add Final Displacement Error (FDE) loss $\mathcal{L}_{\text{FDE}}$ and Final Heading Error (FHE) loss $\mathcal{L}_{\text{FHE}}$ on the last waypoint index $N{-}1$. 
We set the weights as $\lambda_{\text{FDE}}=0.5$ and $\lambda_{\text{FHE}}=0.5$.

\begin{equation}
    \mathcal{L}_{\text{FDE}} = \frac{1}{B} \sum_{b=1}^{B}
    \left\| \hat{\mathbf{p}}_{b,N-1} - \mathbf{p}^{\mathrm{gt}}_{b,N-1} \right\|_2^2,
\end{equation}
where $\mathbf{p}_{b,t}=(x_{b,t}, y_{b,t})$. 
For the heading, the predicted angle $\hat{\theta}_{b,N-1}$ and ground-truth angle $\theta^{\mathrm{gt}}_{b,N-1}$ are first recovered from their respective last-step $\sin$ and $\cos$ outputs using the $\operatorname{atan2}$ function. 
The FHE is then calculated as the squared wrapped angular error:
\begin{equation}
    \mathcal{L}_{\text{FHE}} = \frac{1}{B} \sum_{b=1}^{B} (\Delta\theta_b)^2,
\end{equation}
where the wrapped angular difference $\Delta\theta_b \in [-\pi, \pi]$ is explicitly computed as:
\begin{equation}
    \Delta\theta_b = \operatorname{atan2}\big(\sin(\hat{\theta}_{b,N-1}-\theta^{\mathrm{gt}}_{b,N-1}), \cos(\hat{\theta}_{b,N-1}-\theta^{\mathrm{gt}}_{b,N-1})\big).
\end{equation}

\subsubsection{Motion Loss}
The motion loss $\mathcal{L}_{\text{motion}}$ is composed of a Cross-Entropy (CE) loss $\mathcal{L}_{\text{CE}}$ for token-wise classification, 
a GT-aware smoothness loss $\mathcal{L}_{\text{smooth}}$, and a transition-aware loss $\mathcal{L}_{\text{trans}}$:
\begin{equation}
    \mathcal{L}_{\text{motion}} = \mathcal{L}_{\text{CE}} + \lambda_{\text{smooth}} \mathcal{L}_{\text{smooth}} + \lambda_{\text{trans}} \mathcal{L}_{\text{trans}}. 
\label{eq:motion_loss}
\end{equation}
We set the smoothness loss weight $\lambda_{\text{smooth}}=0.05$ and the transition-aware loss weight $\lambda_{\text{trans}}=0.2$.

\textbf{GT-Aware Smoothness.} To reduce motion-state jitter in hesitant-prone regions, we design a GT-aware smoothness loss that avoids penalizing the network for predicting correct,
clear gear-shift points. Let $\hat{P}_t$ denote the predicted probability distribution of the motion state pair (forward and reverse) at time step $t$. The smoothness loss is implemented as a masked $L_{1}$ norm of consecutive probability changes:
\begin{equation}
    \mathcal{L}_{\text{smooth}} = \frac{1}{\sum m_t} \sum_{t} m_t \|\hat{P}_{t+1} - \hat{P}_t\|_1,
\end{equation}
where $m_t \in \{0, 1\}$ is a tolerance mask. It deactivates the loss calculation ($m_t=0$) within a tolerance window surrounding the ground-truth gear-shift locations, enforcing smoothness exclusively in stable, non-shifting regions.

\textbf{Transition-Aware Motion Design.} 
Beyond regional smoothness, predicting sparse motion state transitions is critical. 
We introduce $\mathcal{L}_{\text{trans}}$ to explicitly supervise these dynamics. 
Let $\hat{\mu}_t \in [0, 1]^2$ be the expected continuous state derived from the predicted token distribution. We estimate the transition probability $\hat{q}_t \in [0, 1]$ between adjacent steps using their normalized $L_1$ distance:
\begin{equation}
    \hat{q}_t = \frac{1}{2} \|\hat{\mu}_{t+1} - \hat{\mu}_t\|_1.
\end{equation}
Let $y_t \in \{0, 1\}$ indicate a ground-truth transition at step $t$ (i.e., a change in forward/reverse tokens). To address the extreme sparsity of these transition events, we optimize $\hat{q}_t$ using a temporally weighted Focal Loss:
\begin{equation}
    \mathcal{L}_{\text{trans}} = \frac{1}{N} \sum_{t} w_t \cdot \operatorname{FocalLoss}(\hat{q}_t, y_t, \gamma),
\end{equation}
where $\operatorname{FocalLoss}(\cdot)$ denotes the standard binary focal loss with the focusing parameter $\gamma$, and $w_t$ handles additional sample weighting. 

Jointly optimizing $\mathcal{L}_{\text{smooth}}$ and $\mathcal{L}_{\text{trans}}$ ensures the model maintains temporal stability while remaining acutely responsive to intended behavioral shifts.

\section{Experiments}

\Figure[bp][width=0.99\linewidth]{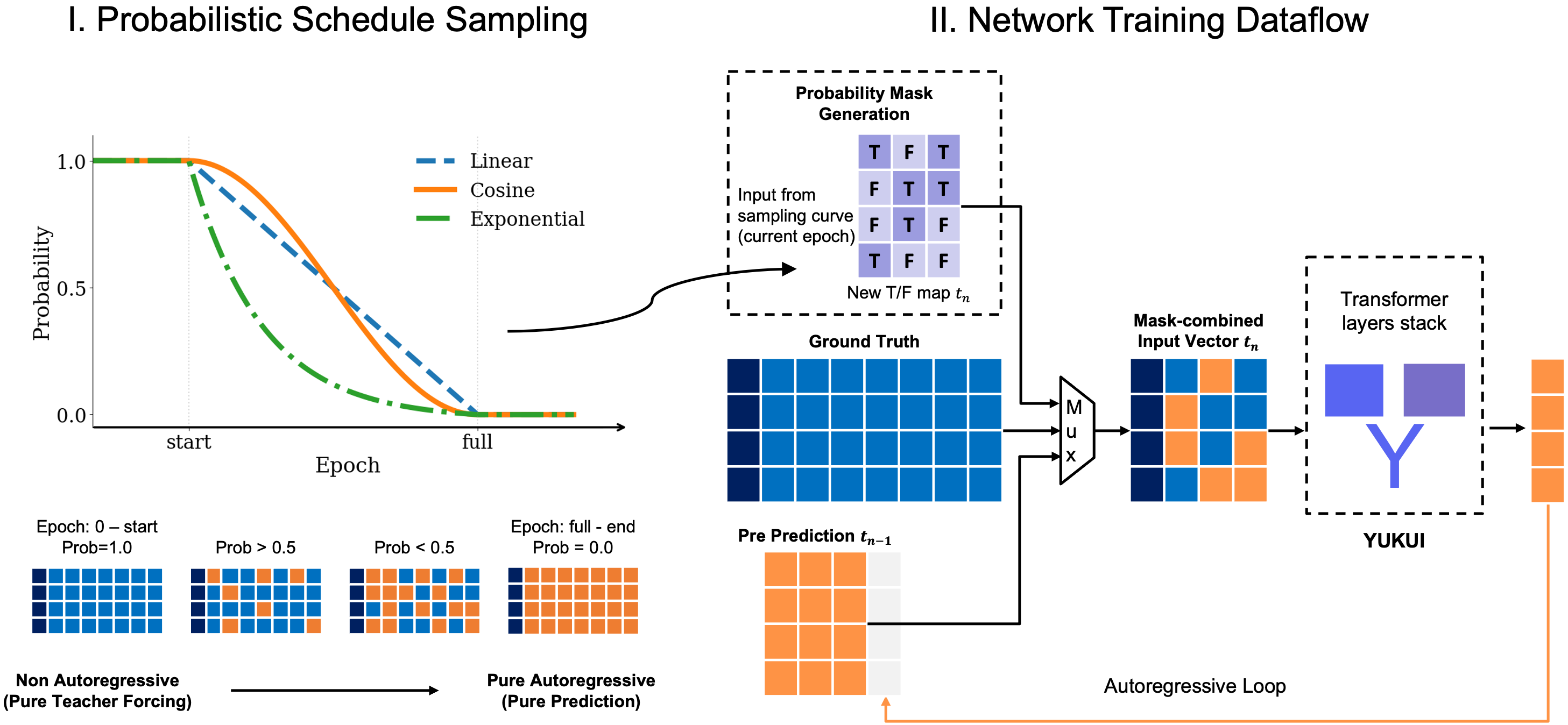}{\textbf{Progressive transition from teacher forcing to autoregressive prediction via Scheduled Sampling.}\label{fig:schedule_sampling}}

\subsection{Training Configuration and Optimization}
Following the data acquisition pipeline of E2E Parking~\cite{10588551}, we collected our parking dataset in the CARLA simulator (detailed in Appendix~\ref{appendix:dataset}). 
All training and validation samples use the CARLA ``Clear Noon'' weather preset.

We train our model for a total of $30$ epochs using a batch size of $16$, distributed across $2$ RTX~4090 GPUs. 
We use the Adam optimizer with an initial learning rate of $2 \times 10^{-4}$. 
A cosine annealing learning rate schedule is employed, which includes warmup and decays the learning rate from $2 \times 10^{-4}$  to $1 \times 10^{-6}$. 
To ensure training stability, gradient clipping is applied with a threshold of $0.5$. 
Mixed precision training (FP16) is enabled to accelerate computation. 
The BEV representation covers $[-10, 10]$\,m along both the $x$ and $y$ axes in the VCS, with a $200 \times 200$ feature map and thus a resolution of $0.1$\,m/grid. 
During training, we add uniform noise of $\pm 0.3$\,m to the target position ($x, y$) and $\pm 2.0^{\circ}$ to the target yaw $\theta$. 
In the trajectory decoder, the maximum value of $N_u$ is 1200; in the motion state decoder, the maximum value of $N_v$ is 100. 
Both decoders generate a sequence of predictions with a length of $30$. 

To address exposure bias, we employ scheduled sampling to bridge teacher forcing (training) and autoregressive inference. 
Scheduled sampling is applied from epoch~$5$ through epoch~$25$. 
During this period, the probability of feeding the ground-truth token to the next time step linearly decays from $1.0$ down to $0.0$. 
After epoch $25$, the model continues training in a fully autoregressive mode. 

All nested sub-loss terms are assigned fixed weighting coefficients and kept unchanged throughout training. 
During the loss calculation, all padding (PAD) tokens are ignored.

\subsection{Evaluation Metrics}
We focus on both the trajectory and motion state prediction performance of YUKUI. 
\subsubsection{Trajectory Evaluation} 
To evaluate the multi-shot parking trajectories, we quantify the error between the predictions and the ground truth using the following metrics. 
To assess the overall spatial accuracy of the trajectory waypoints, we employ Average Displacement Error (ADE), Fourier descriptor difference (Four. Diff.), and Hausdorff distance (Haus. Dist.). 
For waypoint heading performance, we calculate the Average Heading Error (AHE) as the mean absolute angular difference between the predicted and ground-truth poses. 
Because the exact pose of the last waypoint is critical for both the motion optimizer and the control module, we evaluate the trajectory endpoint using the mean Final Displacement Error (mFDE) and the mean Final Heading Error (mFHE), 
i.e., the sample-averaged position and heading errors at the last predicted waypoint.

\subsubsection{Motion State Evaluation}
As we predict the motion state of each waypoint, two vital pieces of information are obtained: 
the general vehicle kinematic state and the trajectory shape change points (i.e., gear-shift positions). 

We classify a parking trajectory according to the number of gear-shift operations observed in the ground-truth demonstration:
\begin{itemize}
\item $1$-shot: $0$ gear shifts
\item $2$-shot: $1$ gear shift
\item $3$-shot: $2$ gear shifts
\item $4$-shot: $3$ gear shifts
\end{itemize}
We denote the shot label by $k \in \{1,2,3,4\}$ and the corresponding number of gear shifts by $g=k-1$.
This classification enables a structured analysis of increasingly complex parking behaviors.

A gear-shift point is a forward$\leftrightarrow$reverse flip in the two-class motion state sequence.
When a predicted motion state sequence is available, predicted gear-shift points follow the same transition rule.
When an explicit motion-state prediction is unavailable, we infer a directional switch from the predicted trajectory.
Let $\mathbf{d}_t=\mathbf{p}_{t+1}-\mathbf{p}_t$ denote the consecutive displacement vector.
A gear-shift candidate is detected when two consecutive displacement vectors reverse their direction, i.e., $\mathbf{d}_{t-1}^{\top}\mathbf{d}_{t}<0$.

\textbf{Motion State Accuracy} is defined as the percentage of timesteps at which the dominant motion state (i.e., the state with the higher probability) is correctly predicted.
Both prediction and ground truth take values in $\{\text{fwd},\text{rev}\}$ only; there is no stationary class to score.
We compute the predicted motion state $\hat{c}_{i,j}$ and the ground-truth motion state  $c_{i,j}$ at each step: 
\begin{equation}
    \hat{c}_{i,j} = \underset{m \in \{\text{fwd}, \text{rev}\}}{\arg\max}(\hat{M}_{i,j}^m) \quad \text{and} 
    \quad c_{i,j} = \underset{m \in \{\text{fwd}, \text{rev}\}}{\arg\max}(M_{i,j}^m). 
\end{equation}
The accuracy is the mean of an indicator function $\mathbbm{1}$ over all timesteps: 
\begin{equation}
    M_{\mathrm{acc}} = \frac{1}{B \cdot N} \sum_{i=1}^B \sum_{j=1}^N \mathbbm{1}(\hat{c}_{i,j} = c_{i,j}) .    
\end{equation}

\textbf{Gear-shift Point Position Error} evaluates the 2D spatial accuracy of correctly predicted gear shifts. 
It measures the ADE between a ground-truth shift waypoint $p$ and its matched prediction $\hat{p}$. 

Trajectories are grouped by $k$-shot category (i.e., by $g$ ground-truth gear shifts).
For each category, we first calculate the mean error for the $n$-th sequential shift point $P_n$ across all samples: 
\begin{equation}
    M_{\text{dist}}(k\text{-shot}, P_n) = \frac{1}{C_{k,n}} \sum_{(p,\hat{p}) \in \mathcal{S}_{k,n}} \|p - \hat{p}\|_2, 
\end{equation}
where $\mathcal{S}_{k,n}$ is the set of all matched pairs for the $n$-th shift in this category, 
and $C_{k,n}$ is the total count. 
Finally, the Gear-shift Average Displacement Error (GS-ADE) for each category is the mean of these sequential gear-shift errors:
\begin{equation}
    M_{\text{GS-ADE}}(k\text{-shot}) = \frac{1}{g} \sum_{n=1}^{g} M_{\text{dist}}(k\text{-shot}, P_n) . 
\end{equation}
For $g=0$ ($1$-shot), $M_{\text{GS-ADE}}$ is undefined.

\textbf{Transition Hit Rate@$\tau$} is the ratio of matched ground-truth transitions to all ground-truth transitions, pooling trajectories in the evaluation set.
Unlike the dense $M_{\mathrm{acc}}$, it isolates gear-shift timing.
We reuse the same $\hat{c}_{i,j}$ and $c_{i,j}$ as above: a transition is any index $j\ge 1$ with $c_{i,j}\neq c_{i,j-1}$ (and analogously for $\hat{c}_{i,j}$).
A ground-truth transition is a hit if the nearest predicted transition falls within a tolerance of $\tau \in \{1, 2, 3\}$ frames; trajectories with no predicted transition contribute no matches.
The three tolerances separate missed transitions from small temporal offsets around the ground-truth shift point. 

\textbf{Gear-shift Count Precision}, \textbf{Count Recall}, and \textbf{Count F1.}
Within each sample, predicted and ground-truth shifts are paired in temporal order as the sequential points $P_n$.
A pair is a true positive once formed by this ordering; the match is not gated by a temporal or distance threshold, and the spatial error of each matched pair is reported by $M_{\text{dist}}$.
Let $N_{\mathrm{pred}}$ and $N_{\mathrm{GT}}$ be the numbers of predicted and ground-truth gear shifts in that sample.
Because the match is ungated, the confusion counts reduce to
\begin{equation}
\begin{aligned}
    \mathrm{TP} &= \min(N_{\mathrm{pred}}, N_{\mathrm{GT}}), \\
    \mathrm{FP} &= \max(N_{\mathrm{pred}} - N_{\mathrm{GT}}, 0), \\
    \mathrm{FN} &= \max(N_{\mathrm{GT}} - N_{\mathrm{pred}}, 0).
\end{aligned}
\end{equation}
These count metrics are separate from Transition Hit Rate@$\tau$: they score how many shifts are recovered, not whether each shift falls inside $\pm\tau$ frames.
Count recall is the ratio of these ordered matches to all ground-truth gear-shift points, pooling trajectories in the evaluation set.
It measures coverage of ground-truth shifts and does not penalize surplus predictions.
Count precision is the ratio of the same matches to all predicted gear-shift points.
Predictions beyond the paired set are counted as false positives, so count precision exposes over-triggering, including on $1$-shot trajectories that contain no ground-truth shift.
The count F1-score is the harmonic mean of count precision and count recall.

\begin{table*}[htbp]
    \centering
    \small
    
    \caption{Performance Benchmarking of Trajectory and Gear-shift Point. }
    \label{tab:exp_compare}
    
    {\setlength{\tabcolsep}{3pt}%
    \renewcommand{\arraystretch}{1.05}%
    \resizebox{\textwidth}{!}{%
    \newcommand{\gsbox}[1]{%
      \begingroup
      \setbox0=\hbox{$\mathbf{0.00 \pm 0.00}$}%
      \makebox[\wd0][c]{#1}%
      \endgroup
    }%
    \newcommand{\gsna}{\gsbox{/}}%
    \newcommand{\gsval}[1]{\gsbox{$#1$}}%
    \newcommand{\gsbf}[1]{\gsbox{$\mathbf{#1}$}}%
    \begin{tabular}{l l cccc cccc}
    \toprule
    
    \multirow{2}{*}{Methods} & \multirow{2}{*}{\shortstack{Input\\Modality}} & \multicolumn{4}{c}{Trajectory Metrics (Mean $\pm$ Std)} & \multicolumn{4}{c}{Gear-shift Point Metrics (Mean $\pm$ Std)}  \\
    \cmidrule(lr){3-6} \cmidrule(lr){7-10}
    & & ADE & Four. Diff. & Haus. Dist. & AHE & 2S[$P_1$] & 3S[$P_1$--$P_2$] & 4S[$P_1$--$P_2$--$P_3$] & GS-ADE \\
    \midrule
    
    Hybrid A*\cite{dolgov2010path}$\dagger$ & Obstacle, 2D goal & $1.39$ & $26.91$ & $1.46$ & $7.6$ & $1.58$ & \gsval{1.81} -- \gsval{1.98} & \gsval{2.82} -- \gsval{2.21} -- \gsval{1.63} & $1.73$ \\
    
    TransFuser\cite{chitta2022transfuser}$\ddagger$ & 4$\times$RGB, 2D goal & $1.07 \pm 0.02$ & $19.93 \pm 0.22$ & $\mathbf{1.14 \pm 0.03}$ & $/$ & $1.25 \pm 0.25$ & \gsval{2.04 \pm 1.14} -- \gsna & \gsna -- \gsna -- \gsna & $1.65 \pm 0.69$ \\
    
    ParkingE2E\cite{10801763} & 4$\times$RGB, BEV slot & $2.45 \pm 0.99$ & $36.86 \pm 6.48$ & $2.76 \pm 1.05$ & $/$ & $3.96 \pm 1.78$ & \gsval{4.43 \pm 2.24} -- \gsval{1.77 \pm 1.04} & \gsval{3.04 \pm 0.96} -- \gsna -- \gsna & $4.07 \pm 1.86$ \\
    
    CAA-Policy\cite{chen2025endtoendvisualautonomousparking} w/o aux.$\S$ & 4$\times$RGB, BEV slot & $3.18 \pm 0.31$ & $66.18 \pm 7.77$ & $4.35 \pm 0.57$ & $19.2 \pm 2.5$ & $2.62 \pm 0.69$ & \gsval{2.33 \pm 1.91} -- \gsval{3.28 \pm 1.11} & \gsna -- \gsna -- \gsna & $2.63 \pm 0.71$ \\
    
    CAA-Policy\cite{chen2025endtoendvisualautonomousparking}              & 4$\times$RGB, BEV slot & $2.70 \pm 0.56$ & $55.92 \pm 9.39$ & $3.84 \pm 0.57$ & $16.3 \pm 7.3$ & $2.34 \pm 0.40$ & \gsval{1.30 \pm 0.46} -- \gsval{2.14 \pm 0.11} & \gsna -- \gsna -- \gsna & $2.32 \pm 0.39$ \\
    
    Ours & 4$\times$RGB, Fourier slot & $\mathbf{1.06 \pm 0.02}$ & $\mathbf{18.56 \pm 0.50}$ & $1.15 \pm 0.03$ & $\mathbf{5.5 \pm 0.0}$ & $\mathbf{1.10 \pm 0.04}$ & \gsbf{0.91 \pm 0.04} -- \gsbf{1.02 \pm 0.05} & \gsbf{0.87 \pm 0.15} -- \gsbf{1.16 \pm 0.09} -- \gsbf{1.23 \pm 0.23} & $\mathbf{1.07 \pm 0.02}$ \\
    
    \bottomrule
    \end{tabular}%
    }%
    }%
    \vspace{0.35em}
    \begin{minipage}{\textwidth}
    \footnotesize
    To ensure a fair comparison, we utilize the official implementations for the baseline methods. \\
    $\dagger$ Hybrid~A* is a deterministic planner; we report a single-run result without mean$\pm$std. \\
    $\ddagger$ A modified TransFuser adapted to a vision-only setting. 
    It sets the 2D coordinates of the target slot as the goal location. \\
    $\S$ CAA-Policy's ``w/o aux.'' set excludes auxiliary tasks (i.e., depth estimation and BEV semantic segmentation), providing an input configuration more closely aligned with the primary benchmarking baselines. 
    \end{minipage}
    
    \end{table*}
 
\begin{table*}[t]
    \centering
    \caption{Ablation Study of Different Model Components. }
    \label{tab:ablation_study}
    \resizebox{\textwidth}{!}{%
    \begin{tabular}{l cc cccccc cccccc}
    \toprule
    \multirow{2}{*}{Input} & \multirow{2}{*}{Motion} & \multirow{2}{*}{Sched.} & \multicolumn{6}{c}{Trajectory Metrics (Mean $\pm$ Std)} & \multicolumn{6}{c}{Motion State Metrics (Mean $\pm$ Std)} \\
    \cmidrule(lr){4-9} \cmidrule(lr){10-15}
     & & & ADE & Four. & Haus. & AHE & mFDE & mFHE & Acc. & Hit Rate@$\tau{=}1$ & Avg. Dist. & Count Prec. & Count Recall & Count F1 \\
    \midrule
    BEV     &              &              & $2.45 \pm 0.99$ & $36.86 \pm 6.48$ & $2.76 \pm 1.05$ & /   & / & / & /     & /     & /     & / & / & / \\
    Fourier &              &              & $2.66 \pm 0.29$ & $46.64 \pm 8.04$ & $3.03 \pm 0.39$ & /   & / & / & /     & /     & /     & / & / & / \\
    Fourier & $\checkmark$ &              & $1.65 \pm 1.00$ & $28.35 \pm 16.24$ & $1.89 \pm 1.09$ & $13.1 \pm 13.0$ & $1.98 \pm 0.89$ & $17.8 \pm 22.1$ & $0.81 \pm 0.05$ & $0.15 \pm 0.11$ & $2.88 \pm 3.08$ & $0.77 \pm 0.05$ & $0.70 \pm 0.28$ & $0.72 \pm 0.19$ \\
    Fourier & $\checkmark$ & $\checkmark$ & $\mathbf{1.06 \pm 0.02}$ & $\mathbf{18.56 \pm 0.50}$ & $\mathbf{1.15 \pm 0.03}$ & $\mathbf{5.5 \pm 0.0}$ & $\mathbf{1.32 \pm 0.03}$ & $\mathbf{4.6 \pm 0.1}$ & $\mathbf{0.84 \pm 0.01}$ & $\mathbf{0.21 \pm 0.01}$ & $\mathbf{1.07 \pm 0.02}$ & $\mathbf{0.82 \pm 0.00}$ & $\mathbf{0.88 \pm 0.01}$ & $\mathbf{0.85 \pm 0.00}$ \\
    \bottomrule
    \end{tabular}
    }
\end{table*}
 
\begin{table*}[tbp]
    \centering
    \caption{Staged selection of sampling schedule and query-infusion operator. }
    \label{tab:scheduler_fusion_ablation}
    \setlength{\tabcolsep}{2.5pt}%
    \renewcommand{\arraystretch}{1.05}%
    \resizebox{\textwidth}{!}{%
    \begin{tabular}{@{}c cc c c cccc cccc cc@{}}
    \toprule
    \multirow{2}{*}{Set} & \multirow{2}{*}{Sched.} & \multirow{2}{*}{Fusion} & Traj. Loss & Mot. Loss & \multicolumn{4}{c}{Trajectory Metrics (Mean $\pm$ Std)} & \multicolumn{4}{c}{Motion State Metrics (Mean $\pm$ Std)} & \multicolumn{2}{c}{Gear-Shift Point Metrics (Mean $\pm$ Std)} \\
    \cmidrule(lr){6-9}\cmidrule(lr){10-13}\cmidrule(lr){14-15}
     &  &  & FDE+FHE & Smooth+Trans. & ADE & AHE & mFDE & mFHE & Acc. & Hit Rate@$\tau{=}1$ & Hit Rate@$\tau{=}2$ & Hit Rate@$\tau{=}3$ & Count Prec. & Count F1 \\
    \midrule
    1 & Cosine      & Add    & -- & -- & $2.05 \pm 0.84$ & $9.1 \pm 3.1$ & $2.30 \pm 0.93$ & $8.5 \pm 3.8$ & $0.76 \pm 0.07$ & $0.15 \pm 0.05$ & $0.24 \pm 0.08$ & $0.31 \pm 0.10$ & $0.70 \pm 0.11$ & $0.74 \pm 0.08$ \\
    2 & Exponential & Add    & -- & -- & $2.70 \pm 0.12$ & $17.7 \pm 5.7$ & $2.91 \pm 0.14$ & $26.8 \pm 13.6$ & $0.72 \pm 0.03$ & $0.08 \pm 0.04$ & $0.12 \pm 0.06$ & $0.16 \pm 0.08$ & $0.69 \pm 0.02$ & $0.61 \pm 0.12$ \\
    3 & Linear      & Add    & -- & -- & $1.66 \pm 0.95$ & $7.4 \pm 2.6$ & $2.03 \pm 1.04$ & $7.7 \pm 3.9$ & $0.79 \pm 0.08$ & $0.17 \pm 0.07$ & $0.26 \pm 0.10$ & $0.34 \pm 0.12$ & $0.79 \pm 0.09$ & $0.80 \pm 0.09$ \\
    4 & Linear      & Concat & -- & -- & $1.77 \pm 0.72$ & $6.7 \pm 1.3$ & $2.16 \pm 0.81$ & $6.4 \pm 2.8$ & $0.78 \pm 0.05$ & $0.16 \pm 0.06$ & $0.24 \pm 0.08$ & $0.31 \pm 0.10$ & $0.78 \pm 0.05$ & $0.77 \pm 0.06$ \\
    5 & Linear      & Gated  & -- & -- & $1.53 \pm 0.77$ & $6.4 \pm 1.6$ & $1.88 \pm 0.90$ & $5.4 \pm 1.5$ & $0.81 \pm 0.05$ & $0.18 \pm 0.06$ & $0.28 \pm 0.08$ & $0.36 \pm 0.10$ & $0.76 \pm 0.11$ & $0.78 \pm 0.10$ \\
    \midrule
    6 & Linear      & Gated  & $\checkmark$ & -- & $1.08 \pm 0.01$ & $5.6 \pm 0.1$ & $1.35 \pm 0.02$ & $4.6 \pm 0.1$ & $0.84 \pm 0.01$ & $\mathbf{0.22 \pm 0.00}$ & $\mathbf{0.32 \pm 0.01}$ & $0.41 \pm 0.01$ & $0.81 \pm 0.00$ & $0.84 \pm 0.00$ \\
    7 & Linear      & Gated  & $\checkmark$ & $\checkmark$ & $\mathbf{1.06 \pm 0.02}$ & $\mathbf{5.5 \pm 0.0}$ & $\mathbf{1.32 \pm 0.03}$ & $\mathbf{4.6 \pm 0.1}$ & $\mathbf{0.84 \pm 0.01}$ & $0.21 \pm 0.01$ & $0.32 \pm 0.00$ & $\mathbf{0.42 \pm 0.01}$ & $\mathbf{0.82 \pm 0.00}$ & $\mathbf{0.85 \pm 0.00}$ \\
    \bottomrule
    \end{tabular}%
    }
\end{table*}

\begin{table}[htbp]
\centering
\caption{End-to-End Inference Cost on a Single NVIDIA GeForce RTX~4090.}
\label{tab:runtime_e2e}
\setlength{\tabcolsep}{3.0pt}%
\renewcommand{\arraystretch}{1.05}%
\resizebox{\columnwidth}{!}{%
\begin{tabular}{l cccc}
\toprule
Methods & Params (M) &  Lat.\ (ms) & FPS & VRAM (GB) \\
\midrule
ParkingE2E \cite{10801763}  & $46.965$  & $167.44$ & $5.97$  & $1.494$ \\
TransFuser \cite{chitta2022transfuser}  & $73.758$  & $37.78$  & $26.47$ & $1.590$ \\
CAA-Policy \cite{chen2025endtoendvisualautonomousparking}  & $35.415$  & $239.58$ & $4.17$  & $1.234$ \\
Ours        & $49.017$  & $98.77$  & $10.12$ & $1.518$ \\
\bottomrule
\end{tabular}%
}%
\vspace{0.35em}
\begin{minipage}{\columnwidth}
\footnotesize
FP32, $B{=}1$, $4{\times}256{\times}256$. \\
Latency: wall-clock (CUDA sync + CPU), images$\to$detokenized trajectory; FPS~$=1000/$latency.
VRAM: \texttt{nvidia-smi} process memory. \\
\end{minipage}
\end{table}

\subsection{Quantitative Results}   
To comprehensively validate the effectiveness of the proposed method, we select Hybrid A* \cite{dolgov2010path} as the baseline, 
and three representative open-source learning-based E2E methods: 
TransFuser \cite{chitta2022transfuser}, ParkingE2E \cite{10801763}, and CAA-Policy \cite{chen2025endtoendvisualautonomousparking}. 
To ensure a fair and reproducible evaluation, we focus our empirical comparisons on methods with the same 30-waypoint trajectory prediction.
To verify that the reported results are not due to random variation, we train each experimental set with random seeds $\{4, 8, 16\}$ and evaluate each run independently.

As shown in Table~\ref{tab:exp_compare}, our approach achieves superior performance across most key metrics for trajectory error. 
As the Modular Plan method, Hybrid A* achieves robust performance during the evaluation, but it underperforms on trajectory similarity to expert demonstrations.  

As the Geometric Plan, ParkingE2E \cite{10801763} is weak on multi-shot trajectories, and gear-shift positions are often far from the expert.
As the Unified Kinematic Plan, CAA-Policy \cite{chen2025endtoendvisualautonomousparking} tightens gear-shift point error by conditioning on control signals and on depth with BEV semantic segmentation supervision.
SunnyParking achieves trajectory accuracy comparable to TransFuser while substantially improving heading and gear-shift prediction, without auxiliary supervision.
This is convenient for real-world deployment when LiDAR/depth cameras and BEV semantic segmentation labels are unavailable.   

Regarding gear-shift point error, we conduct a detailed investigation into the model's advantages in multi-shot parking scenarios. 
While conventional methods degrade on multi-shot trajectories as complexity increases, 
our method maintains high precision at gear-shift points while generating human-like trajectories, even in 4-shot scenarios.

\subsection{Runtime \& Computational Efficiency}
\label{sec:runtime}

We benchmark the four learning-based methods on an NVIDIA RTX~4090.
Since the encoders are similar across methods, the planner module dominates wall-clock cost.

As shown in Table~\ref{tab:runtime_e2e}, TransFuser~\cite{chitta2022transfuser} is fastest (37.78\,ms, 26.47\,FPS) with its non-AR GRU.
Inspired by FlashDrive~\cite{li2026flashdrive}, we use CUDA graphs to cut AR kernel-launch overhead and reach 98.77\,ms (10.12\,FPS), $2.43\times$ faster than CAA-Policy~\cite{chen2025endtoendvisualautonomousparking}.

To assess real-world deployability, 
we contextualize the inference latency within typical parking constraints. 
Assuming a maximum parking speed of $10$\,km/h ($\approx 2.78$\,m/s), 
our optimized $\sim 10$\,Hz execution rate translates to an inter-frame displacement of approximately $0.28$\,m. 
This temporal resolution provides a reasonable reaction margin for downstream controllers to execute precise maneuvers in constrained spaces.

\subsection{Ablation Studies}
\begin{table*}[htbp]
    \centering
    \caption{Leave-one-out Ablation on Trajectory and Motion State Metrics (baseline: full SunnyParking model). }
    \label{tab:ablation_controlled_traj}
    \resizebox{\textwidth}{!}{%
    \begin{tabular}{l l cc ccc cccc cccc}
        \toprule
        \multirow{2}{*}{Variant} & \multirow{2}{*}{Target} & \multirow{2}{*}{Motion} & \multirow{2}{*}{Sched.} & \multicolumn{3}{c}{Loss} & \multicolumn{4}{c}{Trajectory Metrics (Mean $\pm$ Std)} & \multicolumn{4}{c}{Motion State Metrics (Mean $\pm$ Std)} \\
        \cmidrule(lr){5-7}\cmidrule(lr){8-11}\cmidrule(lr){12-15}
         & & & & FDE+FHE & Smooth & Trans. & ADE & AHE & mFDE & mFHE & Acc. & Hit Rate@$\tau{=}1$ & Hit Rate@$\tau{=}2$ & Hit Rate@$\tau{=}3$ \\
        \midrule
        SunnyParking     & Fourier & $\checkmark$ & Linear & $\checkmark$ & $\checkmark$ & $\checkmark$ & $\mathbf{1.06 \pm 0.02}$ & $\mathbf{5.5 \pm 0.0}$ & $\mathbf{1.32 \pm 0.03}$ & $\mathbf{4.6 \pm 0.1}$ & $\mathbf{0.84 \pm 0.01}$ & $\mathbf{0.21 \pm 0.01}$ & $\mathbf{0.32 \pm 0.00}$ & $\mathbf{0.42 \pm 0.01}$ \\
        \midrule
        BEV target          & BEV     & $\checkmark$ & Linear & $\checkmark$ & $\checkmark$ & $\checkmark$ & $2.58 \pm 0.27$ & $22.1 \pm 3.2$ & $3.00 \pm 0.22$ & $34.4 \pm 6.7$ & $0.71 \pm 0.07$ & $0.13 \pm 0.01$ & $0.20 \pm 0.00$ & $0.27 \pm 0.01$ \\
        w/o mot. state   & Fourier & --           & Linear & $\checkmark$ & --           & --           & $1.06 \pm 0.01$ & $5.6 \pm 0.1$ & $1.31 \pm 0.04$ & $4.8 \pm 0.2$ & $/$ & $/$ & $/$ & $/$ \\
        \midrule
        w/o FDE+FHE         & Fourier & $\checkmark$ & Linear & --           & $\checkmark$ & $\checkmark$ & $2.74 \pm 0.08$ & $18.8 \pm 7.5$ & $3.13 \pm 0.07$ & $28.6 \pm 14.9$ & $0.70 \pm 0.06$ & $0.10 \pm 0.03$ & $0.16 \pm 0.05$ & $0.22 \pm 0.06$ \\
        w/o Smooth          & Fourier & $\checkmark$ & Linear & $\checkmark$ & --           & $\checkmark$ & $1.61 \pm 0.93$ & $6.7 \pm 2.1$ & $2.02 \pm 1.18$ & $5.8 \pm 2.0$ & $0.80 \pm 0.06$ & $0.18 \pm 0.06$ & $0.28 \pm 0.08$ & $0.36 \pm 0.10$ \\
        w/o Trans.          & Fourier & $\checkmark$ & Linear & $\checkmark$ & $\checkmark$ & --           & $1.07 \pm 0.03$ & $5.6 \pm 0.1$ & $1.32 \pm 0.05$ & $4.6 \pm 0.1$ & $0.84 \pm 0.01$ & $0.21 \pm 0.00$ & $0.32 \pm 0.01$ & $0.41 \pm 0.00$ \\
        \midrule
        w/o sched. & Fourier & $\checkmark$ & --     & $\checkmark$ & $\checkmark$ & $\checkmark$ & $1.65 \pm 1.00$ & $13.1 \pm 13.0$ & $1.98 \pm 0.89$ & $17.8 \pm 22.1$ & $0.81 \pm 0.05$ & $0.15 \pm 0.11$ & $0.23 \pm 0.17$ & $0.29 \pm 0.21$ \\
        \bottomrule
    \end{tabular}%
    }
\end{table*}

\begin{table*}[htbp]
    \centering
    \caption{Leave-one-out Ablation on Gear-shift Point Metrics (baseline: full SunnyParking model).}
    \label{tab:ablation_controlled_gear}
    \resizebox{\textwidth}{!}{%
    \newcommand{\gsna}{%
      \begingroup
      \setbox0=\hbox{$0.00 \pm 0.00$}%
      \makebox[\wd0][c]{/}%
      \endgroup
    }
    \begin{tabular}{l l cc ccc ccc cccc}
        \toprule
        \multirow{2}{*}{Variant} & \multirow{2}{*}{Target} & \multirow{2}{*}{Motion} & \multirow{2}{*}{Sched.} & \multicolumn{3}{c}{Loss} & \multicolumn{7}{c}{Gear-shift Point Metrics (Mean $\pm$ Std)} \\
        \cmidrule(lr){5-7}\cmidrule(lr){8-14}
         & & & & FDE+FHE & Smooth & Trans. & 2S[$P_1$] & 3S[$P_1$--$P_2$] & 4S[$P_1$--$P_2$--$P_3$] & GS-ADE & Count Recall & Count Prec. & Count F1 \\
        \midrule
        SunnyParking     & Fourier & $\checkmark$ & Linear & $\checkmark$ & $\checkmark$ & $\checkmark$ & $\mathbf{1.10 \pm 0.04}$ & $\mathbf{0.91 \pm 0.04}$ -- $\mathbf{1.02 \pm 0.05}$ & $\mathbf{0.87 \pm 0.15}$ -- $\mathbf{1.16 \pm 0.09}$ -- $\mathbf{1.23 \pm 0.23}$ & $\mathbf{1.07 \pm 0.02}$ & $\mathbf{0.88 \pm 0.01}$ & $\mathbf{0.82 \pm 0.00}$ & $\mathbf{0.85 \pm 0.00}$ \\
        \midrule
        BEV target          & BEV     & $\checkmark$ & Linear & $\checkmark$ & $\checkmark$ & $\checkmark$ & $5.10 \pm 1.43$ & $3.06 \pm 0.59$ -- $2.99 \pm 0.53$ & $5.92 \pm 3.18$ -- $4.26 \pm 0.33$ -- $3.06 \pm 1.10$ & $4.20 \pm 0.97$ & $0.80 \pm 0.05$ & $0.69 \pm 0.04$ & $0.73 \pm 0.02$ \\
        w/o mot. state   & Fourier & --           & Linear & $\checkmark$ & --           & --           & $1.23 \pm 0.06$ & $1.63 \pm 0.03$ -- $1.03 \pm 0.01$ & $1.87 \pm 0.20$ -- $1.16 \pm 0.18$ -- \gsna & $1.45 \pm 0.03$ & $0.33 \pm 0.04$ & $0.95 \pm 0.01$ & $0.49 \pm 0.05$ \\
        \midrule
        w/o FDE+FHE         & Fourier & $\checkmark$ & Linear & --           & $\checkmark$ & $\checkmark$ & $6.02 \pm 1.52$ & $3.92 \pm 0.44$ -- $3.92 \pm 0.46$ & $6.33 \pm 2.57$ -- $4.22 \pm 0.13$ -- $3.70 \pm 1.63$ & $5.00 \pm 0.88$ & $0.72 \pm 0.16$ & $0.65 \pm 0.06$ & $0.67 \pm 0.05$ \\
        w/o Smooth          & Fourier & $\checkmark$ & Linear & $\checkmark$ & --           & $\checkmark$ & $2.29 \pm 1.96$ & $1.58 \pm 1.07$ -- $1.77 \pm 1.21$ & $2.07 \pm 2.15$ -- $2.08 \pm 1.75$ -- $2.23 \pm 1.32$ & $2.02 \pm 1.58$ & $0.82 \pm 0.10$ & $0.75 \pm 0.09$ & $0.78 \pm 0.09$ \\
        w/o Trans.          & Fourier & $\checkmark$ & Linear & $\checkmark$ & $\checkmark$ & --           & $1.14 \pm 0.04$ & $0.96 \pm 0.03$ -- $1.06 \pm 0.04$ & $0.98 \pm 0.07$ -- $1.08 \pm 0.12$ -- $1.11 \pm 0.01$ & $1.11 \pm 0.02$ & $0.87 \pm 0.01$ & $0.82 \pm 0.01$ & $0.84 \pm 0.01$ \\
        \midrule
        w/o sched. & Fourier & $\checkmark$ & --     & $\checkmark$ & $\checkmark$ & $\checkmark$ & $3.59 \pm 4.15$ & $2.06 \pm 2.01$ -- $2.23 \pm 2.03$ & $3.20 \pm 4.23$ -- $2.18 \pm 1.96$ -- $1.25 \pm 0.15$ & $2.88 \pm 3.08$ & $0.70 \pm 0.28$ & $0.77 \pm 0.05$ & $0.72 \pm 0.19$ \\
        \bottomrule
    \end{tabular}%
    }
\end{table*}

In order to validate the effectiveness of each component in the proposed method, 
we conduct ablation studies to investigate the impact of the model components, training strategy, and the loss functions. 

\subsubsection{Progressive Addition of Proposed Components}
Table~\ref{tab:ablation_study} validates the contribution of each component relative to conventional baselines:  
\begin{itemize}
    \item \textbf{Input Representation:} Comparing BEV features against our proposed Fourier-based embedding.
    \item \textbf{Model Components:} Evaluating the baseline single trajectory decoder versus our dual-branch architecture.
    \item \textbf{Training Strategy:} Assessing the improvement of Scheduled Sampling. 
\end{itemize}

Replacing BEV target features with Fourier encoding and gated query infusion does not lower ADE and Hausdorff errors on its own (2.45$\rightarrow$2.66\,m and 2.76$\rightarrow$3.03\,m); Fourier difference increases (36.86$\rightarrow$46.64).
The model is given a continuous target prior, yet without motion state awareness and scheduled sampling the overall trajectory shape is not well preserved. 
Adding the motion state branch completes the YUKUI planner, yet without scheduled sampling the plan stays unstable and gear-shift localization remains unreliable.
Conditioning motion on the trajectory is not enough while the decoder is trained only on ground-truth history, so geometry at direction changes is still inconsistent.
Scheduled sampling then cuts Fourier and Hausdorff errors and improves gear-shift count recall, while sharply reducing AHE ($13.1 \rightarrow 5.5$).
It reduces the train--test exposure gap and helps the model stay stable on harder parking cases.

\def\BenchRowLabel#1{%
  \begin{tabular}[b]{@{}c@{}}%
    \parbox[c][\benchpanelht][c]{1.4em}{\centering\rotatebox{90}{\footnotesize\sffamily #1}}\\
    \rule{0pt}{\benchcapband}%
  \end{tabular}%
}
\FigureContentWide[htbp]{%
    \setlength{\tabcolsep}{1pt}%
    \renewcommand{\arraystretch}{1}%
    \setbox0=\hbox{\includegraphics[width=0.148\linewidth]{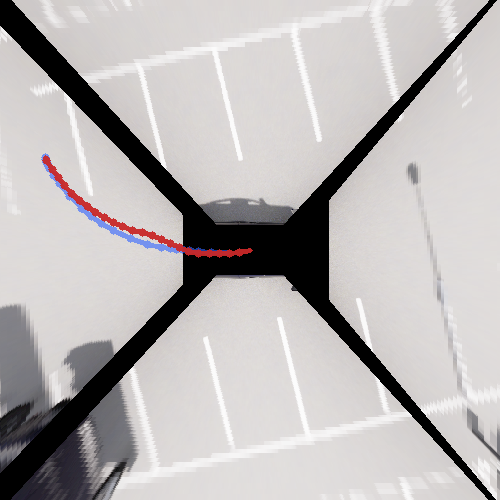}}%
    \edef\benchpanelht{\the\ht0}%
    \setbox2=\hbox{%
      \begin{subfigure}[b]{0.148\linewidth}%
        \includegraphics[width=\linewidth]{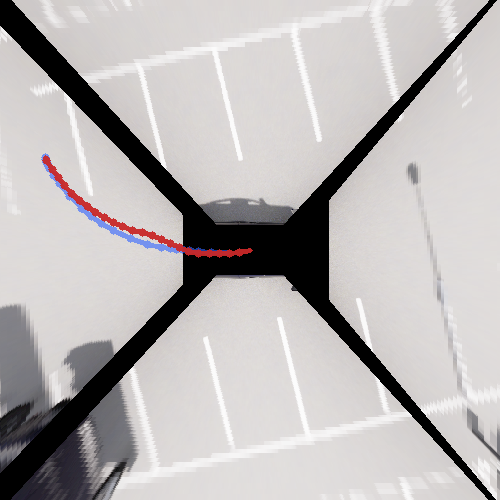}%
        \subcaption{}%
      \end{subfigure}%
    }%
    \edef\benchcapband{\the\dimexpr\ht2+\dp2-\ht0\relax}%
    \addtocounter{subfigure}{-1}%
    \begin{tabular}{@{}c@{\hspace{3pt}}c@{\hspace{2pt}}c@{\hspace{2pt}}c@{\hspace{2pt}}c@{\hspace{2pt}}c@{\hspace{2pt}}c@{}}
        &
        \scriptsize\sffamily 1-shot &
        \scriptsize\sffamily 2-shot &
        \scriptsize\sffamily 2-shot &
        \scriptsize\sffamily 3-shot &
        \scriptsize\sffamily 3-shot &
        \scriptsize\sffamily 4-shot \\[2pt]
        \BenchRowLabel{Hybrid A*} &
        \begin{subfigure}[b]{0.148\linewidth}
            \includegraphics[width=\linewidth]{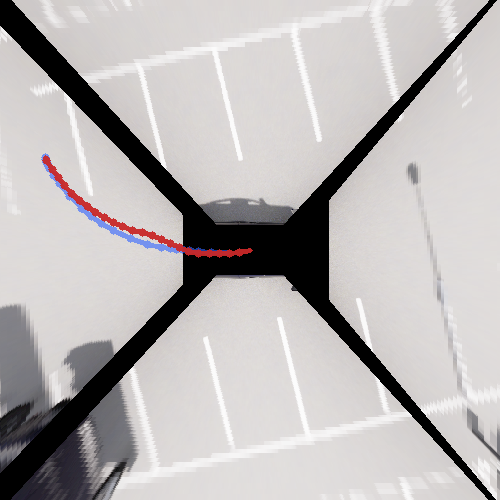}
            \subcaption{}
            \label{fig:bench_ha_a}
        \end{subfigure} &
        \begin{subfigure}[b]{0.148\linewidth}
            \includegraphics[width=\linewidth]{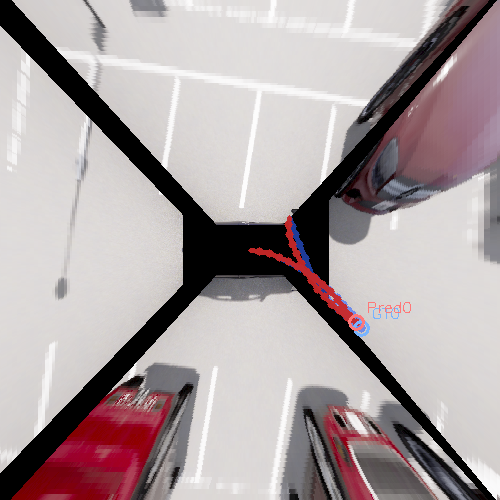}
            \subcaption{}
        \end{subfigure} &
        \begin{subfigure}[b]{0.148\linewidth}
            \includegraphics[width=\linewidth]{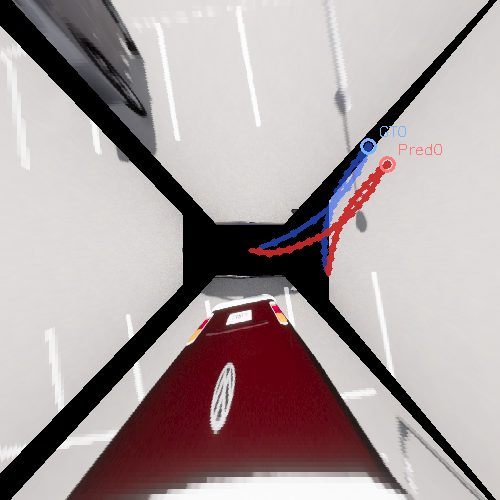}
            \subcaption{}
        \end{subfigure} &
        \begin{subfigure}[b]{0.148\linewidth}
            \includegraphics[width=\linewidth]{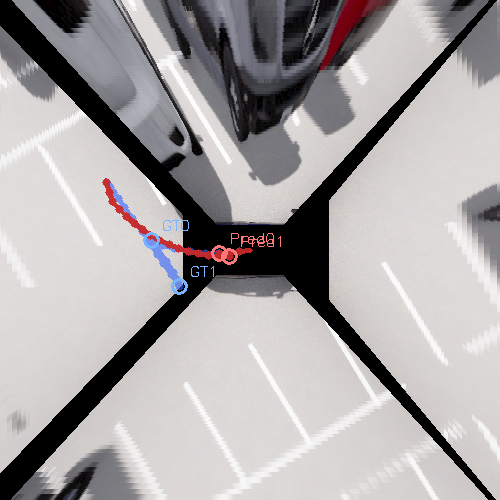}
            \subcaption{}
            \label{fig:bench_ha_d}
        \end{subfigure} &
        \begin{subfigure}[b]{0.148\linewidth}
            \includegraphics[width=\linewidth]{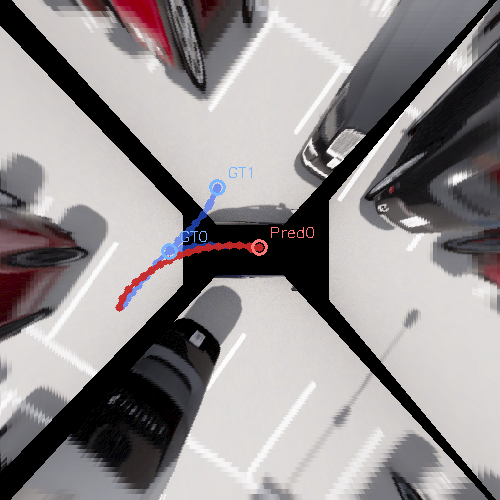}
            \subcaption{}
        \end{subfigure} &
        \begin{subfigure}[b]{0.148\linewidth}
            \includegraphics[width=\linewidth]{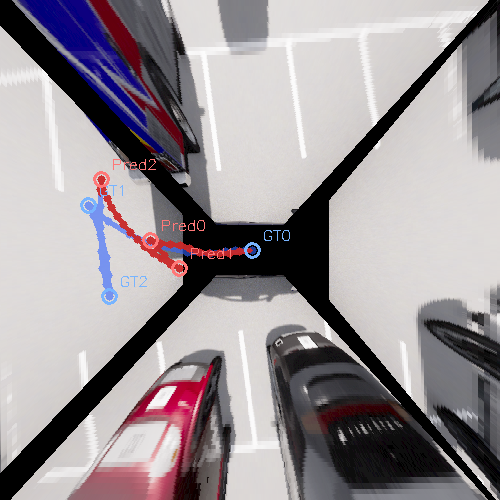}
            \subcaption{}
            \label{fig:bench_ha_f}
        \end{subfigure} \\[2pt]
        \BenchRowLabel{TransFuser} &
        \begin{subfigure}[b]{0.148\linewidth}
            \includegraphics[width=\linewidth]{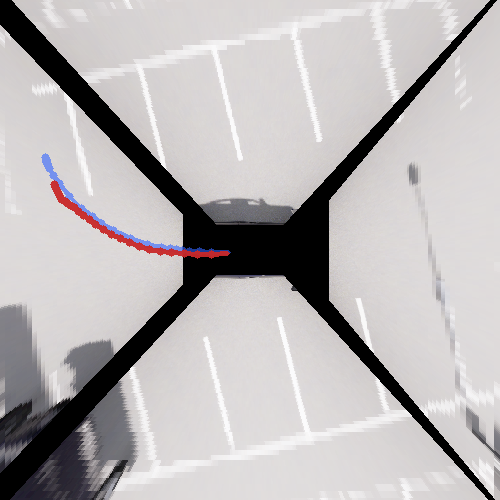}
            \subcaption{}
        \end{subfigure} &
        \begin{subfigure}[b]{0.148\linewidth}
            \includegraphics[width=\linewidth]{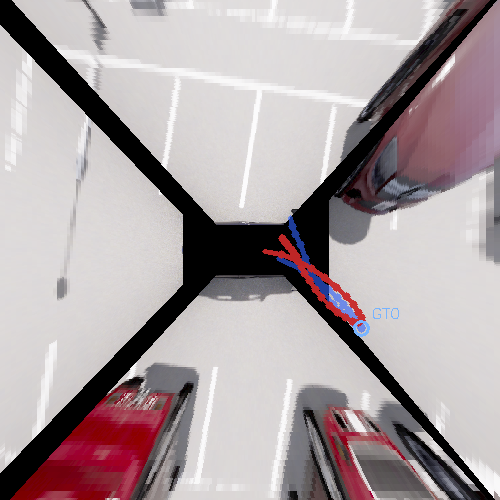}
            \subcaption{}
        \end{subfigure} &
        \begin{subfigure}[b]{0.148\linewidth}
            \includegraphics[width=\linewidth]{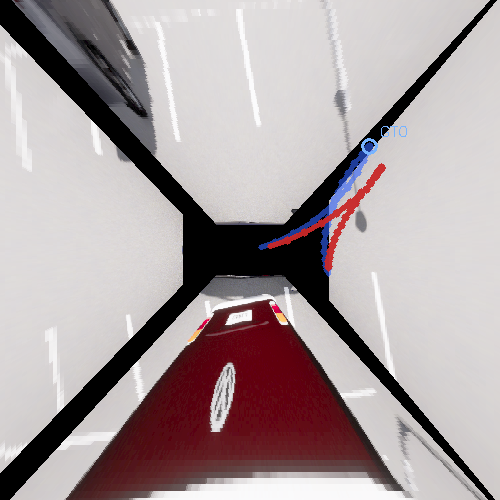}
            \subcaption{}
        \end{subfigure} &
        \begin{subfigure}[b]{0.148\linewidth}
            \includegraphics[width=\linewidth]{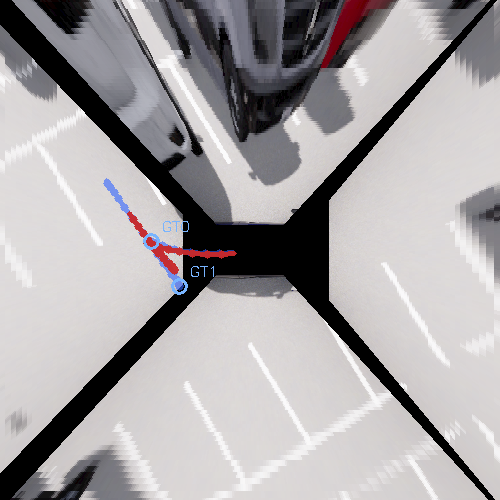}
            \subcaption{}
        \end{subfigure} &
        \begin{subfigure}[b]{0.148\linewidth}
            \includegraphics[width=\linewidth]{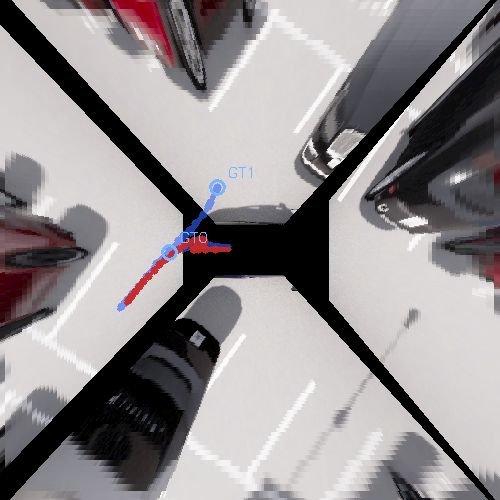}
            \subcaption{}
        \end{subfigure} &
        \begin{subfigure}[b]{0.148\linewidth}
            \includegraphics[width=\linewidth]{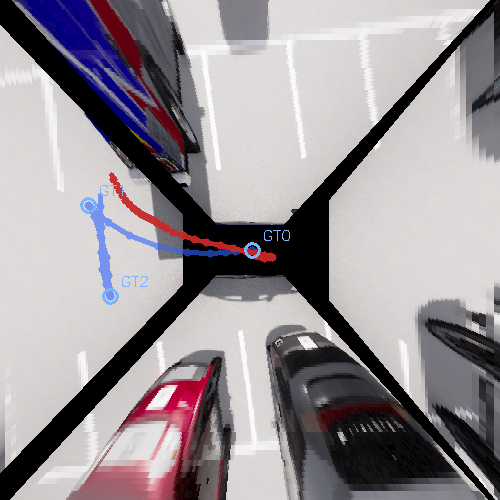}
            \subcaption{}
        \end{subfigure} \\[2pt]
        \BenchRowLabel{Ours} &
        \begin{subfigure}[b]{0.148\linewidth}
            \includegraphics[width=\linewidth]{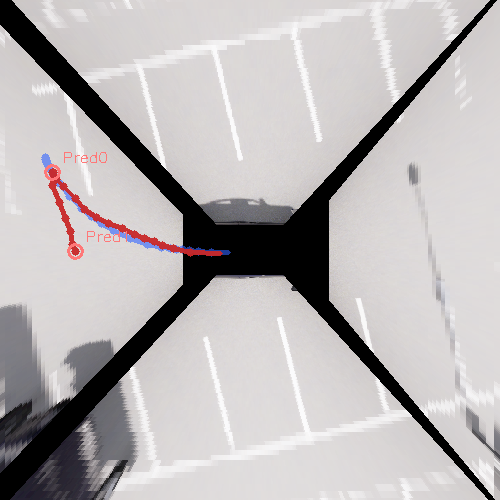}
            \subcaption{}
            \label{fig:bench_ours_m}
        \end{subfigure} &
        \begin{subfigure}[b]{0.148\linewidth}
            \includegraphics[width=\linewidth]{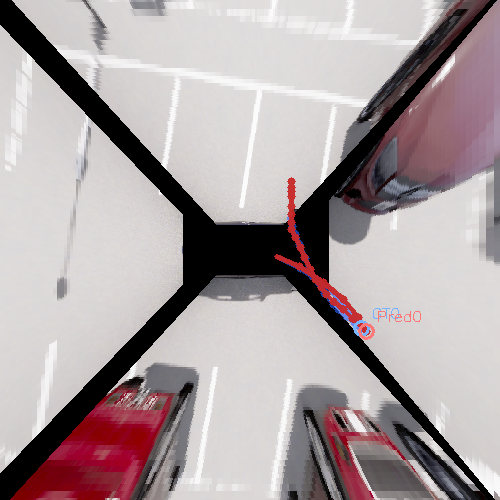}
            \subcaption{}
        \end{subfigure} &
        \begin{subfigure}[b]{0.148\linewidth}
            \includegraphics[width=\linewidth]{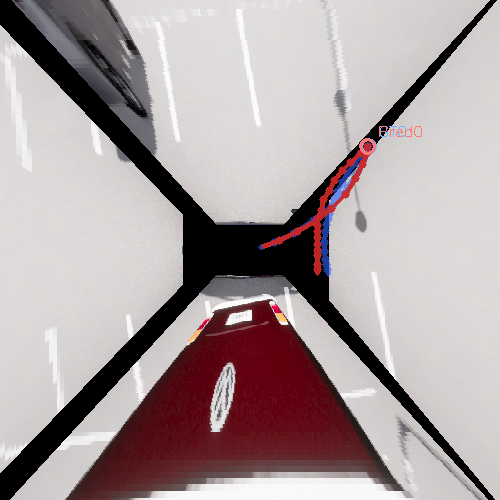}
            \subcaption{}
        \end{subfigure} &
        \begin{subfigure}[b]{0.148\linewidth}
            \includegraphics[width=\linewidth]{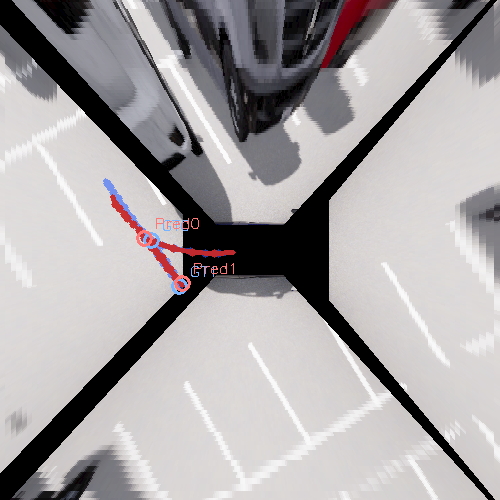}
            \subcaption{}
            \label{fig:benchmarking_traj_case_1}
        \end{subfigure} &
        \begin{subfigure}[b]{0.148\linewidth}
            \includegraphics[width=\linewidth]{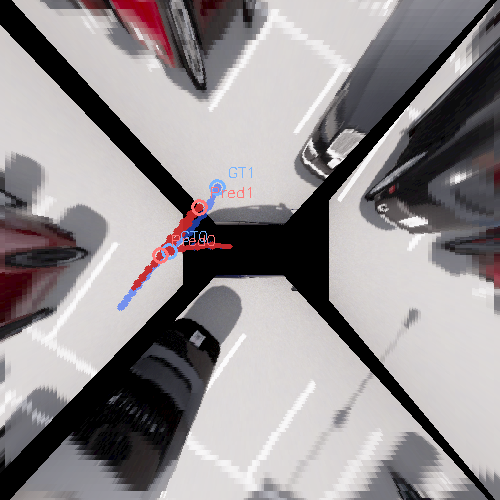}
            \subcaption{}
        \end{subfigure} &
        \begin{subfigure}[b]{0.148\linewidth}
            \includegraphics[width=\linewidth]{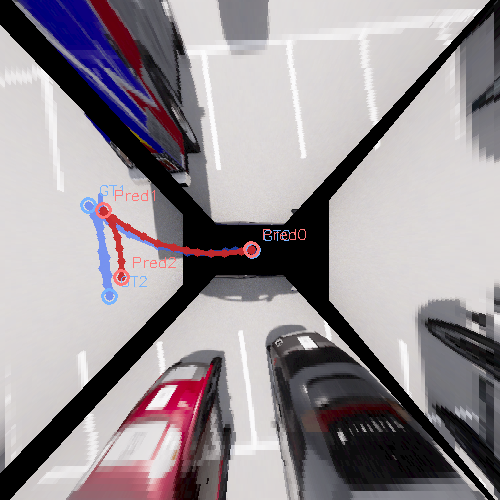}
            \subcaption{}
            \label{fig:benchmarking_traj_case_2}
        \end{subfigure}
    \end{tabular}%
}{Qualitative comparison on the held-out validation set.
    Blue denotes the ground truth; red denotes the prediction.
    Hybrid A* serves as a strong baseline, generating continuous trajectories directed at the target point. 
    However, it occasionally produces maneuvers that violate human driving behavior: 
    in (\subref{fig:bench_ha_d}), it executes frequent motion state switches over extremely short distances, 
    and the gear-shift timing and locations in (\subref{fig:bench_ha_f}) contradict human intuition. 
    The learning-based TransFuser can generate a more human-like trajectory. 
    However, it performs poorly at gear-shift points due to lacking explicit motion state modeling. 
    Its predicted trajectories also suffer from minor, unnatural curvatures; while these slight bends help waypoint-based KPIs, they harm real-world vehicle usability. 
    Our method aligns tightly with the human expert trajectory, 
    and the model accurately anticipates gear-shift timing and count, demonstrating human-like prediction. 
    We also observe a limitation: 
    in (\subref{fig:bench_ours_m}), our method tends to perform multiple gear shifts even in 1-shot scenarios. 
    In such a simple case (\subref{fig:bench_ha_a}), Hybrid A* yields a neater trajectory.
    \label{fig:benchmarking_traj_gear_viz}
    }

\def\WeatherRowLabel#1{%
  \begin{tabular}[b]{@{}c@{}}%
    \parbox[c][\dowpanelht][c]{1.15em}{\centering\rotatebox{90}{\footnotesize\sffamily #1}}\\
    \rule{0pt}{\dowcapband}%
  \end{tabular}%
}
\FigureContentWide[tbp]{%
    \setlength{\tabcolsep}{1pt}%
    \renewcommand{\arraystretch}{1}%
    \setbox0=\hbox{\includegraphics[width=0.148\linewidth]{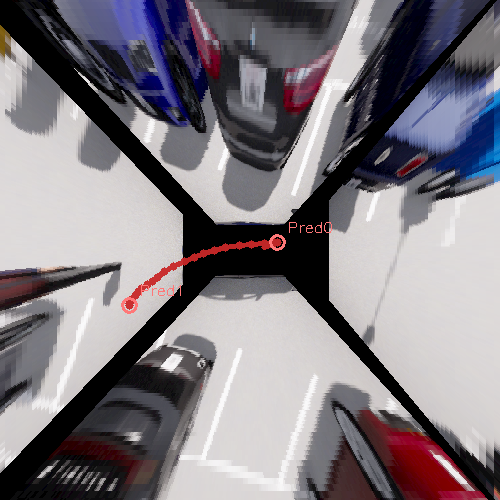}}%
    \edef\dowpanelht{\the\ht0}%
    \setbox2=\hbox{%
      \begin{subfigure}[b]{0.148\linewidth}%
        \includegraphics[width=\linewidth]{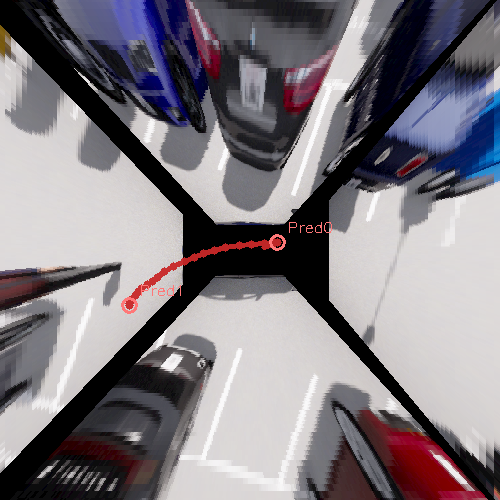}%
        \subcaption{}%
      \end{subfigure}%
    }%
    \edef\dowcapband{\the\dimexpr\ht2+\dp2-\ht0\relax}%
    \addtocounter{subfigure}{-1}%
    \begin{tabular}{@{}c@{\hspace{3pt}}c@{\hspace{2pt}}c@{\hspace{3pt}}|@{\hspace{3pt}}c@{\hspace{2pt}}c@{\hspace{2pt}}c@{\hspace{2pt}}c@{}}
        &
        \multicolumn{3}{c}{\footnotesize\sffamily\textbf{In-Distribution Weather}} &
        \multicolumn{3}{c}{\footnotesize\sffamily\textbf{Robustness to OOD Weather}} \\[1pt]
        &
        \scriptsize\sffamily Clear Noon  &
        \scriptsize\sffamily Clear Noon &
        \scriptsize\sffamily Clear Noon (Reference) &
        \scriptsize\sffamily Cloudy Sunset  &
        \scriptsize\sffamily HardRain Sunset  &
        \scriptsize\sffamily Clear Night  \\[2pt]
        \WeatherRowLabel{Pedestrian} &
        \begin{subfigure}[b]{0.148\linewidth}
            \includegraphics[width=\linewidth]{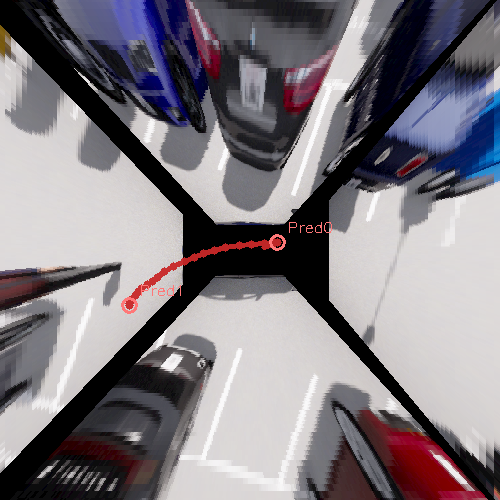}
            \subcaption{}
            \label{fig:dynamic_obstacle_weather_a}
        \end{subfigure} &
        \begin{subfigure}[b]{0.148\linewidth}
            \includegraphics[width=\linewidth]{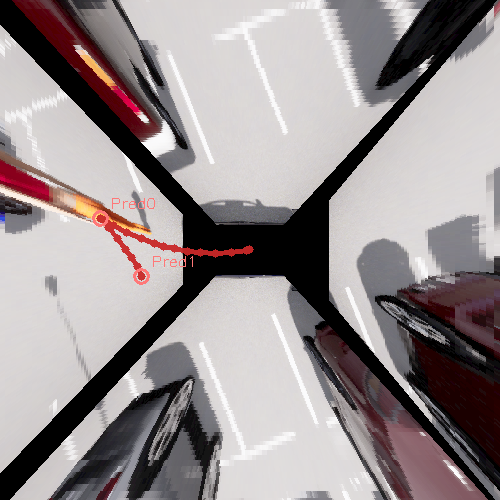}
            \subcaption{}
            \label{fig:dynamic_obstacle_weather_b}
        \end{subfigure} &
        \begin{subfigure}[b]{0.148\linewidth}
            \includegraphics[width=\linewidth]{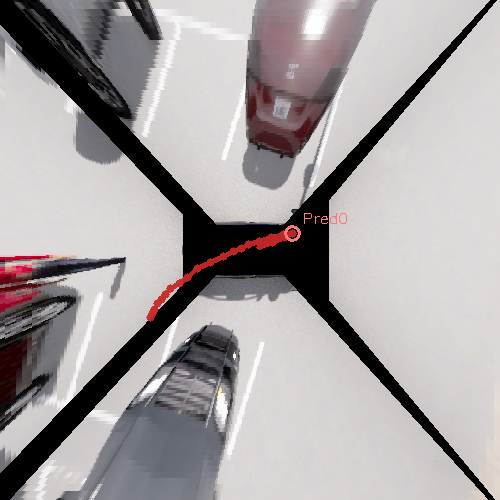}
            \subcaption{}
            \label{fig:dynamic_obstacle_weather_c}
        \end{subfigure} &
        \begin{subfigure}[b]{0.148\linewidth}
            \includegraphics[width=\linewidth]{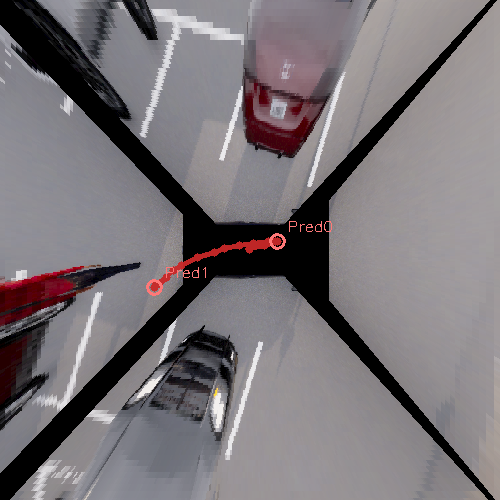}
            \subcaption{}
            \label{fig:dynamic_obstacle_weather_d}
        \end{subfigure} &
        \begin{subfigure}[b]{0.148\linewidth}
            \includegraphics[width=\linewidth]{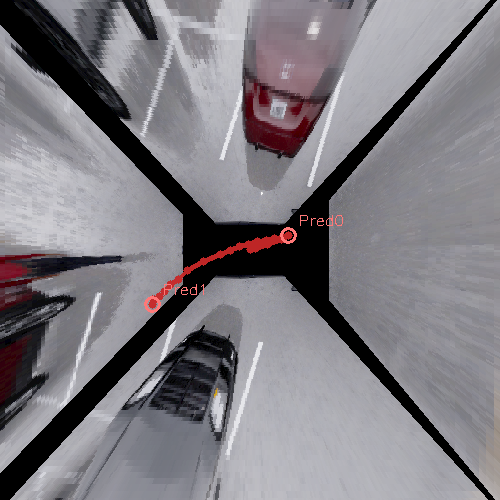}
            \subcaption{}
            \label{fig:dynamic_obstacle_weather_e}
        \end{subfigure} &
        \begin{subfigure}[b]{0.148\linewidth}
            \includegraphics[width=\linewidth]{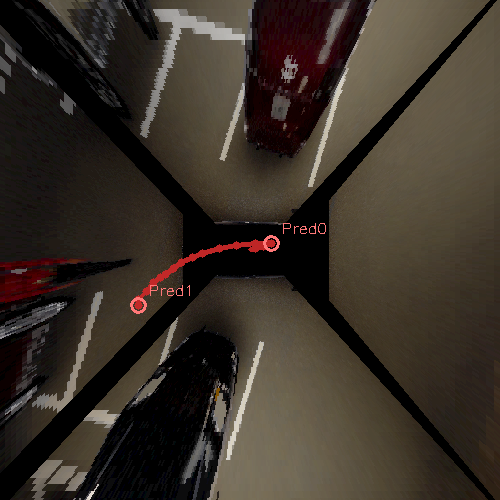}
            \subcaption{}
            \label{fig:dynamic_obstacle_weather_f}
        \end{subfigure} \\[2pt]
        \WeatherRowLabel{Vehicle} &
        \begin{subfigure}[b]{0.148\linewidth}
            \includegraphics[width=\linewidth]{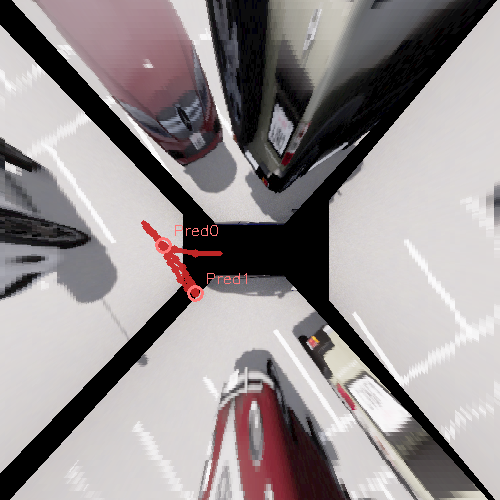}
            \subcaption{}
            \label{fig:dynamic_obstacle_weather_g}
        \end{subfigure} &
        \begin{subfigure}[b]{0.148\linewidth}
            \includegraphics[width=\linewidth]{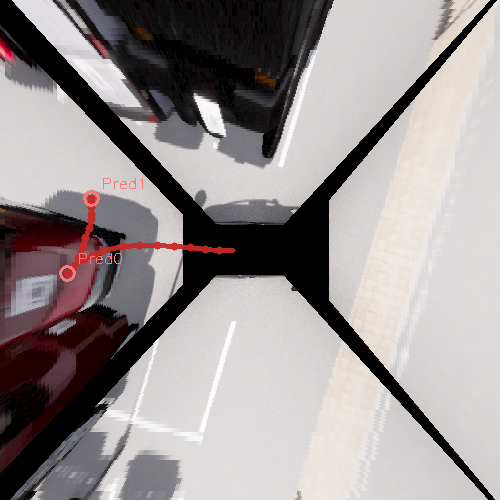}
            \subcaption{}
            \label{fig:dynamic_obstacle_weather_h}
        \end{subfigure} &
        \begin{subfigure}[b]{0.148\linewidth}
            \includegraphics[width=\linewidth]{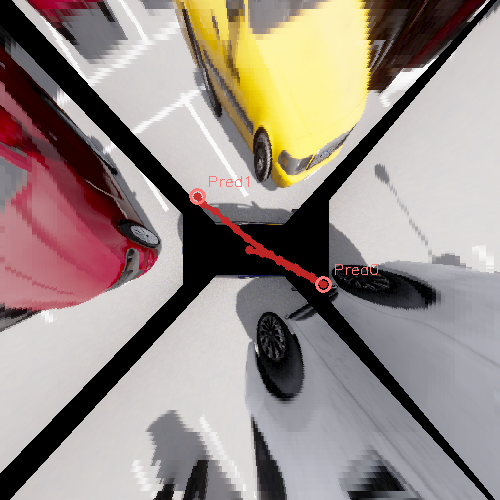}
            \subcaption{}
            \label{fig:dynamic_obstacle_weather_i}
        \end{subfigure} &
        \begin{subfigure}[b]{0.148\linewidth}
            \includegraphics[width=\linewidth]{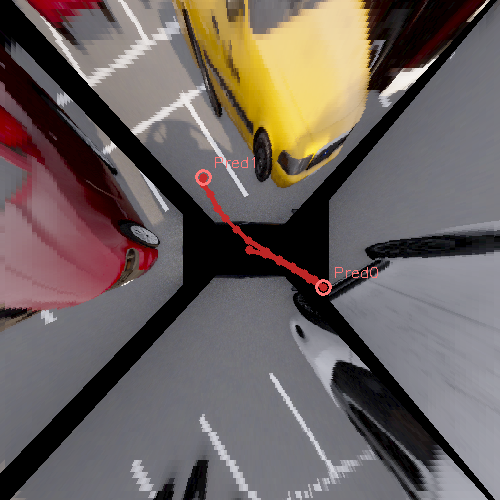}
            \subcaption{}
            \label{fig:dynamic_obstacle_weather_j}
        \end{subfigure} &
        \begin{subfigure}[b]{0.148\linewidth}
            \includegraphics[width=\linewidth]{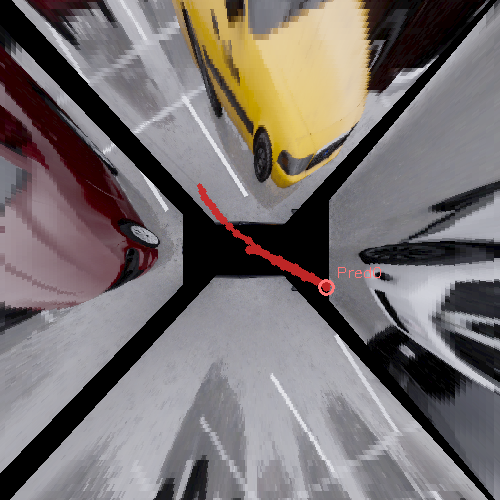}
            \subcaption{}
            \label{fig:dynamic_obstacle_weather_k}
        \end{subfigure} &
        \begin{subfigure}[b]{0.148\linewidth}
            \includegraphics[width=\linewidth]{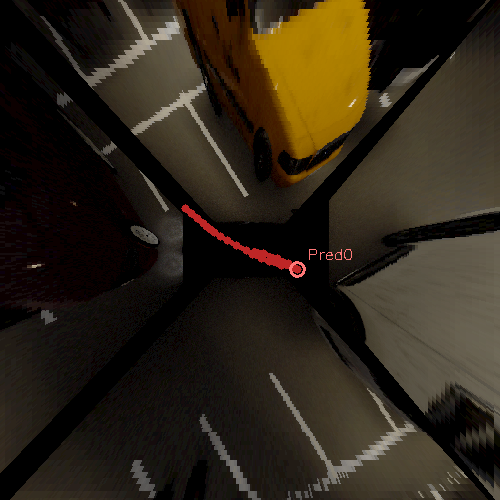}
            \subcaption{}
            \label{fig:dynamic_obstacle_weather_l}
        \end{subfigure} \\[2pt]
        \WeatherRowLabel{Multiple} &
        \begin{subfigure}[b]{0.148\linewidth}
            \includegraphics[width=\linewidth]{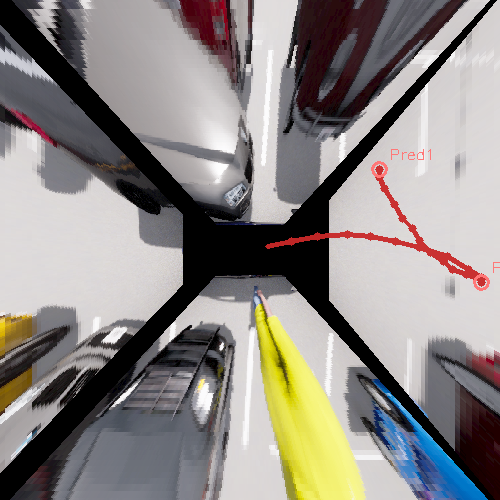}
            \subcaption{}
            \label{fig:dynamic_obstacle_weather_m}
        \end{subfigure} &
        \begin{subfigure}[b]{0.148\linewidth}
            \includegraphics[width=\linewidth]{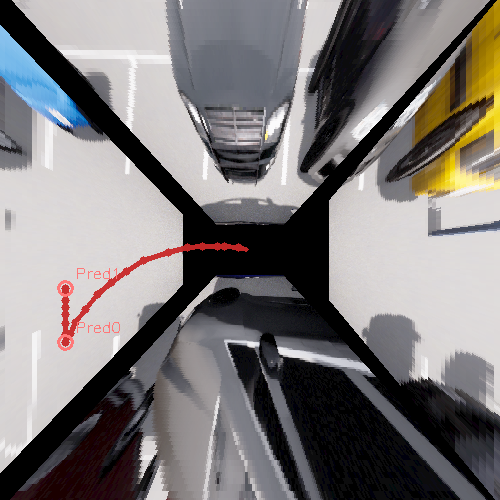}
            \subcaption{}
            \label{fig:dynamic_obstacle_weather_n}
        \end{subfigure} &
        \begin{subfigure}[b]{0.148\linewidth}
            \includegraphics[width=\linewidth]{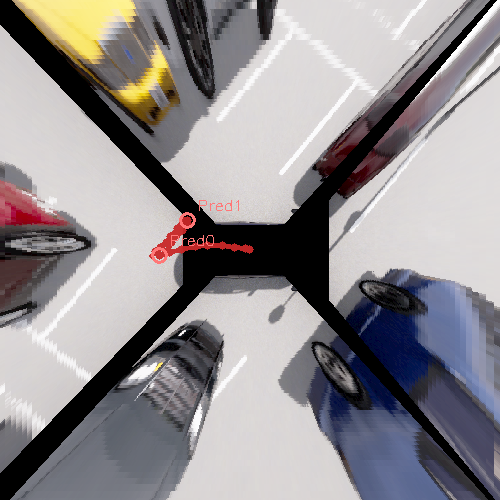}
            \subcaption{}
            \label{fig:dynamic_obstacle_weather_o}
        \end{subfigure} &
        \begin{subfigure}[b]{0.148\linewidth}
            \includegraphics[width=\linewidth]{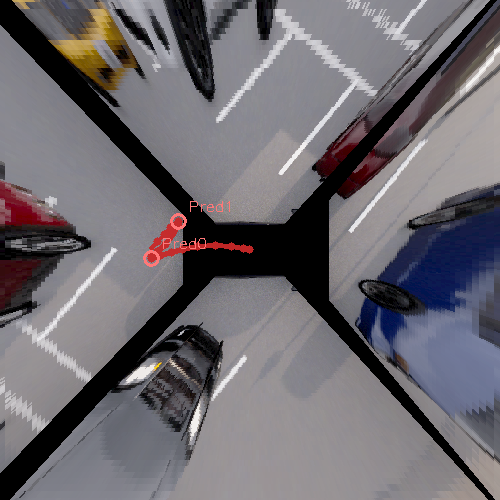}
            \subcaption{}
            \label{fig:dynamic_obstacle_weather_p}
        \end{subfigure} &
        \begin{subfigure}[b]{0.148\linewidth}
            \includegraphics[width=\linewidth]{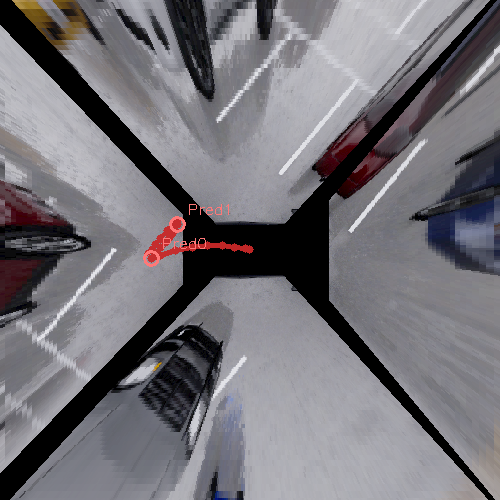}
            \subcaption{}
            \label{fig:dynamic_obstacle_weather_q}
        \end{subfigure} &
        \begin{subfigure}[b]{0.148\linewidth}
            \includegraphics[width=\linewidth]{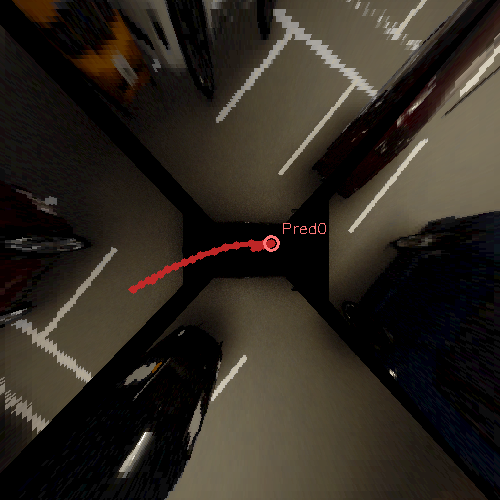}
            \subcaption{}
            \label{fig:dynamic_obstacle_weather_r}
        \end{subfigure}
    \end{tabular}%
}{Qualitative stress tests with dynamic obstacles under in-distribution and OOD weather conditions.
    Red denotes the prediction.
    Cols~1--3 show distinct scenarios under in-distribution weather;
    Col~3 is the shared Clear Noon reference;
    Cols~4--6 evaluate the same reference scenario under OOD weather.
    \label{fig:dynamic_obstacle_weather_viz}
    }

\def\HeatmapRowLabel#1{%
  \begin{tabular}[b]{@{}c@{}}%
    \parbox[c][\hmappanelht][c]{1.4em}{\centering\rotatebox{90}{\footnotesize\sffamily #1}}\\
    \rule{0pt}{\hmapcapband}%
  \end{tabular}%
}
\FigureContentSingle[!htbp]{%
    \setlength{\tabcolsep}{0.5pt}%
    \renewcommand{\arraystretch}{0.85}%
    \setbox0=\hbox{\includegraphics[width=0.42\linewidth]{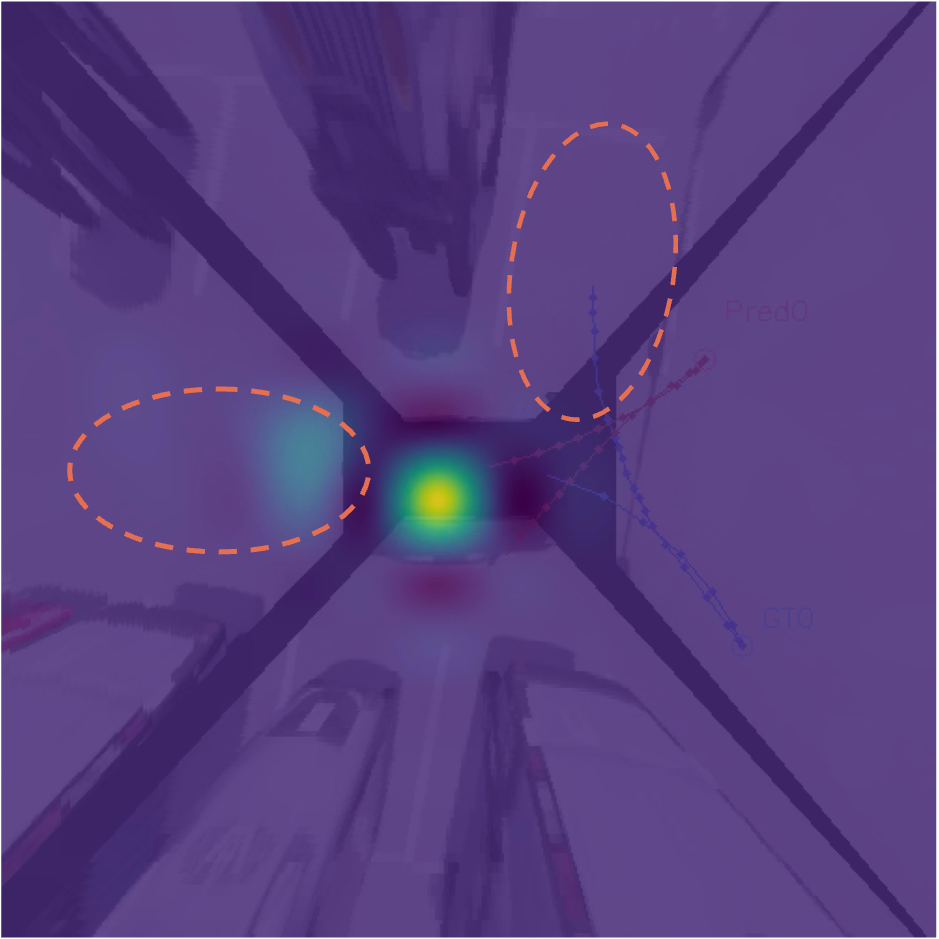}}%
    \edef\hmappanelht{\the\ht0}%
    \setbox2=\hbox{%
      \begin{subfigure}[b]{0.42\linewidth}%
        \includegraphics[width=\linewidth]{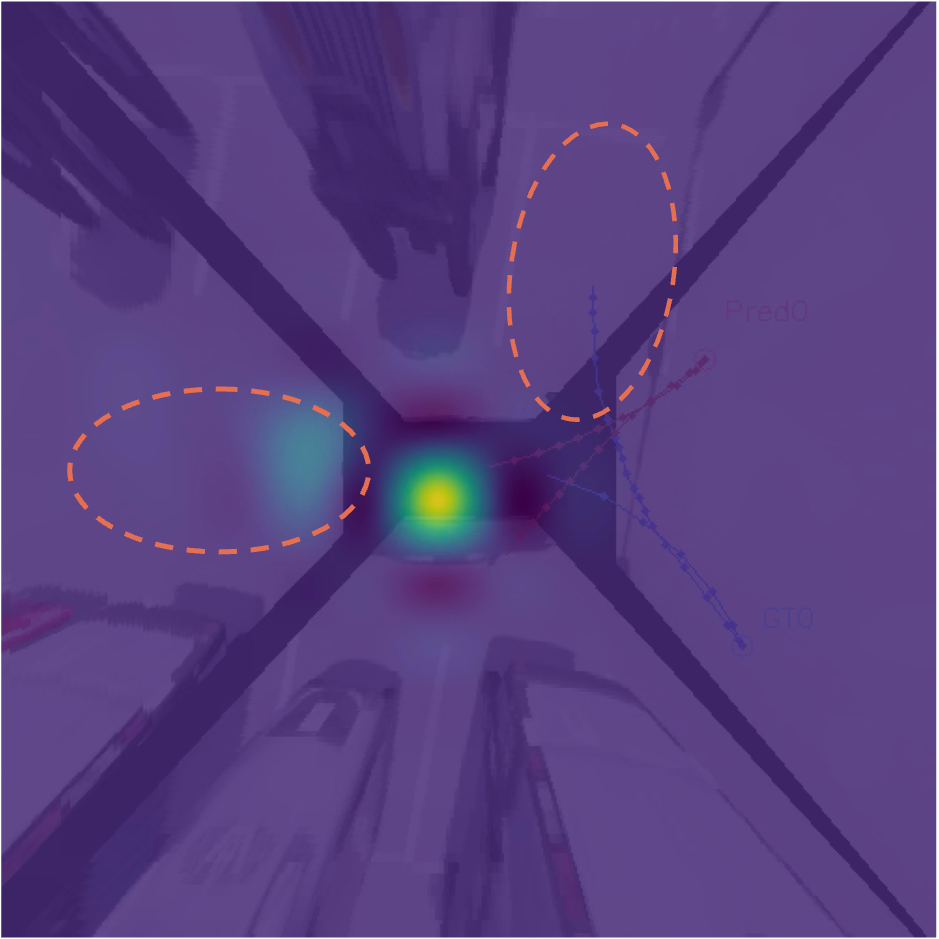}%
        \subcaption{}%
      \end{subfigure}%
    }%
    \edef\hmapcapband{\the\dimexpr\ht2+\dp2-\ht0\relax}%
    \addtocounter{subfigure}{-1}%
    \begin{tabular}{@{}c@{\hspace{3pt}}c@{\hspace{2pt}}c@{}}
        \HeatmapRowLabel{ParkingE2E} &
        \begin{subfigure}[b]{0.42\linewidth}
            \includegraphics[width=\linewidth]{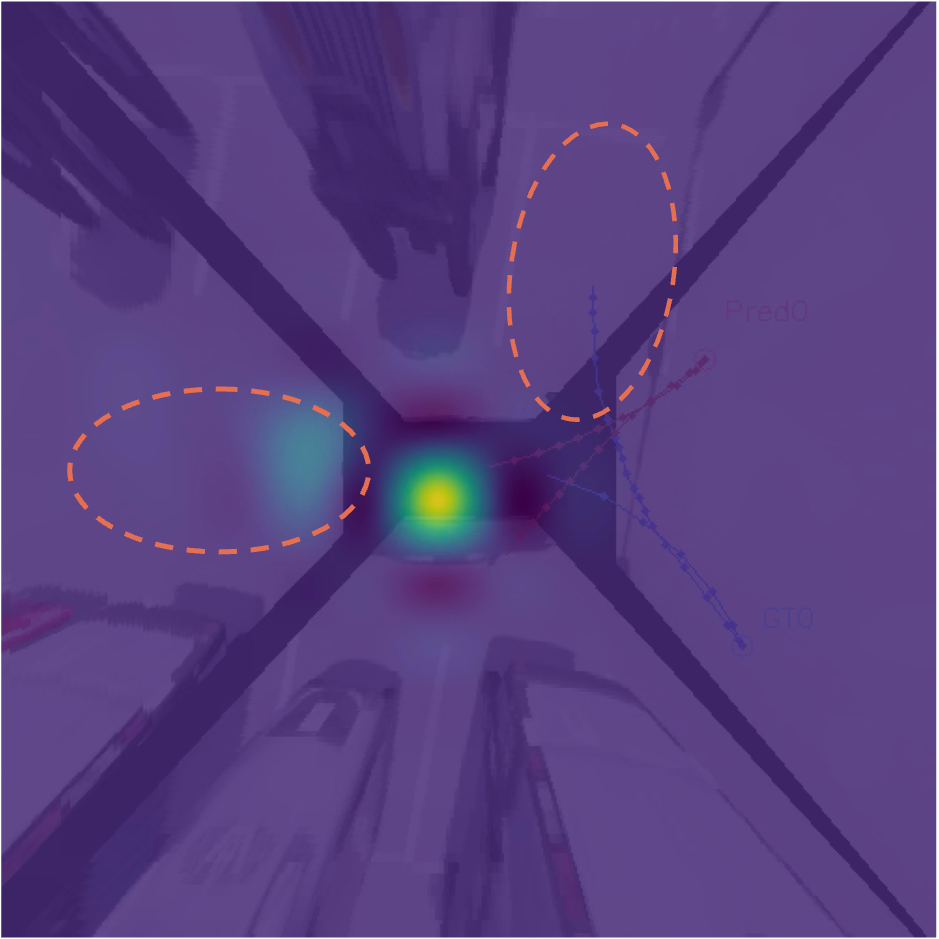}
            \subcaption{}
            \label{fig:heatmap_baseline_a}
        \end{subfigure} &
        \begin{subfigure}[b]{0.42\linewidth}
            \includegraphics[width=\linewidth]{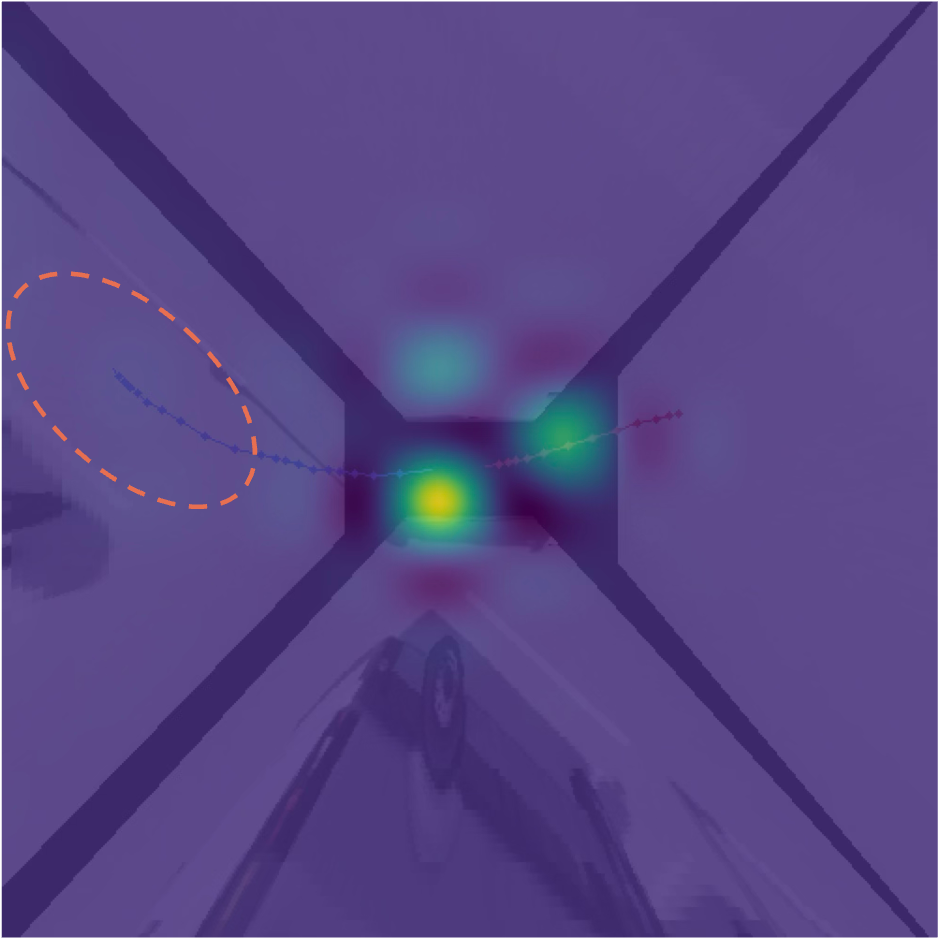}
            \subcaption{}
            \label{fig:heatmap_baseline_b}
        \end{subfigure} \\[0.5pt]
        \HeatmapRowLabel{Ours} &
        \begin{subfigure}[b]{0.42\linewidth}
            \includegraphics[width=\linewidth]{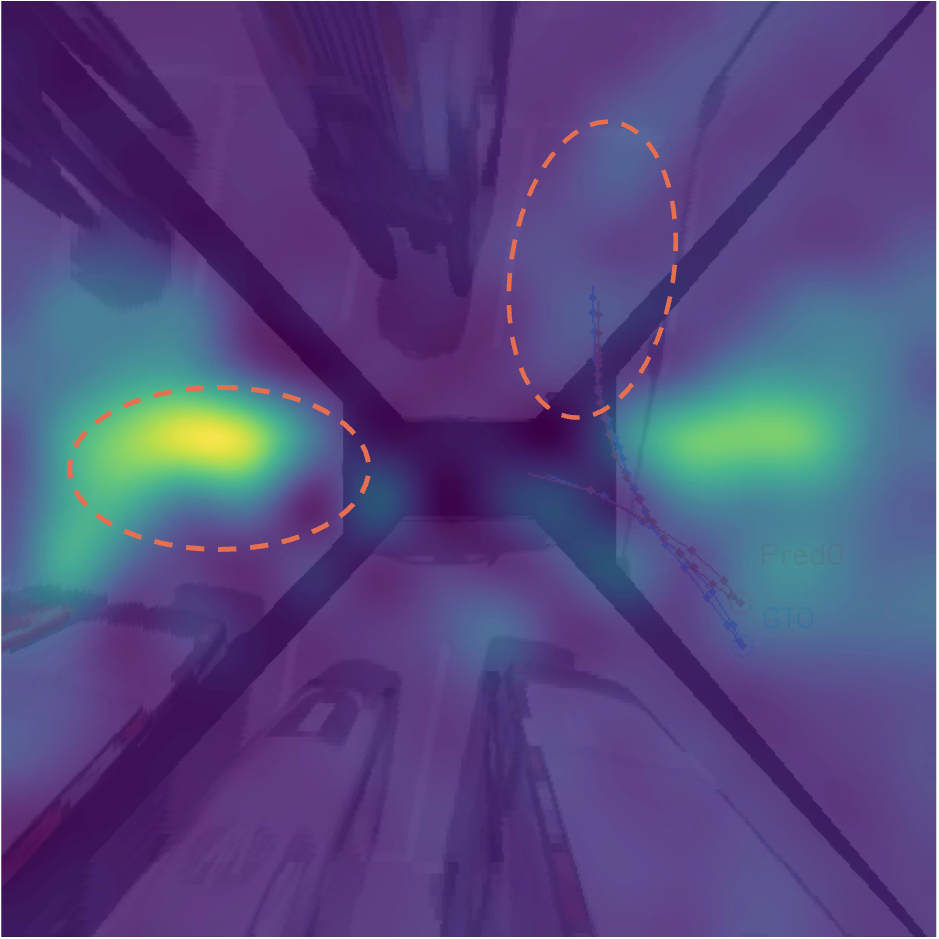}
            \subcaption{}
            \label{fig:heatmap_ours_a}
        \end{subfigure} &
        \begin{subfigure}[b]{0.42\linewidth}
            \includegraphics[width=\linewidth]{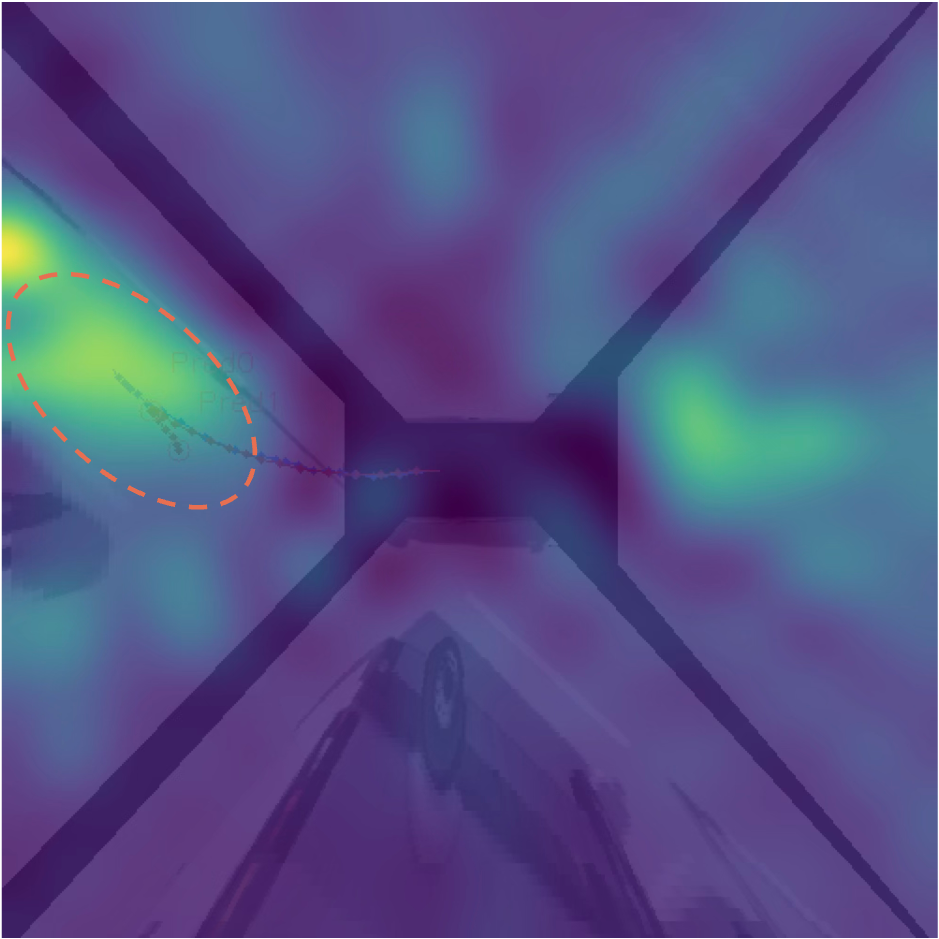}
            \subcaption{}
            \label{fig:heatmap_ours_b}
        \end{subfigure}
    \end{tabular}%
}{\textbf{Comparison of Attention Heatmaps between ParkingE2E (BEV-based method) and Ours (Fourier-based method).
    (\subref{fig:heatmap_baseline_a}) and (\subref{fig:heatmap_ours_a}) show a rear-slot facing configuration, where the ego vehicle is oriented horizontally toward the target slot;
    (\subref{fig:heatmap_baseline_b}) and (\subref{fig:heatmap_ours_b}) show a rear-slot perpendicular configuration, where the ego rear is already aligned with the target slot.
    Our method exhibits an attention shift as the relative position between the ego rear and the target slot changes.}\label{fig:heatmap_viz}}

\subsubsection{Best Combination Validation}

Table~\ref{tab:scheduler_fusion_ablation} stages the selection one factor at a time.
Sets 1--3 fix additive fusion and turn auxiliary losses off, and vary the schedule.
Sets 3--5 fix the linear schedule, still with auxiliary losses off, and vary the infusion operator.
Sets 6--7 keep that linear schedule and gated infusion, fix the endpoint loss, and vary the motion loss.

Under additive fusion, the linear scheduler is the strongest of the three, so the later sets adopt it.
With the schedule fixed, concatenation is not uniformly better than addition: heading comes down, while displacement does not.
Gated fusion is the operator that improves the two together; gear-shift count precision remains slightly higher under addition.
This superiority stems from the gate's ability to adaptively modulate fusion complexity.
By interpolating between a straightforward additive residual and a fully parameterized projection,
it enables fine-grained, per-channel and per-location feature interaction while preserving the foundational spatial priors.
On this choice, the endpoint loss alone (Set~6) sharply lowers trajectory error and its variance.
Adding the motion loss (Set~7) further improves trajectory and gear metrics, with only a negligible drop in the strictest transition hit rate ($\tau{=}1$), and is the setting retained for the full SunnyParking model. 

\subsubsection{Leave-One-Out Component Contribution}

To evaluate the contribution of each proposed module, 
we conduct a leave-one-out ablation study using the full SunnyParking model as the baseline. 
As shown in Table~\ref{tab:ablation_controlled_traj} and Table~\ref{tab:ablation_controlled_gear}, 
removing any single component degrades trajectory or motion metrics under the full-model setting.

The motion state branch remains necessary even when overall trajectory metrics look competitive. 
Without it (\textit{w/o mot. state}), ADE stays low at $1.06$, yet this masks a failure at gear-shift points: 
count recall drops from $0.88$ to $0.33$, and gear-shift point errors in complex maneuvers (e.g., 4-shot) become severely degraded. 
Without explicit motion modeling, the model shows substantially weaker kinematic consistency around directional switches.

For target representation, replacing Fourier encoding with a BEV heatmap (\textit{BEV target}) raises mFHE from $4.6$ to $34.4$. 
The losses serve as regularizers under the full model: omitting endpoint losses (\textit{w/o FDE+FHE}) damages terminal accuracy; 
removing the smoothing loss (\textit{w/o Smooth}) increases variance and yields irregular paths; 
discarding the transition loss (\textit{w/o Trans.}) leaves trajectory and motion-state metrics nearly unchanged, 
yet gear-shift errors increase (GS-ADE $1.07 \rightarrow 1.11$\,m; 4-shot $P_1$ $0.87 \rightarrow 0.98$\,m; count F1 $0.85 \rightarrow 0.84$). 
Training without the scheduling strategy (\textit{w/o sched.}) yields high variance and fails to jointly maintain geometric trajectory accuracy and motion state consistency.

Each omitted component hurts a distinct axis of performance. 
Fourier encoding, the motion state branch, the full auxiliary loss set, and the linear scheduler together yield the best trade-off between geometric trajectory accuracy, kinematic state prediction, and gear-shift consistency. 

\subsection{Qualitative Analysis} \label{sec:qualitative_analysis}

We evaluate our model by visualizing predicted trajectories and gear-shift points across scenarios of varying complexity.
Despite being trained exclusively under ``Clear Noon'' environments without interactive agents, our model is evaluated on challenging unseen scenarios with moving pedestrians and vehicles, under adverse conditions such as ``Cloudy Sunset'', ``HardRain Sunset'', and ``Clear Night''. 

As shown in Fig.~\ref{fig:benchmarking_traj_gear_viz}, 
our method exhibits greater robustness than the baselines as scenario complexity increases, maintaining accurate gear-shift prediction while preserving trajectory accuracy.
In simpler cases, the model may still plan multi-shift maneuvers.
In parking tasks, the expert maneuver represents only one of many feasible solutions. 
Rather than strictly overfitting to the ground truth, our model successfully captures the underlying human-like pattern, 
suggesting robustness under the tested qualitative OOD scenarios.

Fig.~\ref{fig:dynamic_obstacle_weather_viz} jointly probes two OOD axes: photometric variation and interactive agents.

\textbf{Visual Robustness:} Holding the parking layout fixed and varying only weather and lighting illustrates robustness to visual domain shifts.
Variations in weather and lighting drastically alter the scene's photometric properties captured by RGB cameras, 
posing a well-known challenge for neural networks. 
SunnyParking mitigates these domain shifts through two key factors: 
First, a recent benchmarking study \cite{xie2025benchmarking} indicates that LSS-based BEV encoders inherently resist photometric perturbations. 
Once visual features are projected into the BEV space, 
the domain gap presented to the downstream planner is substantially minimized. 
Second, the parking task is strictly governed by drivable areas, target slot, and vehicle kinematics. 
This tightly bounds the well-defined trajectory distribution, 
making the planner less sensitive to raw texture variations. 
Comparing the ``Clear Noon'' references with their OOD counterparts (e.g., 
Figs.~\ref{fig:dynamic_obstacle_weather_i} vs. \ref{fig:dynamic_obstacle_weather_j}--\ref{fig:dynamic_obstacle_weather_l}) 
shows that cloudy and rain shifts barely perturb the planned path, 
whereas poor nighttime illumination occasionally degrades trajectory quality.  

\textbf{Parking Tests with Dynamic Obstacles:}
Dynamic obstacles are introduced exclusively during evaluation: pedestrians cross access aisles and NPC vehicles occupy corridors.
Such reactive avoidance scenarios are absent from the training demonstrations.
The planner operates strictly in a single-frame manner, lifting visual features into $\mathcal{F}_{\mathrm{BEV}}$ without temporal history or trajectory forecasting.

While the model generally maintains coherent rollouts toward the target slot, failure cases emerge during reverse maneuvers.
When a reverse parking trajectory is targeted, the learned parking prior dominates instantaneous occupancy cues.
As shown in Fig.~\ref{fig:dynamic_obstacle_weather_b} and Fig.~\ref{fig:dynamic_obstacle_weather_h}, the planner fails to temporarily detour around or yield to dynamic obstacles situated behind the ego-vehicle, especially under occlusion.
Without avoidance supervision or temporal prediction, single-frame BEV occupancy cannot override the strong parking prior under tight reverse kinematics, resulting in trajectories that intersect obstacle boundaries.

\textbf{Attention Mechanism Visualization:} We visualize the attention heatmaps of the Target-Spatial fusion module to understand the mechanisms driving the Fourier-encoding gains. 
As shown in Fig.~\ref{fig:heatmap_viz}, there is a significant attention degradation in the baseline (i.e., ParkingE2E \cite{10801763}) during multi-shot scenarios. 
It frequently suffers from periodic artifacts due to position encoding errors, and in severe cases, collapses into rigid patterns decoupled from the input—essentially ``memorizing'' average dataset positions rather than planning based on the scene. 
In contrast, by explicitly modeling planning intent, our method demonstrates superior active understanding. The model proactively shifts attention to the vehicle's rear (see Fig.~\ref{fig:heatmap_ours_a}) and the target slot (see Fig.~\ref{fig:heatmap_ours_b}) when predicting reverse motions. 
This ensures attention remains strictly aligned with the immediate planning intent throughout the long-term maneuver.

\section{Discussion and Limitations}
\label{sec:limitations}
\textbf{Preference for Multi-Shot Maneuvers:}
While SunnyParking reliably executes long-horizon maneuvers (Sec.~\ref{sec:qualitative_analysis}, Table~\ref{tab:exp_compare}), 
its predicted shot count reflects a conservative human prior rather than geometric optimality. 
This bias stems from the demonstration distribution (Appendix~\ref{appendix:dataset_stats}) 
and from objectives that lack shot-efficiency incentives: 
endpoint metrics (mFDE/mFHE) do not penalize redundant intermediate shifts, 
and the transition loss penalizes missed shifts far more than spurious extra ones. 
Human-style fidelity therefore does not imply shot-count optimality; 
closing the gap would require explicit efficiency objectives or rebalanced demonstrations.

\textbf{Generalization to Dynamic Obstacles:}
Interactive agents are absent from training, and the planner has no temporal model of other traffic participants; collision-aware parking under this OOD regime is therefore outside the present claims.
Explicit agent prediction or a world model~\cite{Feng2025ASO, 10.1007/978-3-031-73195-2_4, zheng2024occworld, yang2025driving} is left for future work (Sec.~\ref{sec:related_generative}).

\section{Conclusion}
In this study, we propose an end-to-end dual-branch trajectory prediction framework. 
By introducing an additional motion state branch, the model significantly enhances its ability to predict transition points (e.g., gear shifts) in multi-shot parking trajectories. 
We employ Fourier feature mapping to encode target points into query vectors interacting with BEV features for cross-domain information fusion. 
Experimental results demonstrate that our dual-branch model exhibits superior trajectory handling in complex parking scenarios while effectively capturing human expert styles. 
In particular, unlike conventional methods that degrade on complex multi-shot maneuvers, our framework sustains robust long-horizon gear-shift prediction with kinematic fidelity. 
We plan to extend our study to more complex real-world scenarios and improve semantic understanding of unconstrained environments.


%

\appendices
\section{CARLA Human-driven Parking Dataset Details}
\label{appendix:dataset}

To create a comprehensive and challenging dataset, we generate a series of parking scenarios within the \texttt{Town04\_Opt} environment in CARLA 0.9.13. 
The generation process is twofold: populating the environment with static vehicles and systematically placing the ego vehicle at diverse initial poses.

\Figure[bp][width=0.99\linewidth]{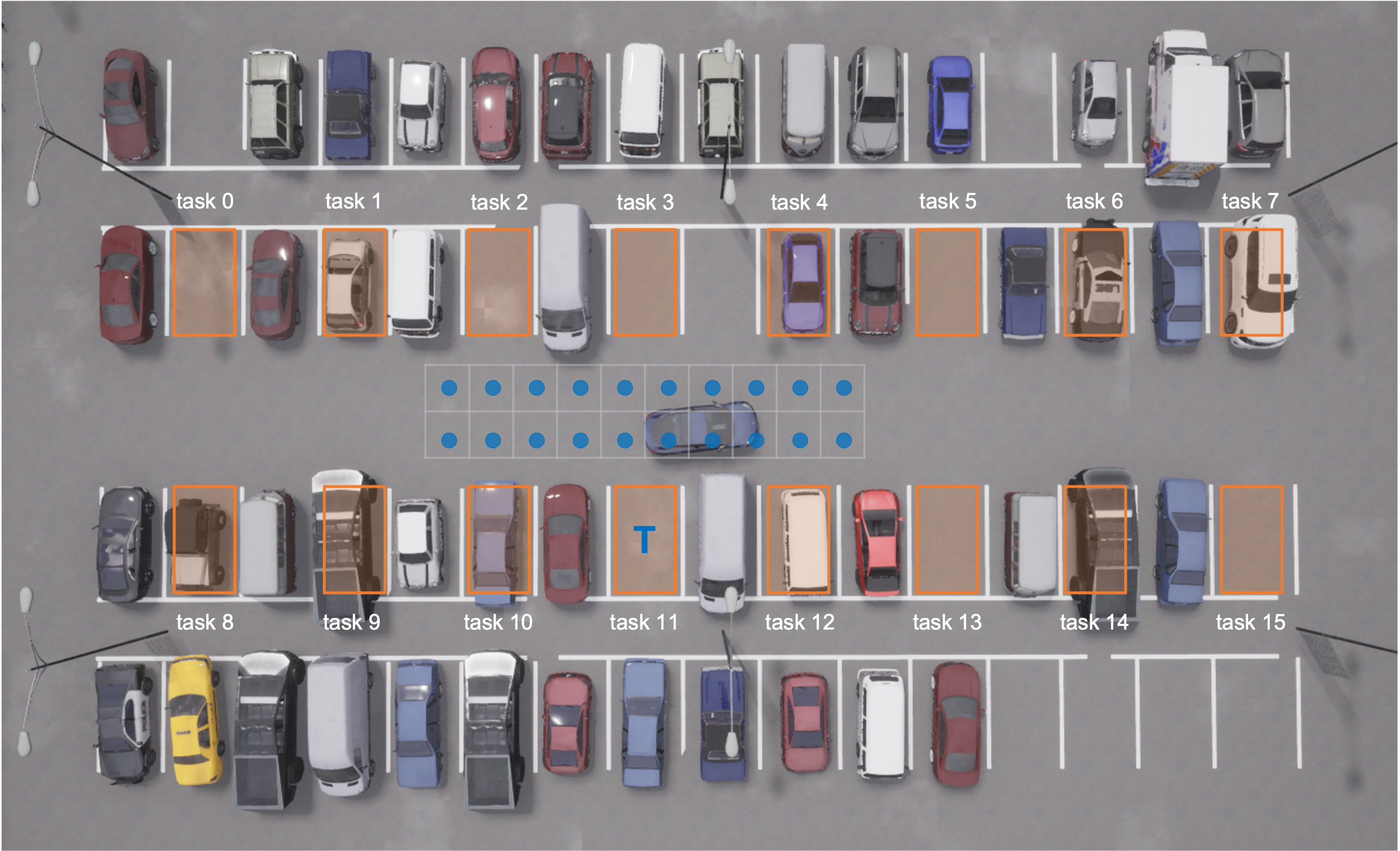}{\textbf{Data Collection Scenario of the Parking Lot in BEV. The orange boxes represent the 16 target slots, 
while the blue spots are the start points for a specific task T (i.e., ``task 11''). 
The ego vehicle pose is determined by (\ref{eq:spawn_pose}).} \label{fig:dataset_grid}}

\subsection{Static Environment Generation}
As shown in Fig.~\ref{fig:dataset_grid}, for each of the 16 designated target slots, we populate the remaining parking spaces with static vehicles to simulate a realistic parking lot environment. 
The pose $(x, y, \psi)$ of each static vehicle is randomized as follows:

\begin{itemize}
    \item \textbf{Base Orientation ($\psi_{\text{base}}$):} We simulate the two most common parking patterns for the static vehicles on perpendicular slots. The base orientation is randomly set to either $0^\circ$ (head-in parking) or $180^\circ$ (back-in parking).
    \item \textbf{Yaw Jitter ($\delta_{\psi}$):} A small random yaw offset is added to simulate imperfect parking, drawn from a clipped normal distribution over $[-8^\circ, 8^\circ]$. 
\end{itemize}
The final heading of a static vehicle is $\psi_{\text{vehicle}} = \psi_{\text{base}} + \delta_{\psi}$. 

\subsection{Ego Vehicle Initial Pose Generation}
To ensure comprehensive coverage of starting positions around each target slot, we devise a systematic generation strategy combining a deterministic grid with stochastic jitter. 

The base pose $(x_{\text{base}} , y_{\text{base}} , \psi_{\text{base}} )$ is positioned relative to the target slot's entrance, ensuring the vehicle starts on the main access lane. 
Its yaw is set perpendicular to the lane. 
A deterministic 2D grid $\mathcal{G}$ is defined to systematically sample start points covering a wide area in front of the target slot. 
An offset $(x_{\text{offset}}, y_{\text{offset}})$ is selected from this grid for each scenario: 
\begin{align}
    x_{\text{offset}} & \in \{-x_{r}, -x_{r} + x_{s}, \ldots, x_{r}\},  \\
    y_{\text{offset}} & \in \{-y_{r}, -y_{r} + y_{s}, \ldots, y_{r}\}, 
\end{align}
where $x_r=1$\,m, and $y_r=10$\,m, with step sizes $x_{s}=1$\,m, and $y_{s}=2$\,m. 
To guarantee each spawn pose is unique and to introduce fine-grained variations, a stochastic jitter term is applied to both position and orientation: $x_{\text{jitter}}, y_{\text{jitter}} \sim \mathcal{U}(-0.2, 0.2)\,\mathrm{m}$ and $\psi_{\text{jitter}} \sim \mathcal{U}(-15^\circ, 15^\circ)$.

The final spawn pose is thus the sum of these components, creating a diverse yet structured distribution of initial conditions for robust model training and evaluation.
\begin{equation}
\begin{aligned}
    x_{\text{spawn}} &= x_{\text{base}} + x_{\text{offset}} + x_{\text{jitter}},  \\
    y_{\text{spawn}} &= y_{\text{base}} + y_{\text{offset}} + y_{\text{jitter}},  \\
    \psi_{\text{spawn}} &= \psi_{\text{base}} + \psi_{\text{jitter}}. 
\label{eq:spawn_pose}
\end{aligned}
\end{equation}

\subsection{Data Acquisition}

\Figure[bp][width=0.9\linewidth]{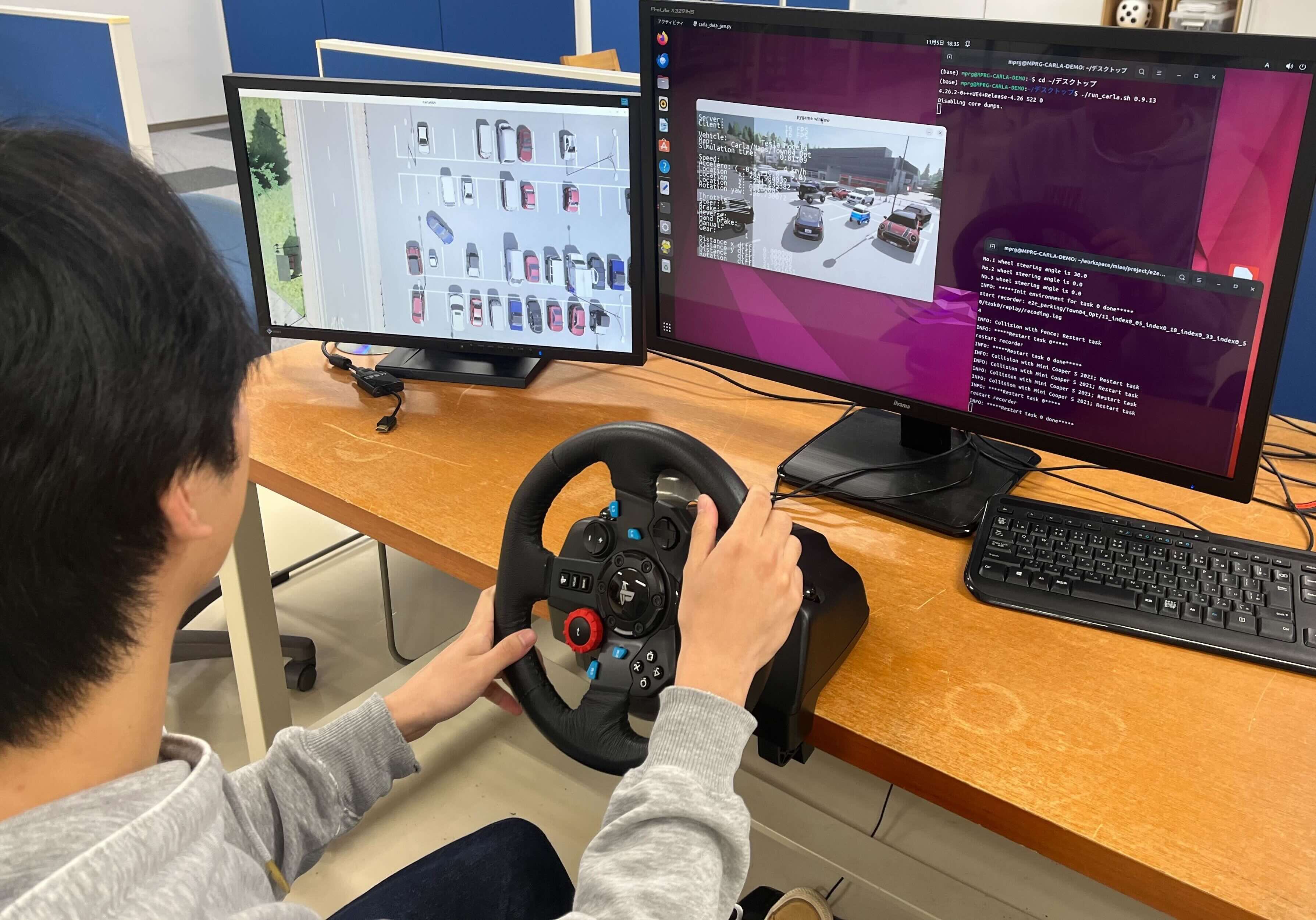}{\textbf{A human driver is performing a parking maneuver.} \label{fig:hattori}}

Nine drivers performed parking maneuvers using the Logitech G29 controller to operate the steering wheel, gearbox, throttle, and brake pedals. 
This hardware configuration produces smooth, continuous control signals closer to real-vehicle actuation than those from stepwise computer peripherals. 

We configured an operational steering limit of $30^{\circ}$ to mirror the soft-limit strategy of real-world chassis controllers. 
This design aligns the simulation with physical engineering practices, ensuring the generated parking maneuvers faithfully replicate real-vehicle behaviors. 

We set the criteria for successful parking as the Euclidean distance being less than $0.25$ m and the heading angle being less than $2.5^\circ$. 
These thresholds are dataset inclusion criteria rather than evaluation thresholds and therefore do not directly enter the metrics reported in Table~\ref{tab:exp_compare}.
These limits align with the ISO 20900~\cite{iso20900_2023} standard for automated parking systems and are stricter than established configurations \cite{10588551} (Euclidean distance $<0.4$ m and heading angle $<5^\circ$). 
Multimodal data are synchronously collected during park-in task execution at a $5$ Hz frequency.
When processing large-scale visual data, both feature selection and data acquisition strategies are critical to extract meaningful information efficiently \cite{bolon2020feature}.
Parking is a low-speed maneuver, so consecutive surround-view frames at a higher rate are nearly identical and add little new visual evidence.
Kabir~\emph{et~al.}~\cite{kabir2023machine} similarly emphasize controlled image acquisition for stable CNN-based visual estimation; our $5$\,Hz rate limits redundant frames for the same reason.
The collected modalities specifically comprise:

\begin{itemize}
    \item Perception Data: Four surround-view RGB camera images and corresponding depth maps. 
    \item Ego-Vehicle State: High-precision vehicle positioning data (including position $x, y$ and orientation $\psi$), alongside detailed chassis signals (e.g., speed, acceleration, throttle, brake, and gear status).
    \item Task Information: Precise position and orientation of the target parking slot. 
    \item Environment Data: Category, position, orientation, 
    and 3D bounding box information for surrounding static obstacles (e.g., other vehicles). 
\end{itemize}

\begin{table}[tbp]
\centering
\caption{Consolidated Sensor Specifications}
\label{tab:sensor_and_camera_specs}
\begin{tabular}{@{}l  c l l @{}}
\toprule
\textbf{Sensor} & \textbf{Count} & \textbf{Resolution} & \textbf{FOV} \\
\midrule
RGB Camera & $4$ & $400 \times 300$ px & $100^{\circ}$ \\

Depth Camera & $4$ & $400 \times 300$ px & $100^{\circ}$ \\
\midrule

\multirow{2}{*}{IMU} & \multirow{2}{*}{$1$} & \multicolumn{2}{l}{Accelerometer (3-axis),     }  \\
                     		&              		      & \multicolumn{2}{l}{Gyroscope (3-axis), Compass}  \\
\midrule
GNSS                 	&          $1$         & \multicolumn{2}{l}{Provides Latitude, Longitude} \\
\midrule
\multirow{2}{*}{Vehicle State} & \multirow{2}{*}{--} & \multicolumn{2}{l}{Position, Orientation, Speed,} \\
                     		&         	           & \multicolumn{2}{l}{Throttle, Brake, Steer, Gearbox} \\

\bottomrule
\end{tabular}
\end{table}

\subsection{Data Analysis} \label{subsec:data_analysis}
\Figure[bp][width=0.99\linewidth]{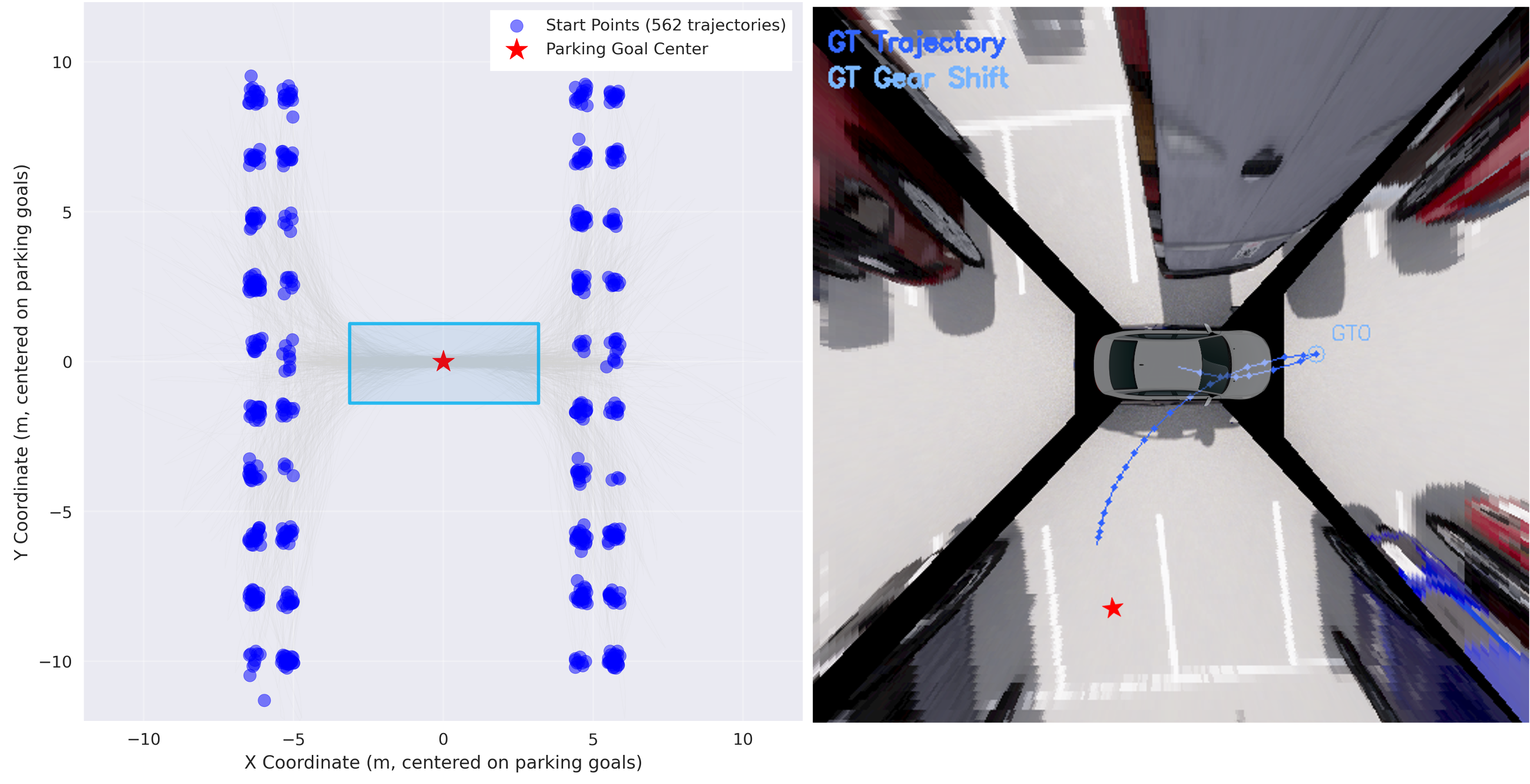}{\textbf{Distribution Map of Parking Start Point. 
All target parking slots are normalized to the center of the image. It can be seen that for parking on one side, there are 20 start points (2 columns by 10 rows), 
accompanied by a certain amount of jitter.} \label{fig:dataset_distribution}}

\FigureContentWide[htbp]{%
    \begin{subfigure}{\textwidth}
        \centering
        \includegraphics[width=0.28\textwidth]{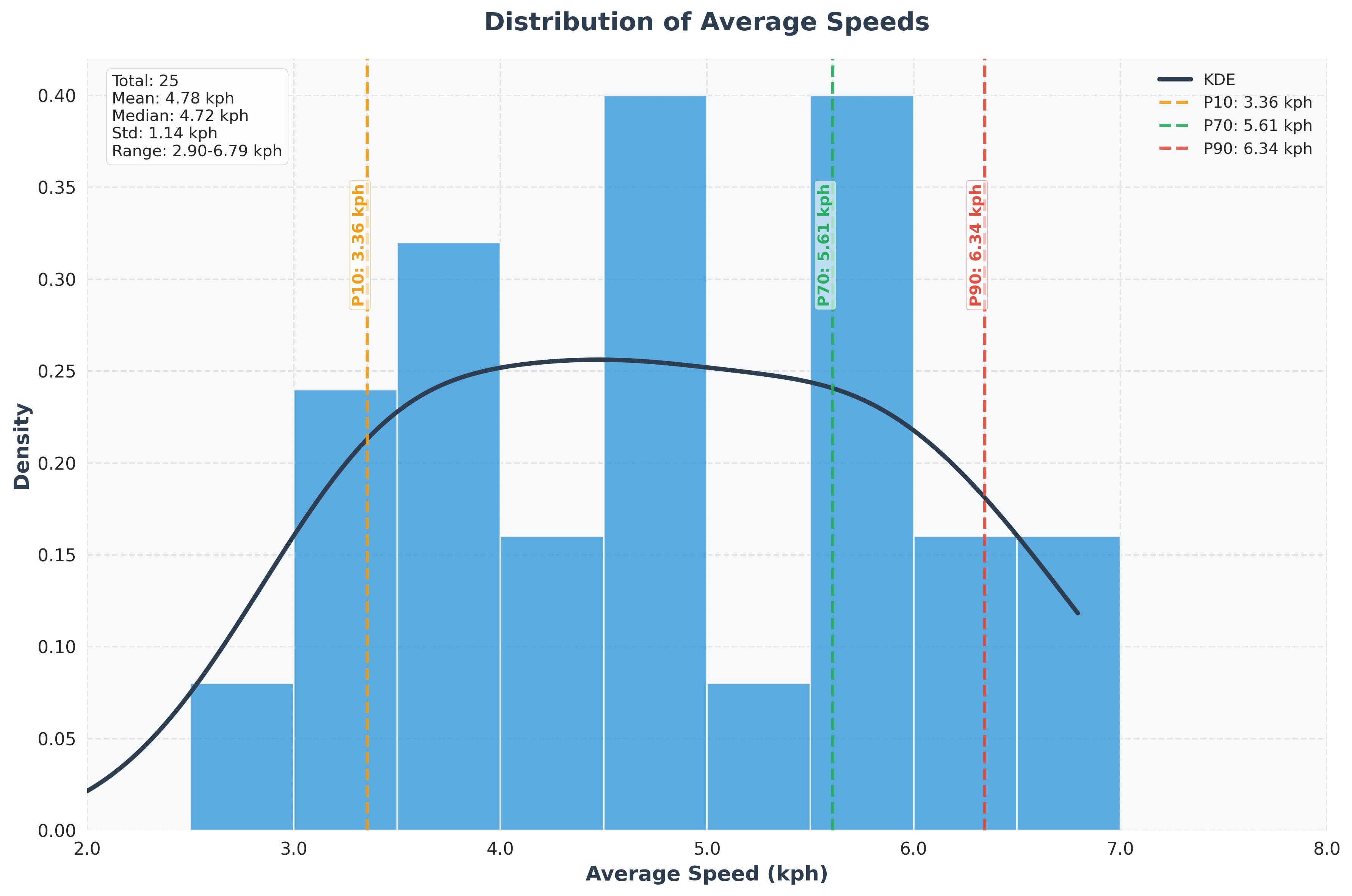}
        \hspace{\fill}
        \includegraphics[width=0.28\textwidth]{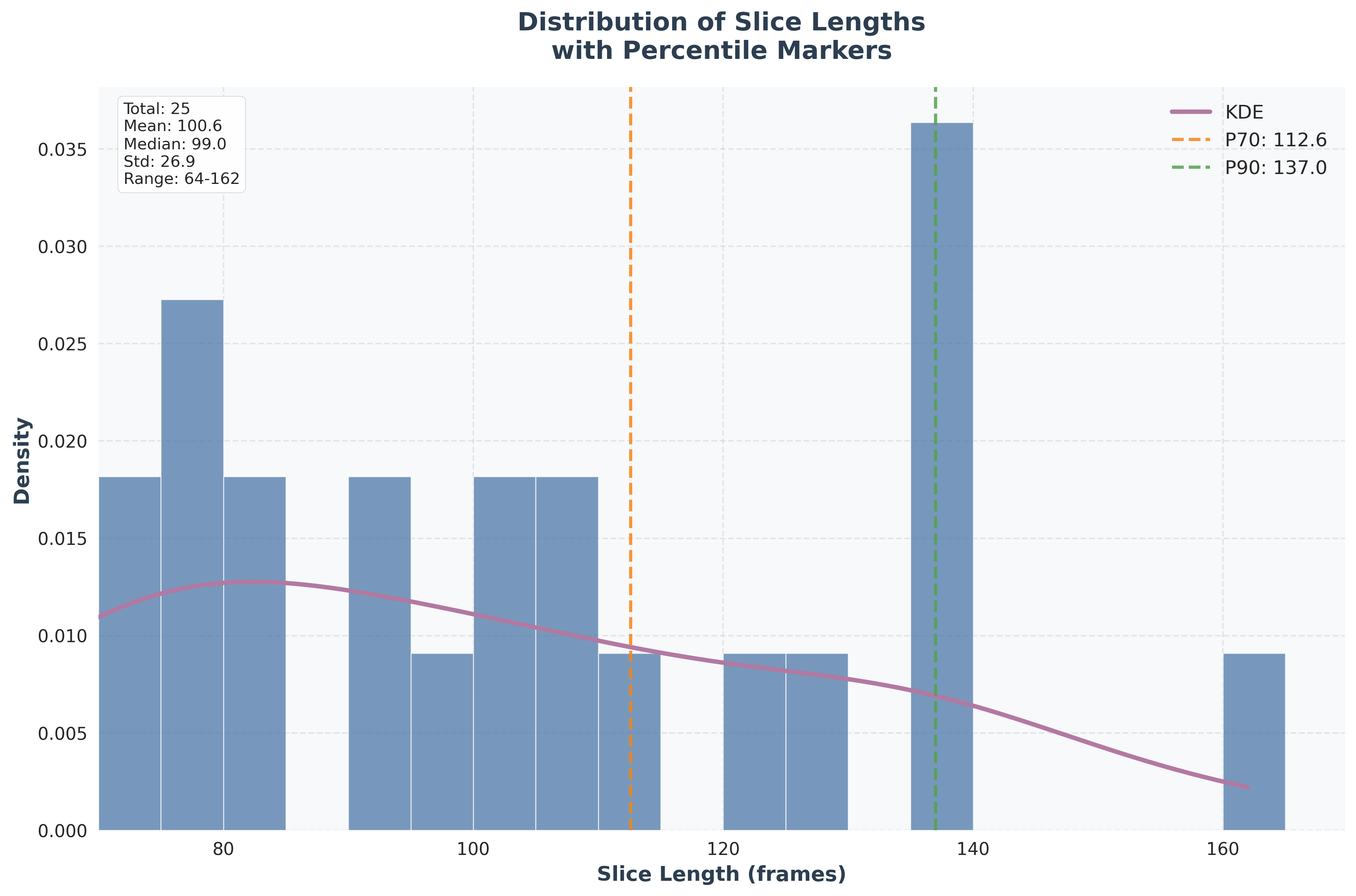}
        \hspace{\fill}
        \includegraphics[width=0.28\textwidth]{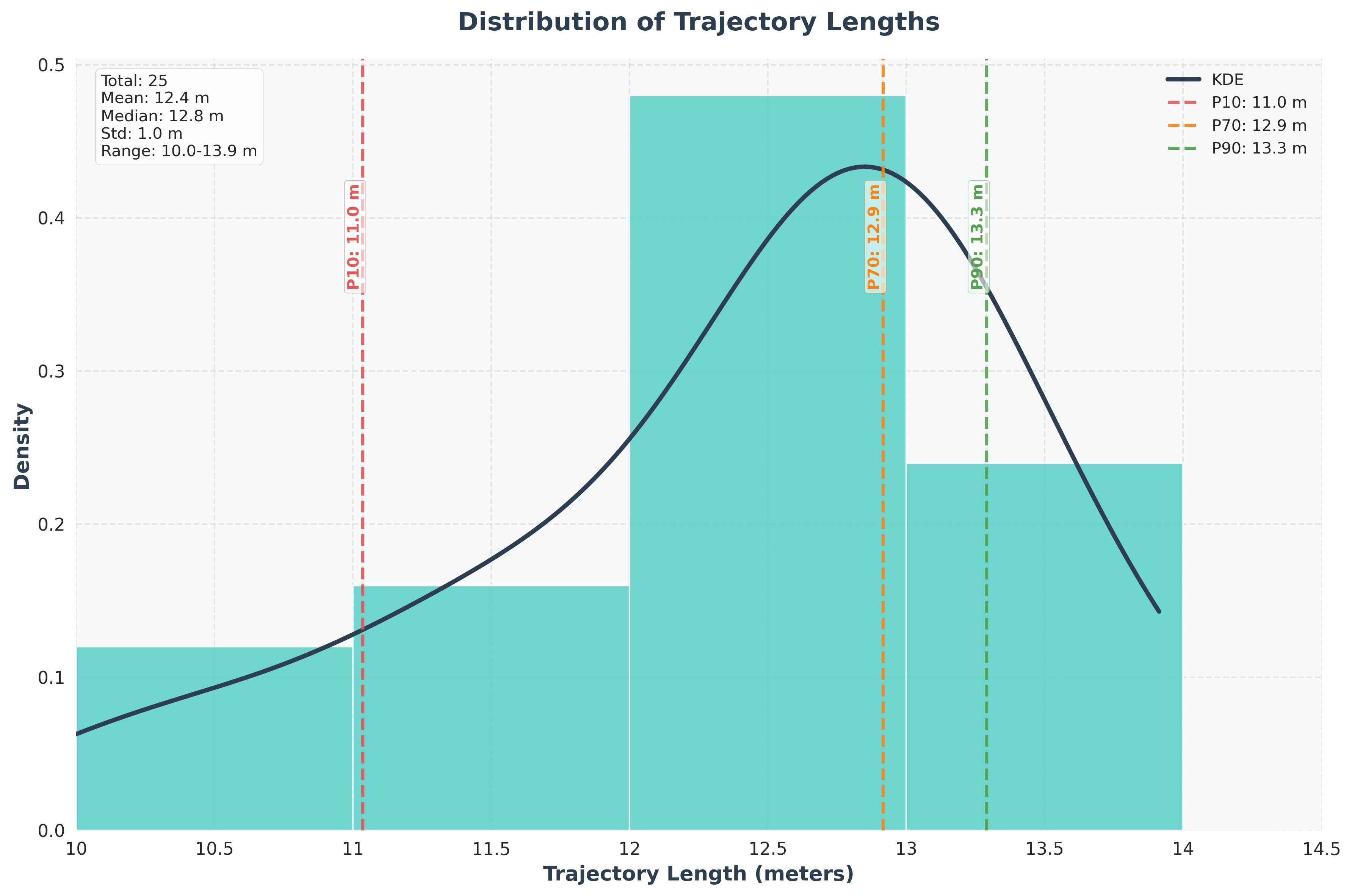}
        \caption{$1$-shot: 0 Gear Shifts}
        \label{fig:group_gear_0}
    \end{subfigure}

    \begin{subfigure}{\textwidth}
        \centering
        \includegraphics[width=0.28\textwidth]{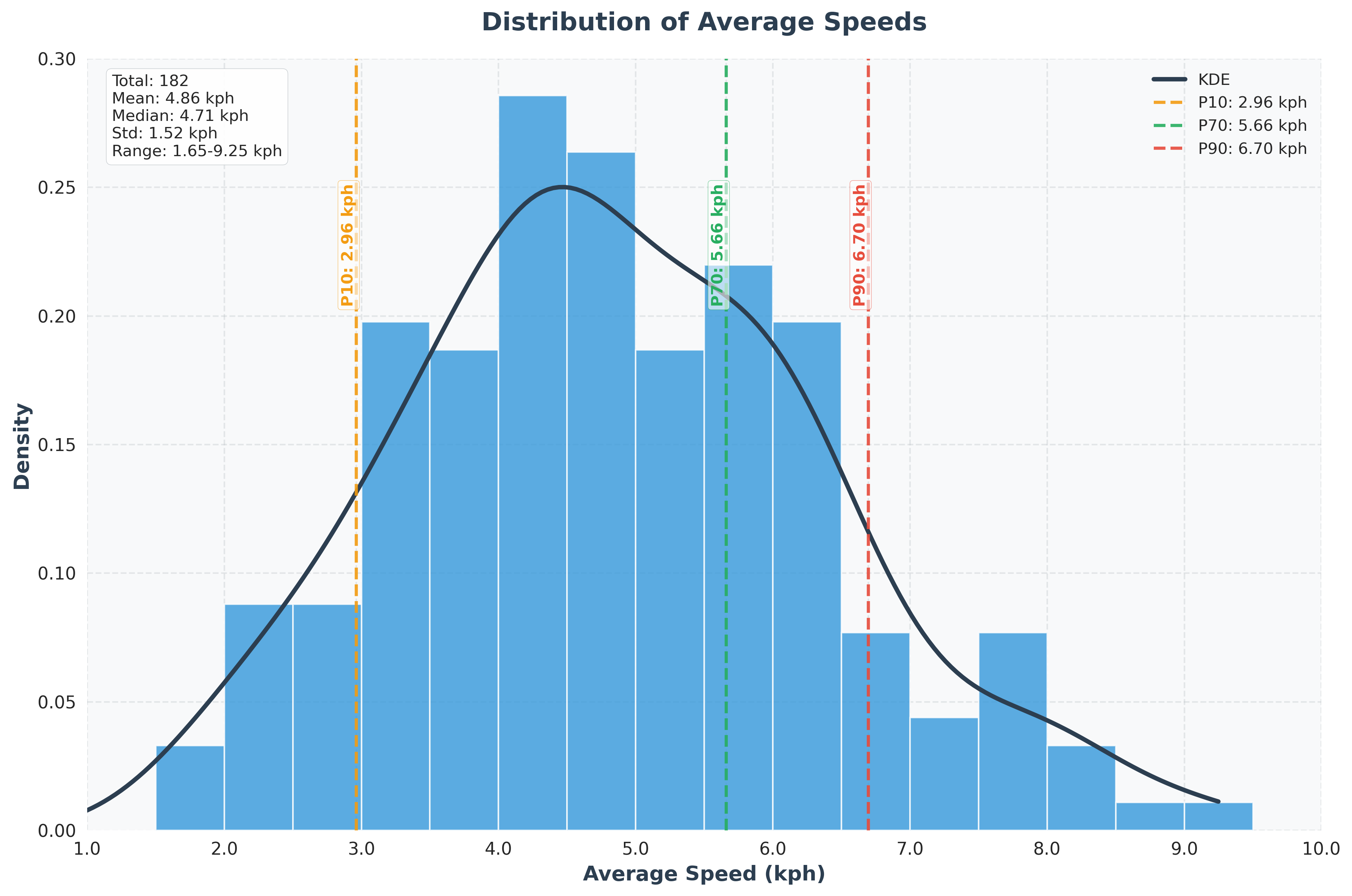}
        \hspace{\fill}
        \includegraphics[width=0.28\textwidth]{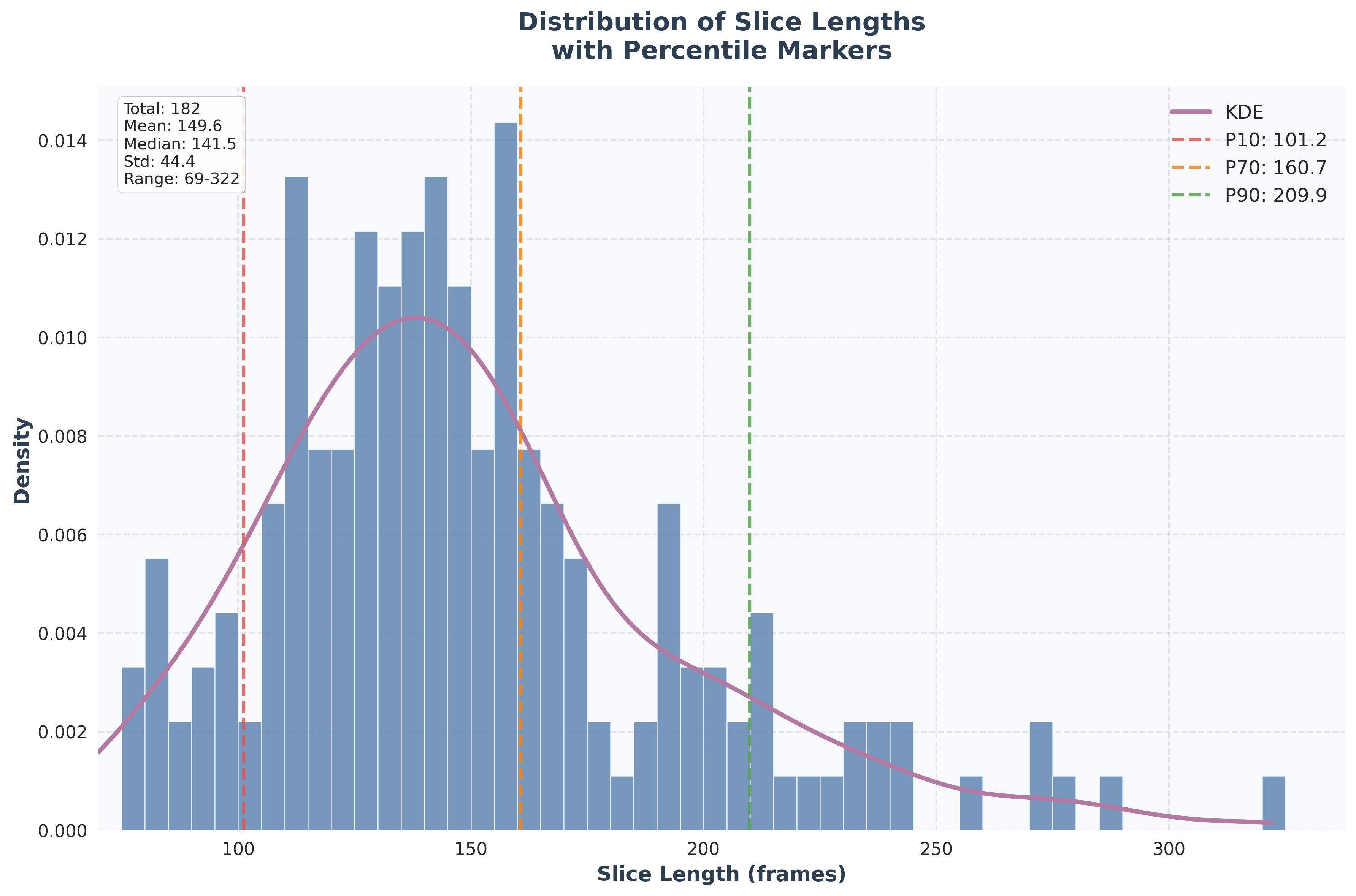}
        \hspace{\fill}
        \includegraphics[width=0.28\textwidth]{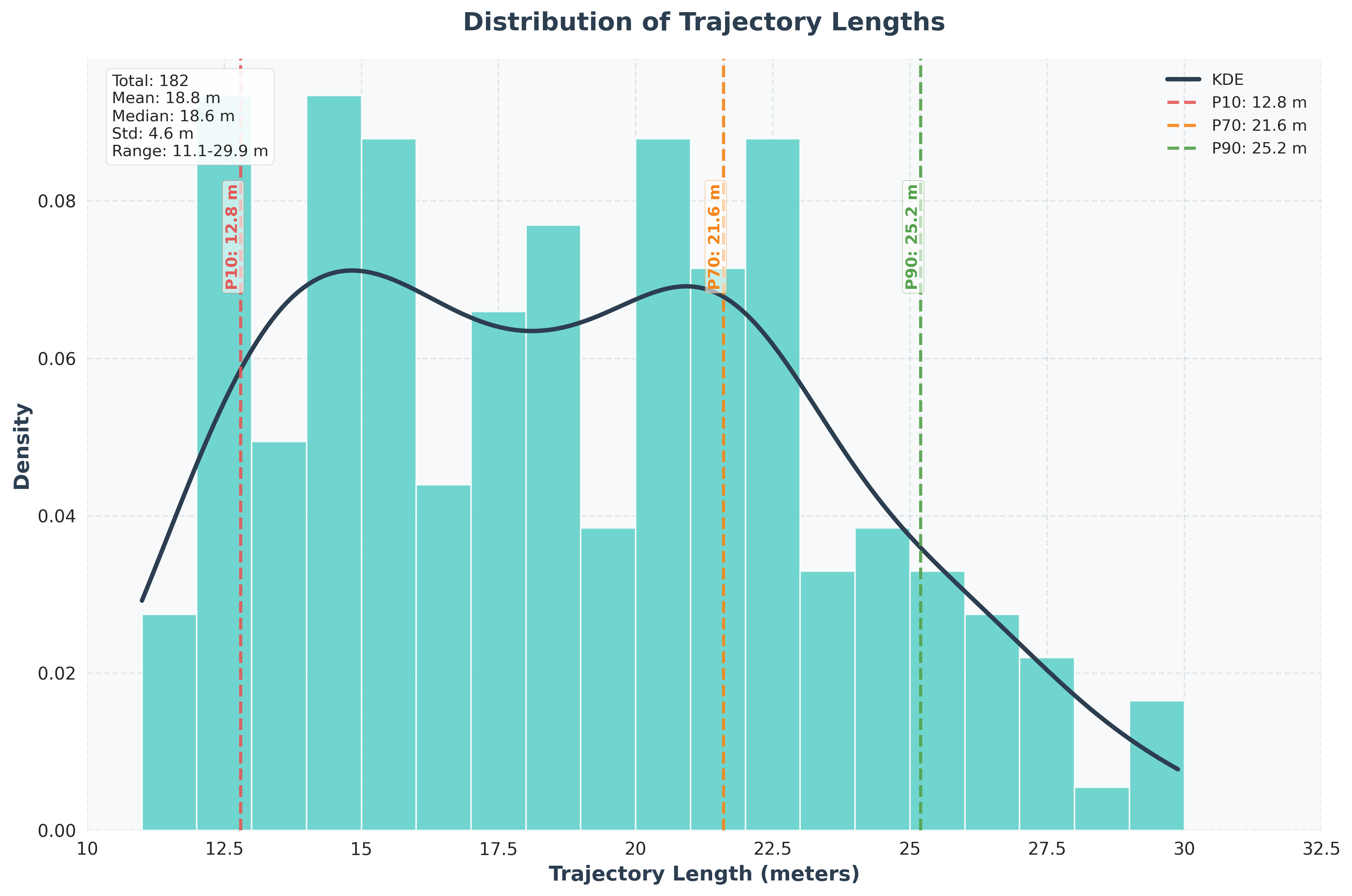}
        \caption{$2$-shot: 1 Gear Shift}
        \label{fig:group_gear_1}
    \end{subfigure}

    \begin{subfigure}{\textwidth}
        \centering
        \includegraphics[width=0.28\textwidth]{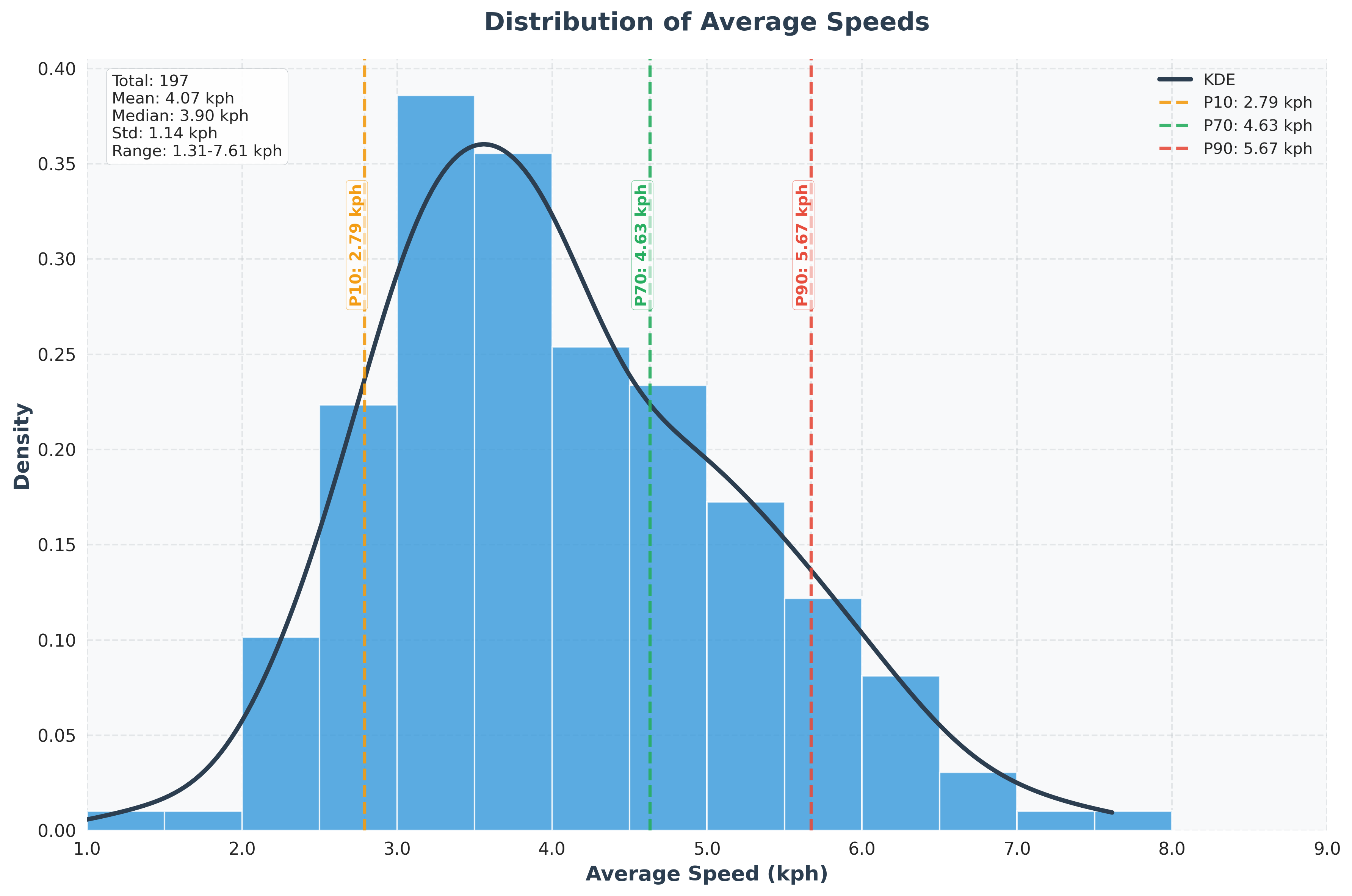}
        \hspace{\fill}
        \includegraphics[width=0.28\textwidth]{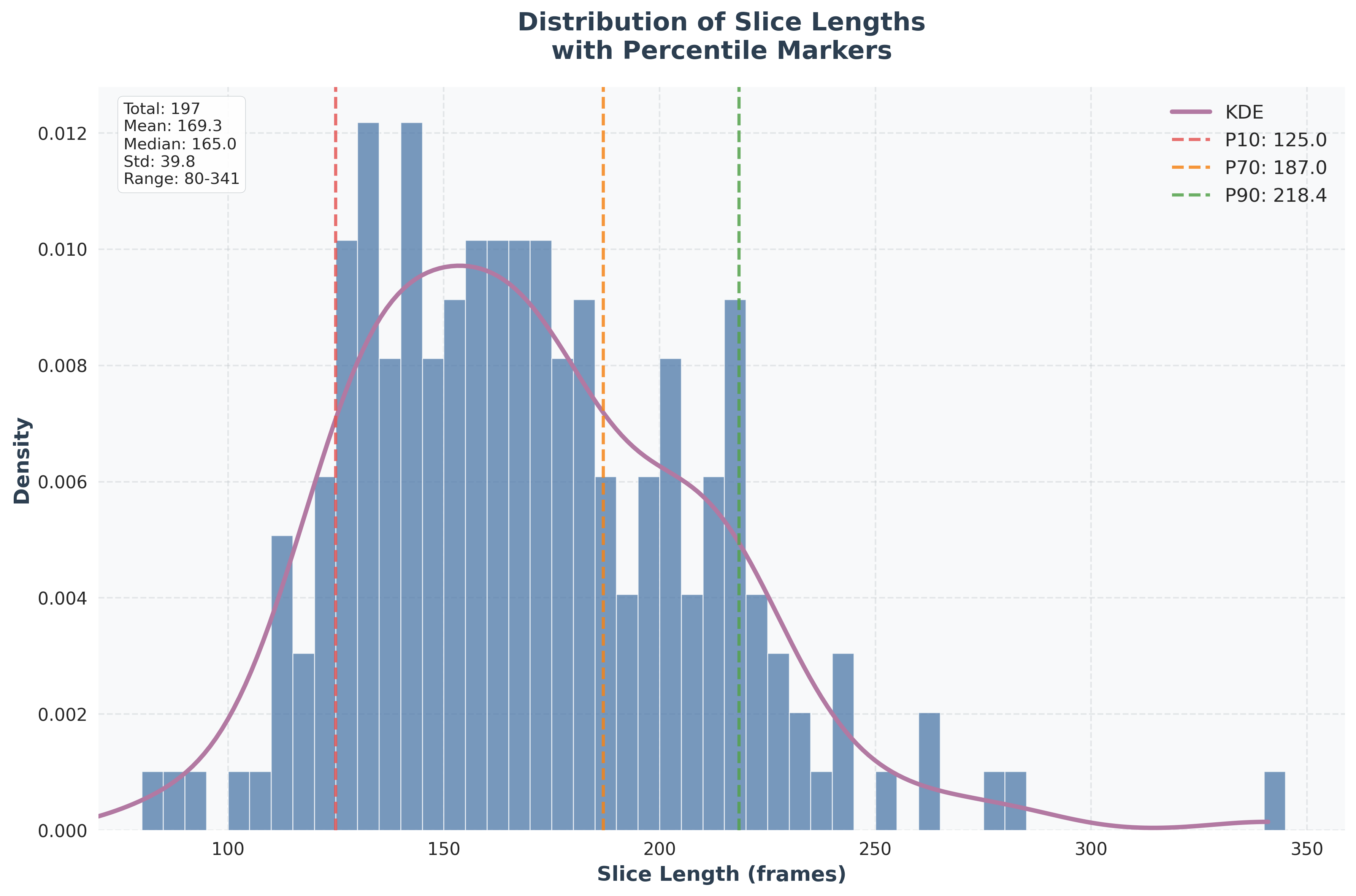}
        \hspace{\fill}
        \includegraphics[width=0.28\textwidth]{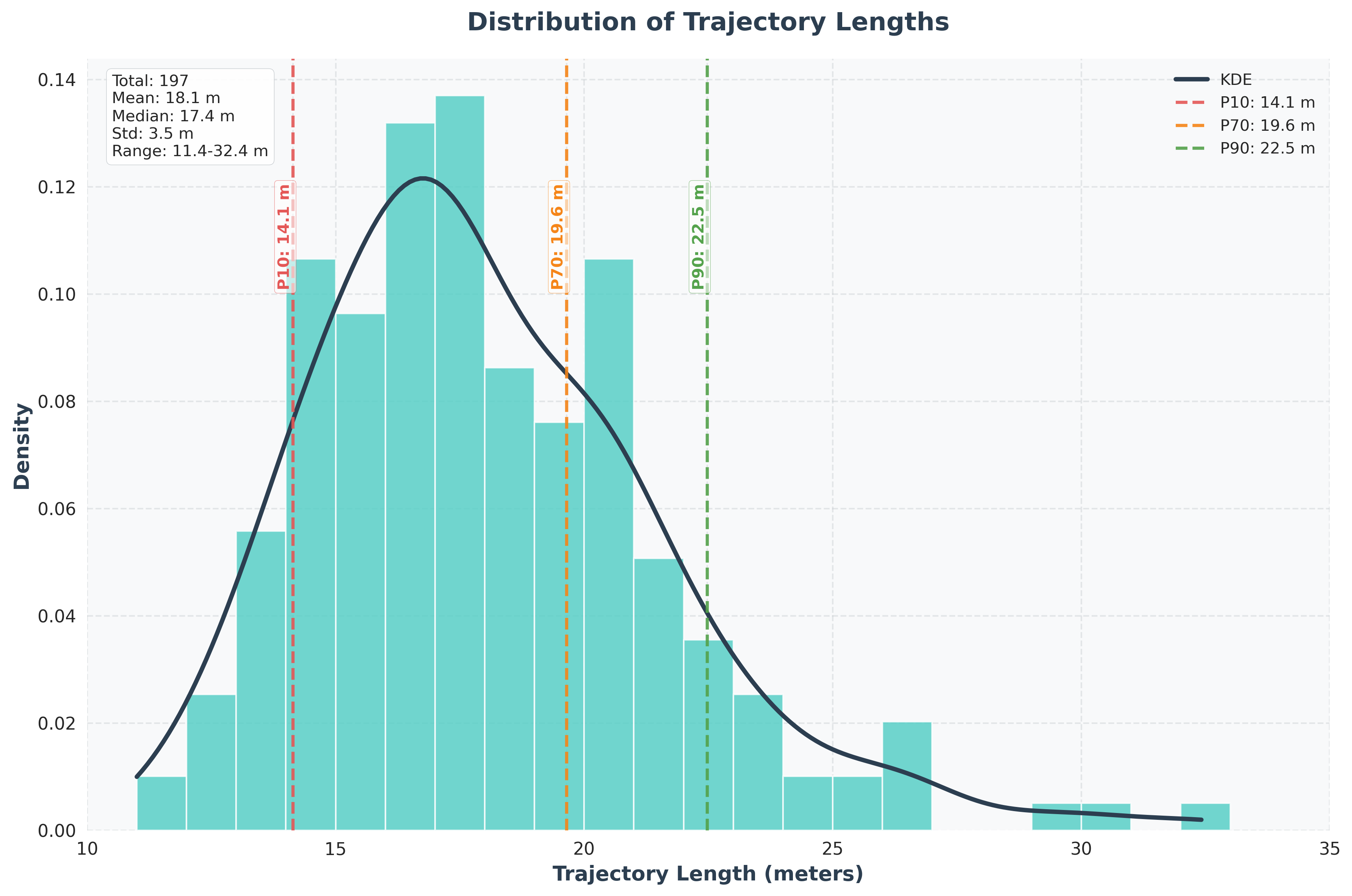}
        \caption{$3$-shot: 2 Gear Shifts}
        \label{fig:group_gear_2}
    \end{subfigure}

    \begin{subfigure}{\textwidth}
        \centering
        \includegraphics[width=0.28\textwidth]{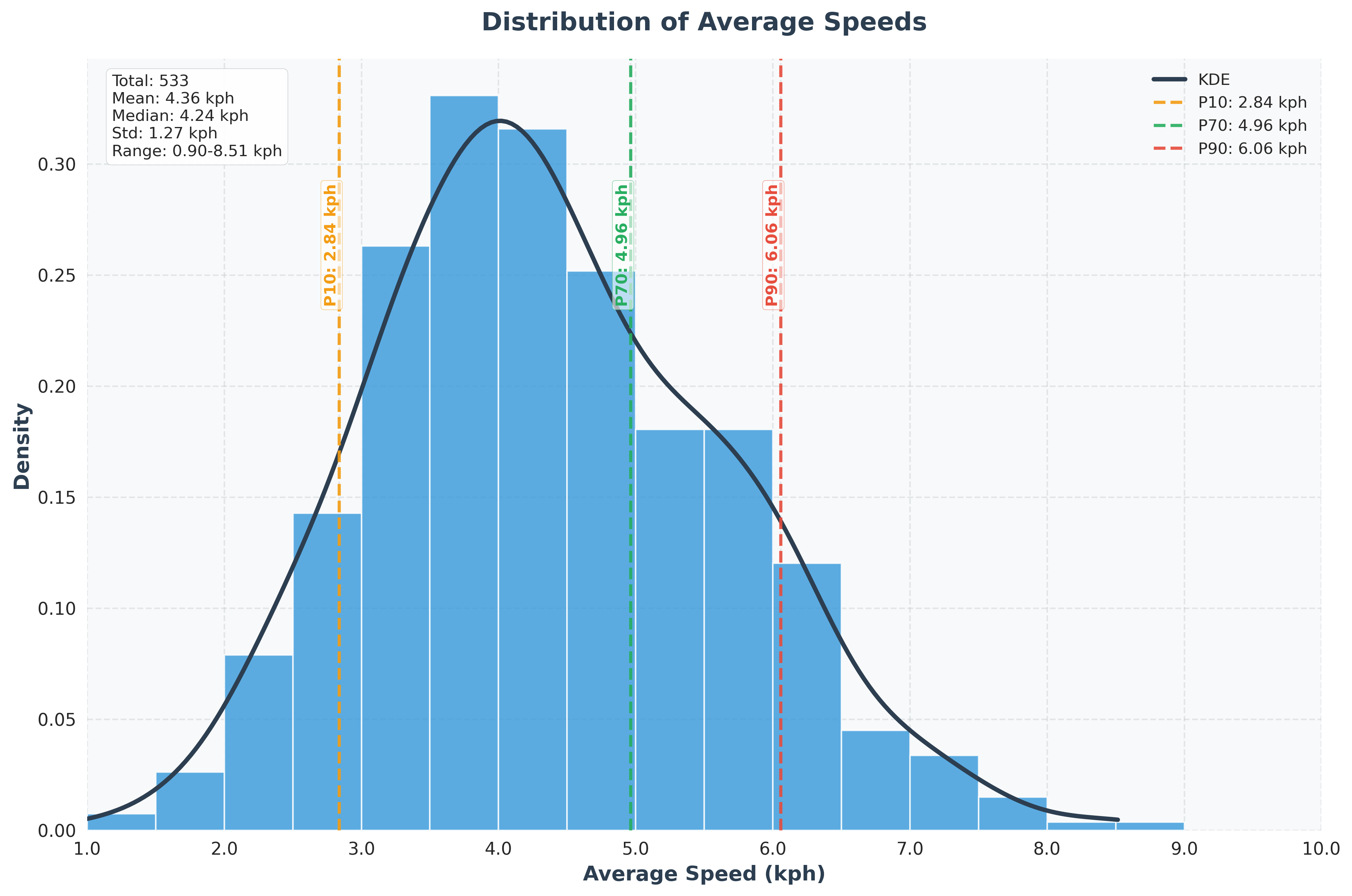}
        \hspace{\fill}
        \includegraphics[width=0.28\textwidth]{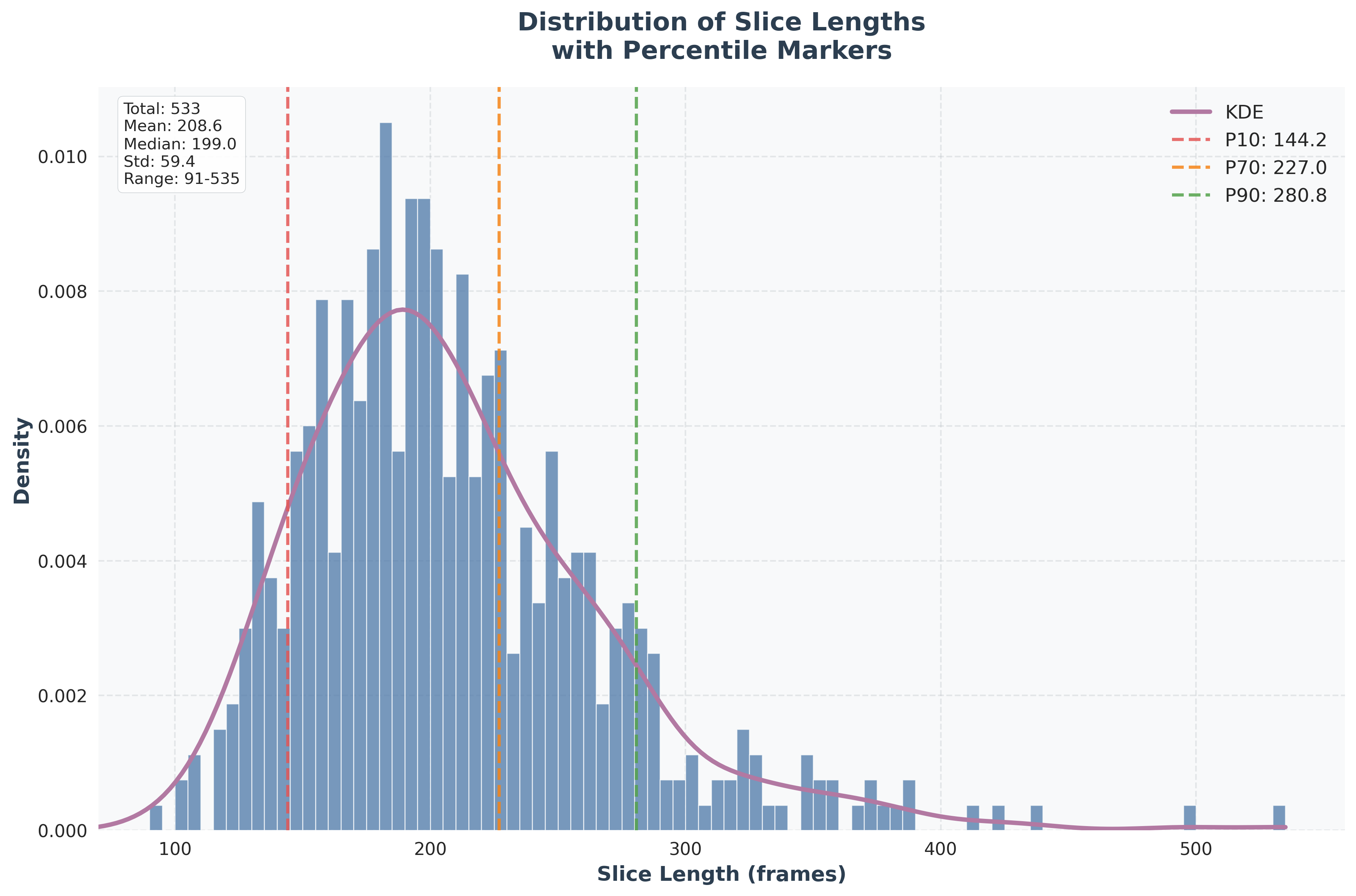}
        \hspace{\fill}
        \includegraphics[width=0.28\textwidth]{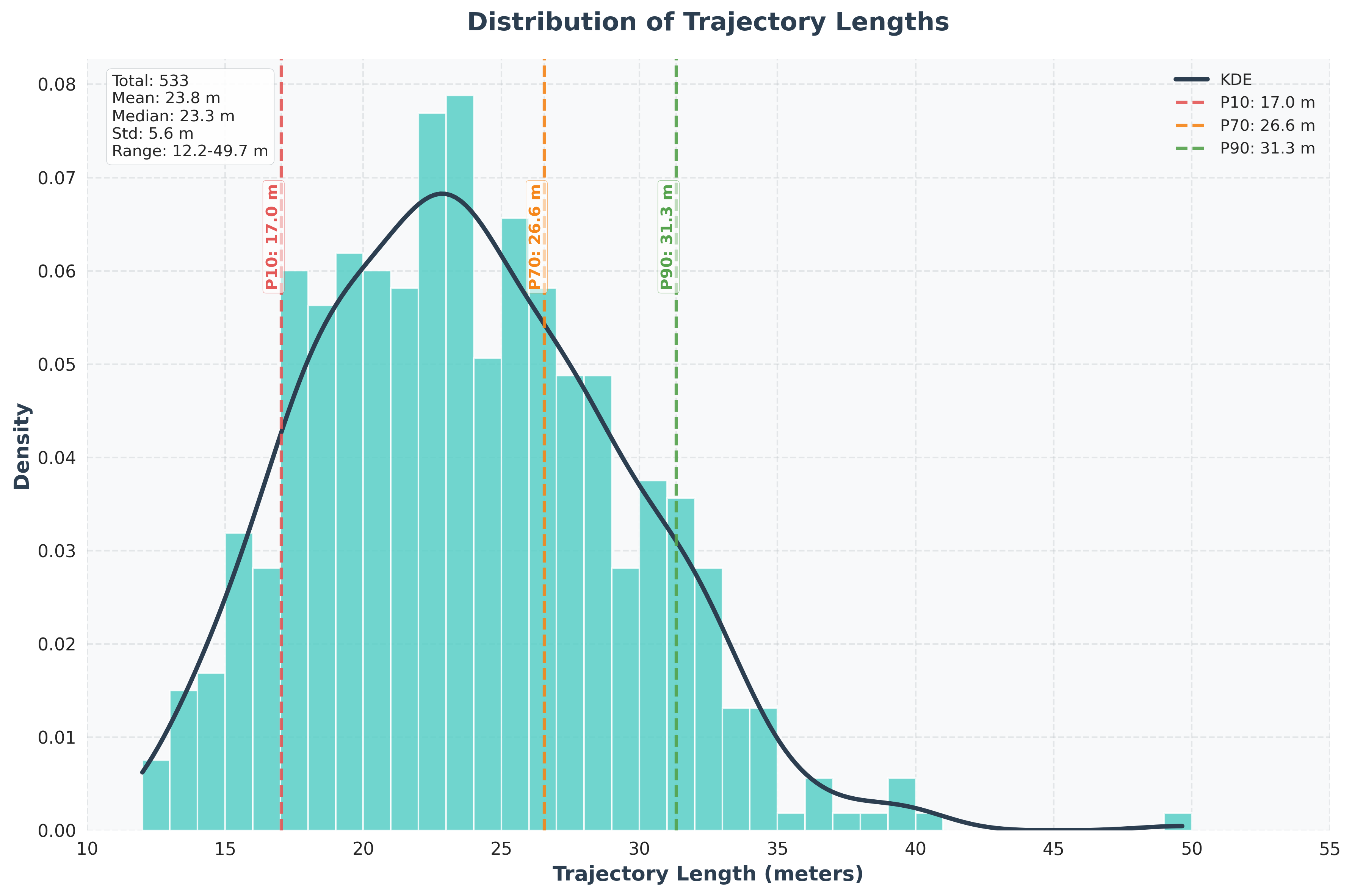}
        \caption{$4$-shot: 3 Gear Shifts}
        \label{fig:group_gear_3}
    \end{subfigure}%
}{Overview of data distributions for core metrics, grouped by the number of gear shifts. From left to right: Average Speed, Slice Length, and Trajectory Length.\label{fig:data_distribution_grid}}

The raw collected data contained significant redundancy and non-standard driving behaviors.
To construct a focused, efficient dataset that aligns with human drivers' actual parking habits and intentions \cite{10830257}, 
we first concentrated the data on parking tasks involving 0 to 3 gear changes. 
Then we defined the \emph{valid slice} of parking tasks: 
the start time is when the vehicle's throttle first produces actual acceleration, and the end time is when the vehicle comes to a complete stop at the target parking slot with a speed of zero, while ignoring instances where the gearbox passes through Neutral during gear changes.
The resulting slice therefore contains only Drive and Reverse segments; near-zero speed at the terminal stop does not introduce a third motion state class. 

Based on this valid slice, subsequent statistical analyses were conducted. 
All trajectory tasks were categorized by gear shift count ($N_{\text{shift}} \in \{0, 1, 2, 3\}$).
Statistical analysis was performed across three dimensions: 
average speed, slice length, and trajectory length. 

As shown in Fig.~\ref{fig:data_distribution_grid}, the original data distribution is extensive. 
To avoid any extreme outliers and obtain the most representative core data, we retained only those trajectories within each category where all three dimensions fell within the $[10\%, 90\%]$ range. 
This formed our core dataset.
We consider zero gear changes ($1$-shot parking) to represent the smoothest and most efficient parking maneuver, serving as an ideal baseline.
The specific interval thresholds applied for each category are as follows:
\begin{table}[htbp]
\centering
\caption{Data filtering thresholds and gear-shift distribution per maneuver.}
\label{tab:filtering_gear_shift}
\resizebox{\columnwidth}{!}{%
\begin{tabular}{@{}l c c c c c c@{}} 
    \toprule
    \multirow{2}{*}{Category} & Avg. Speed & Slice Len.  & Traj. Len. & Maneuvers & Mean & Timestep \\ 
     &  (km/h) &  (frames) &  (m) & (\%) & Timesteps & Share \\ 
    \midrule
    $1$-shot  & \multicolumn{3}{c}{Unconstrained} & 25~(4.45\%) & 62.4 & 1.9\% \\ 
    $2$-shot  & $[2.90, 6.70]$ & $[101, 210]$ & $[12.8, 25.2]$ & 111~(19.75\%) & 107.3 & 14.9\% \\ 
    $3$-shot  & $[2.79, 5.70]$ & $[125, 219]$ & $[14.0, 22.5]$ & 120~(21.35\%) & 130.8 & 19.6\% \\ 
    $4$-shot  & $[2.83, 6.00]$ & $[144, 280]$ & $[17.0, 31.0]$ & 306~(54.45\%) & 166.8 & 63.6\% \\ 
    \midrule
    Total & & & & 562~(100\%) & & 100\% \\ 
\bottomrule
\end{tabular}%
}
\end{table}

\section{Detailed Dataset Statistics and Generalization Analysis}
\label{appendix:dataset_stats}

We characterize the post-filtering scale of SunnyParking, the trajectory-level distribution of gear shifts, 
and the coupling between starting geometry and demonstration style. 
These analyses expose the multi-shot priors governing model training and evaluation.

\subsection{Corpus Scale and Data Split}
Table~\ref{tab:overall_scale} summarizes the dataset scale. 
Each recorded parking maneuver is treated as one sample.
The 562 maneuvers cover all nine drivers, all 16 target slots, and the start-pose grid in (\ref{eq:spawn_pose}), and are randomly partitioned into training and validation sets. 
No maneuver is shared between the two sets.
Validation reports in-distribution performance on held-out maneuvers from the same domain.
Driver identities and target-slot identities are not explicitly held out across the two subsets; therefore, this evaluation should be interpreted as maneuver-level generalization rather than cross-driver or cross-slot generalization. 
Furthermore, static frames at the start and end of a maneuver are pruned during preprocessing, yielding the effective timesteps for supervision.

\begin{table}[tbp]
\centering
\caption{Overall scale of the SunnyParking dataset. Retained timesteps exclude static frames trimmed during preprocessing.}
\label{tab:overall_scale}
\begin{tabular}{lccc}
\toprule
Quantity & Train & Validation & Total \\
\midrule
Trajectories & 507 & 55 & 562 \\
Retained timesteps & 72,319 & 7,845 & 80,164 \\
Raw timesteps (before trim) & 90,692 & 9,674 & 100,366 \\
\bottomrule
\end{tabular}
\end{table}

\subsection{Trajectory-Level Gear-Transition Distribution}
Table~\ref{tab:filtering_gear_shift} details the distribution of gear shifts per trajectory. 
The dataset is heavily dominated by multi-shift maneuvers. 
Specifically, trajectories with 3 shifts (i.e., 4-shot parking) constitute the majority of the dataset both in maneuver count (54.45\%) and total timesteps (63.6\%). 
In contrast, true 1-shot parking maneuvers with no gear shifts are exceedingly rare, 
accounting for only 4.45\% of the maneuvers and a mere 1.9\% of the training frames. 

\subsection{Correlation Between Start Geometry and Maneuver Style}
To investigate whether the scarcity of 1-shot maneuvers is strictly caused by geometrically harder initial states, 
Table~\ref{tab:start_geometry} compares the starting poses across different maneuver styles. 

As shown in Table~\ref{tab:start_geometry}, 
while all maneuvers begin nearly perpendicular to the target slot ($|\Delta\mathrm{yaw}| \approx 90^{\circ}$), 
true 1-shot maneuvers are characterized by starting further away with a larger lateral offset. 
In fact, typical 4-shot trajectories are often necessitated by unfavorable initial poses, even when starting closer to the target.

Among styles, 3-shot starts are closest to the 1-shot envelope. 
Many 3-shot maneuvers thus begin in geometries where a 1-shot path is feasible, 
yet the dataset predominantly provides reverse--forward--reverse templates.

\FigureContentSingle[tbp]{%
    \begin{subfigure}[b]{0.48\linewidth}
        \centering
        \includegraphics[height=5cm]{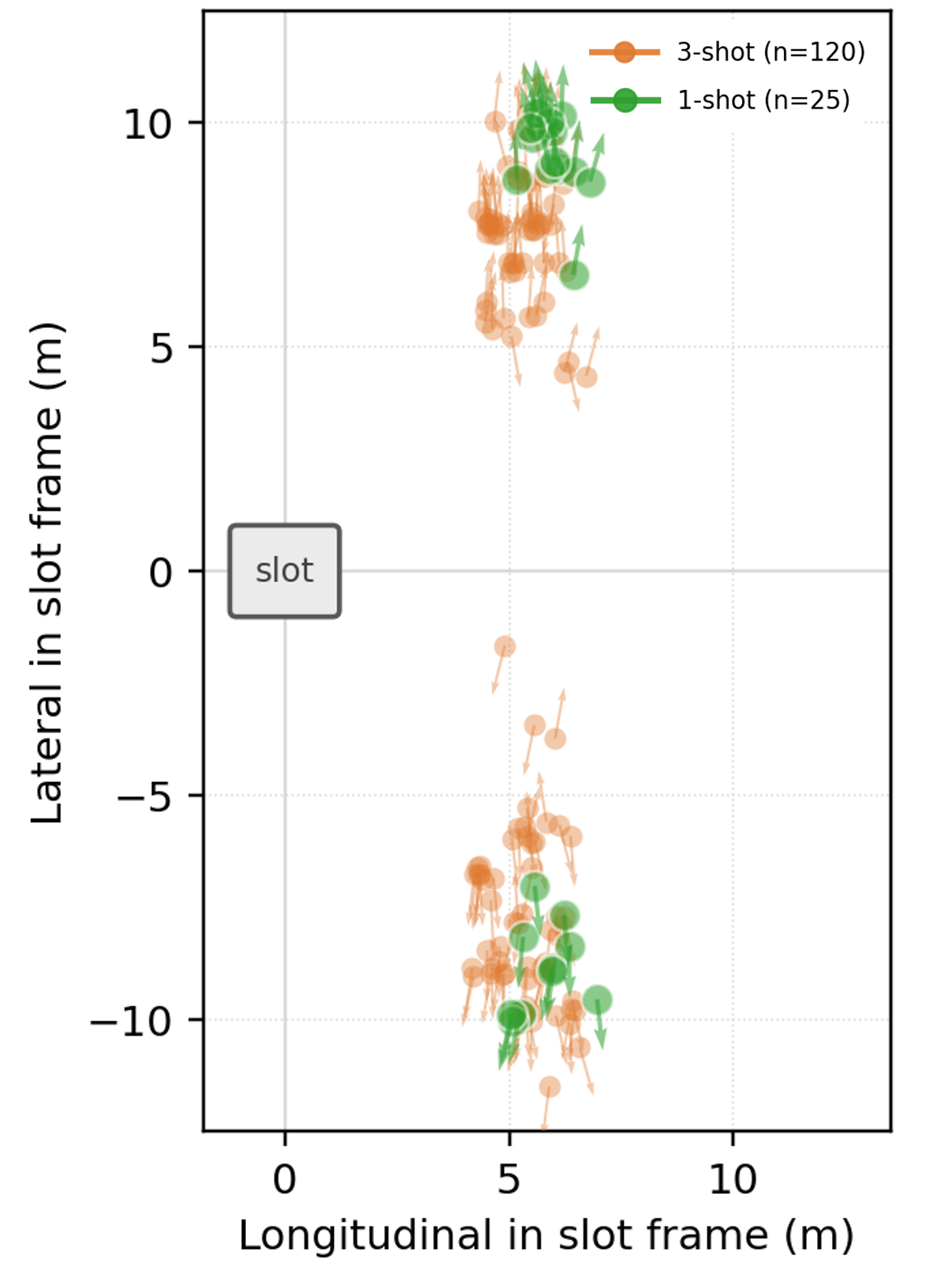}
        \caption{Start Position Distribution}
        \label{fig:start_position_distribution}
    \end{subfigure}%
    \hfill
    \begin{subfigure}[b]{0.48\linewidth}
        \centering
        \includegraphics[height=5cm]{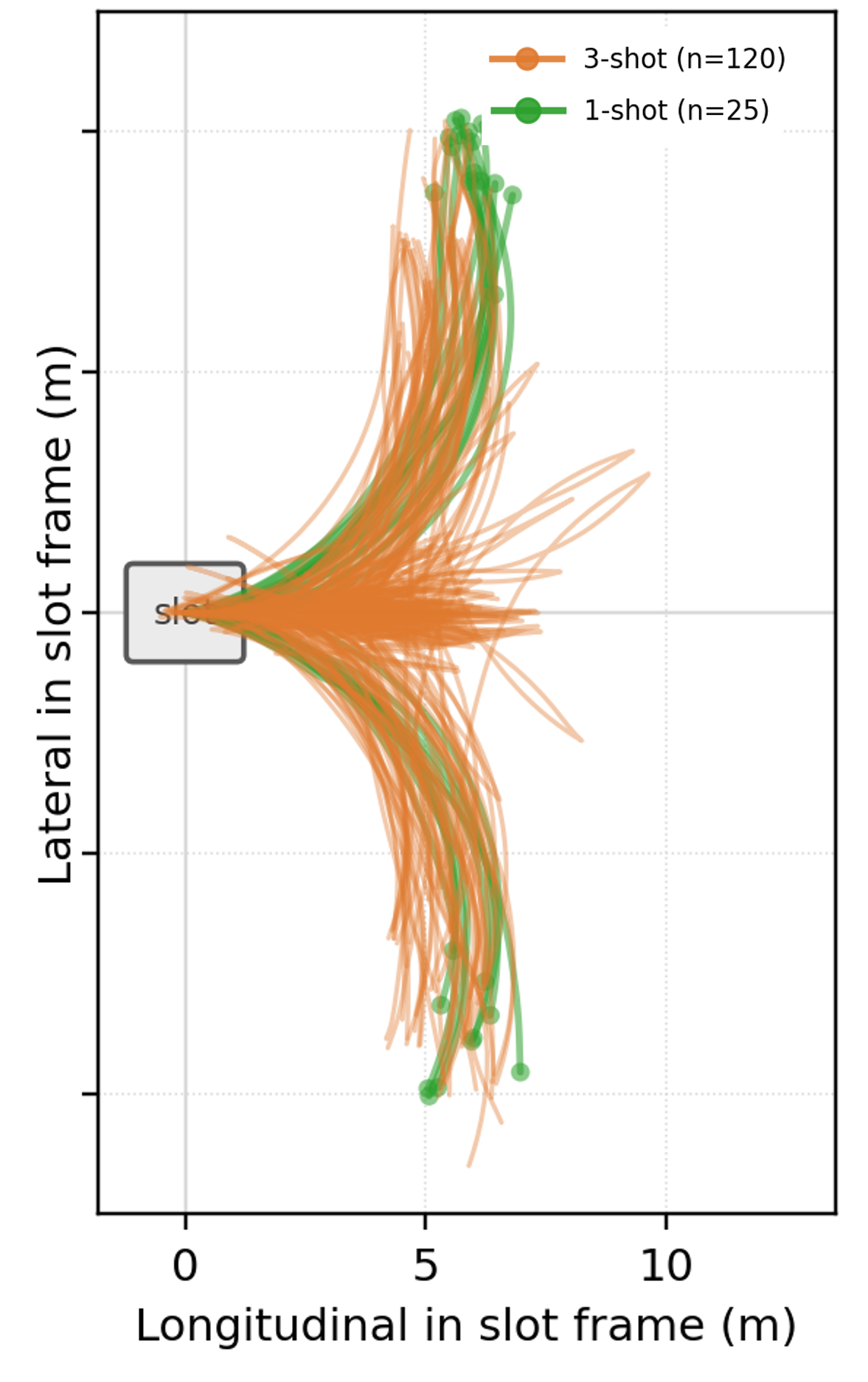}
        \caption{Trajectory Distribution}
        \label{fig:trajectory_distribution}
    \end{subfigure}%
}{Start poses and trajectories of 1-shot vs.\ 3-shot maneuvers.
Human experts strongly prefer multi-shot park-ins: even for starts in a similar geometric envelope, 
3-shot trajectories dominate the dataset, 
while 1-shot cases are rare and occur mainly at larger lateral offsets with a more spacious aisle.
\label{fig:start_traj_distribution}}

Accordingly, under human experts' preference for multi-shot park-ins, the SunnyParking dataset inherits the same style bias in its trajectory distribution.

\begin{table}[tbp]
\centering
\caption{Comparison of median starting geometry relative to the target pose across maneuver styles.}
\label{tab:start_geometry}
\begin{tabular}{lcccc}
\toprule
Metric (Median) & 1-shot & 2-shot & 3-shot & 4-shot \\
\midrule
Distance to goal & \textbf{11.03\,m} & 7.27\,m & \textbf{9.43\,m} & 7.45\,m \\
$|\mathrm{lateral}|$ offset & \textbf{9.14\,m} & 4.85\,m & \textbf{7.75\,m} & 4.79\,m \\
$|\Delta\mathrm{yaw}|$ & $91.6^{\circ}$ & $91.4^{\circ}$ & $90.6^{\circ}$ & $92.5^{\circ}$ \\
\bottomrule
\end{tabular}
\end{table}

\section{Motion State Ground Truth from Chassis Gear}
\label{appendix:motion_state_gt}

The motion state branch is trained as a two-class problem, $\{\text{forward}, \text{reverse}\}$.
On the raw chassis traces ($562$ tasks, $100{,}366$ frames; $80{,}164$ samples after the valid-slice trim in Appendix~\ref{appendix:dataset}), the gearbox takes only Drive ($32{,}730$ frames) and Reverse ($67{,}636$ frames), with no contradictory cases. 
A frame is marked stationary when $|v| \le \tau$, using the speed magnitude.
Stationary frames inherit the motion state of the preceding motion state segment instead of receiving their own class.

Table~\ref{tab:motion_state_tau} sweeps $\tau \in \{0.01, 0.02, 0.05, 0.10\}\,\mathrm{m/s}$. 
Raising the threshold only changes how many frames are judged as dwell, from $9.8\%$ at $0.01\,\mathrm{m/s}$ to $15.3\%$ at $0.10\,\mathrm{m/s}$. 
Among frames that remain moving, the forward/reverse length share stays within $0.42$ percentage points of the gear-only split ($32.61/67.39$). 
No threshold produces a forward$\leftrightarrow$reverse flip. 
After each dwell inherits the preceding segment, every $\tau$ matches the gear-derived labels on $100\%$ of frames, which is the ground truth used by the trainer. 

\begin{table}[tbp]
\centering
\caption{Speed-threshold sweep for motion state ground truth ($100{,}366$ raw frames, $562$ tasks). F/R length is the share of frames with $|v| > \tau$. Agreement is the frame match with the gear-derived labels after dwells inherit the preceding segment.}
\label{tab:motion_state_tau}
\setlength{\tabcolsep}{4pt}%
\renewcommand{\arraystretch}{1.05}%
\begin{tabular}{lccc}
\toprule
$\tau$ (m/s) & Dwell (\%) & F/R length & Agreement (\%) \\
\midrule
gear only & $0$     & $32.61/67.39$ & $100$ \\
$0.01$    & $9.77$  & $32.28/67.72$ & $100$ \\
$0.02$    & $10.66$ & $32.27/67.73$ & $100$ \\
$0.05$    & $12.95$ & $32.23/67.77$ & $100$ \\
$0.10$    & $15.29$ & $32.19/67.81$ & $100$ \\
\bottomrule
\end{tabular}
\end{table}

\bibliographystyle{IEEEtran}
\bibliography{ref_bibtex}


\begin{IEEEbiography}[{\includegraphics[width=1in,height=1.25in,clip,keepaspectratio]{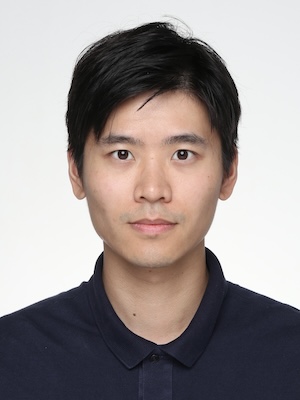}}]{Jishu Miao} received the B.E. degree in Mechatronics Engineering from the School of Mechanical and Electrical Engineering, Shaoxing University, China in 2018. 
    He received the M.Eng. degree in Computer Science from the Department of Computer Science, Chubu University, Japan in 2021. 
    He is currently a part-time Ph.D. student in the Machine Perception and Robotics Group at Chubu University. 
    His research interests include autonomous driving, 3D scene reconstruction, and multimodal data fusion.
\end{IEEEbiography}

\begin{IEEEbiography}[{\includegraphics[width=1in,height=1.25in,clip,keepaspectratio]{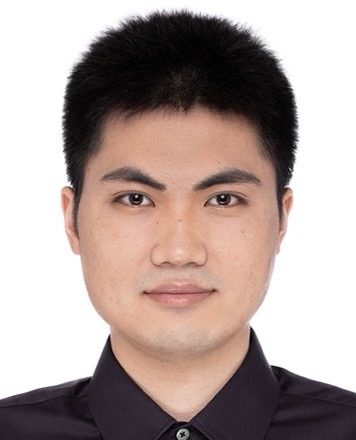}}]{Han Chen} received the B.S. and M.S. degrees in Software Engineering from the School of Software Engineering, Tongji University, China in 2016 and 2021, respectively. 
    During his graduate studies from 2018 to 2021, 
    his research focused on intelligent transportation systems. 
    He is currently pursuing a part-time graduate degree at the College of Automotive Engineering, Tongji University, China, where his research interests include autonomous driving in parking scenarios.
\end{IEEEbiography}

\begin{IEEEbiography}[{\includegraphics[width=1in,height=1.25in,clip,keepaspectratio]{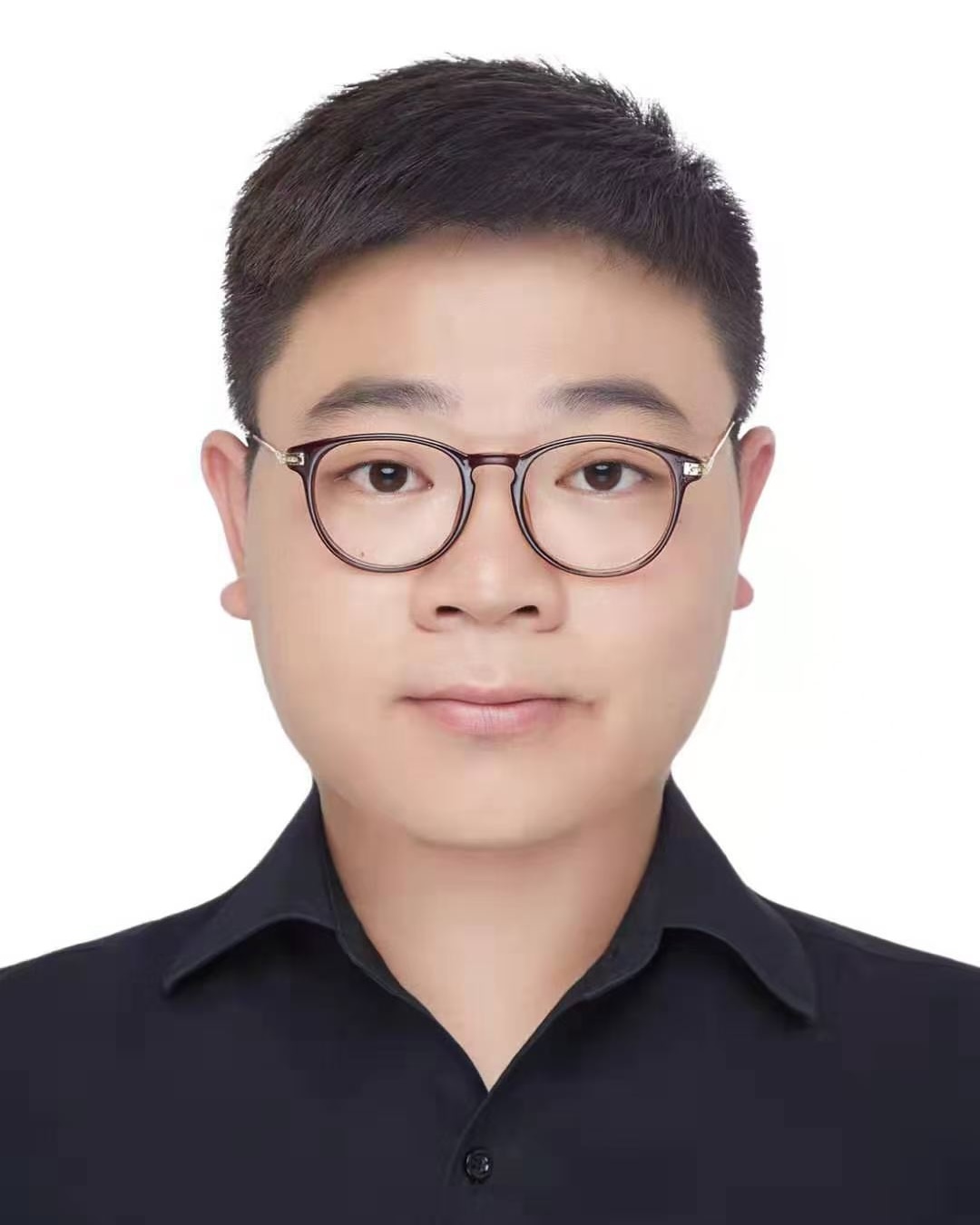}}]{Jiankun Zhai} received the B.S. degree in mechanical engineering and the M.S. degree in vehicle engineering from Beijing Institute of Technology, Beijing, China, in 2018 and 2021, respectively. 
    From 2021 to January 2026, he worked in the field of autonomous driving. 
    Since March 2026, he has been pursuing the Ph.D. degree in mechanical and electrical engineering with Harbin Institute of Technology, Harbin, China. 
    His research interests include medical robotics, surgical robotics, robotic control, and intelligent robotic systems.
\end{IEEEbiography}

\begin{IEEEbiography}[{\includegraphics[width=1in,height=1.25in,clip,keepaspectratio]{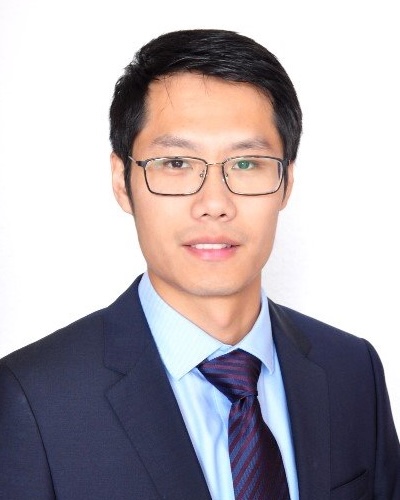}}]{Qi Liu} received his Bachelor's degree in Mechanical Engineering and Automation from Zhejiang University, China, and his Master's degree in Commercial Vehicle Technology and Ph.D. from the Robotic Research Lab at Technische Universität Kaiserslautern, Germany.
    His research interests include autonomous driving, robotics, algorithm design, and vehicle software engineering. 
\end{IEEEbiography}

\begin{IEEEbiography}[{\includegraphics[width=1in,height=1.25in,clip,keepaspectratio]{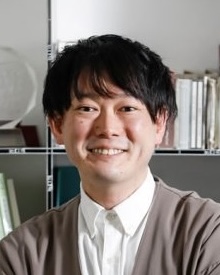}}]{Tsubasa Hirakawa} received his Ph.D. degree from the Department of Computer Science, Hiroshima University, Japan, in 2017. 
    From 2017 to 2019 he was a research fellow at Chubu University. 
    He was a designated associate professor of the Chubu Institute for Advanced Studies, Chubu University, Japan from 2019 to 2021. 
    He is now a lecturer of Chubu University, Japan since 2021. He was a Fellowship of the Japan Society for the Promotion of Science from 2014 to 2017. 
    He was a visiting researcher at ESIEE Paris, France, in 2014 and 2015.
\end{IEEEbiography}

\begin{IEEEbiography}[{\includegraphics[width=1in,height=1.25in,clip,keepaspectratio]{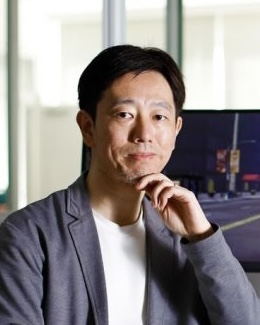}}]{Takayoshi Yamashita} received his Ph.D. degree from the Department of Computer Science, Chubu University, Japan, in 2011. 
    He worked in OMRON Corporation from 2002 to 2014. He was a lecturer and an associate professor of the Department of Computer Science, Chubu University, Japan, from 2014 to 2017 and from 2017 to 2021, respectively. 
    He is now a professor of the Department of Computer Science, Chubu University, Japan since 2021. 
    His current research interests include object detection, object tracking, human activity understanding, pattern recognition and machine learning. 
    He is a member of the IEEE, the IEICE and the IPSJ.
\end{IEEEbiography}

\begin{IEEEbiography}[{\includegraphics[width=1in,height=1.25in,clip,keepaspectratio]{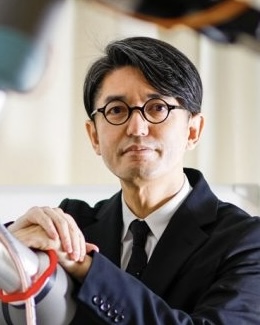}}]{Hironobu Fujiyoshi} received his Ph.D. in Electrical Engineering from Chubu University, Japan, in 1997. 
    From 1997 to 2000 he was a post-doctoral fellow at the Robotics Institute of Carnegie Mellon University, Pittsburgh, PA, USA, working on the DARPA Video Surveillance and Monitoring (VSAM) effort and the humanoid vision project for the HONDA Humanoid Robot. 
    He is now a professor of the Department of Robotics, Chubu University, Japan. From 2005 to 2006, he was a visiting researcher at Robotics Institute, Carnegie Mellon University. 
    His research interests include computer vision, video understanding and pattern recognition. 
    He is a member of the IEEE, the IEICE, the IPSJ, and the IEE.
\end{IEEEbiography}
\EOD

\end{document}